\documentclass[11pt]{article}
\usepackage[preprint]{acl}
\setcitestyle{authoryear,round,citesep={;},aysep={,},yysep={,}}
\usepackage{times}
\usepackage{latexsym}
\usepackage[T1]{fontenc}
\usepackage[utf8]{inputenc}
\usepackage{microtype}
\usepackage{xurl}
\usepackage{inconsolata}
\usepackage{graphicx}
\usepackage{booktabs}
\usepackage{multirow}
\usepackage{amsmath}
\usepackage{amssymb}
\usepackage{xcolor}
\usepackage{fontawesome5}
\usepackage{array}
\usepackage{colortbl}
\usepackage{tikz}
\usetikzlibrary{arrows.meta,positioning,fit,backgrounds}
\definecolor{ours}{HTML}{E8F1FB}
\definecolor{head}{HTML}{F3F3F3}
\definecolor{gain}{HTML}{1E7B34}
\definecolor{loss}{HTML}{B3261E}
\newcommand{\se}[1]{\,{\scriptsize$\pm$#1}}
\providecommand{\num}[1]{\ifcsname num@#1\endcsname\csname num@#1\endcsname\else\textcolor{red}{??#1??}\fi}
\expandafter\def\csname num@aob-advisory-clean\endcsname{85.5}
\expandafter\def\csname num@aob-advisory-correct\endcsname{11.4}
\expandafter\def\csname num@aob-advisory-unknown\endcsname{25.4}
\expandafter\def\csname num@aob-advisory-wrong\endcsname{63.1}
\expandafter\def\csname num@aob-advisoryhint-clean\endcsname{82.5}
\expandafter\def\csname num@aob-advisoryhint-correct\endcsname{9.8}
\expandafter\def\csname num@aob-advisoryhint-unknown\endcsname{52.9}
\expandafter\def\csname num@aob-advisoryhint-wrong\endcsname{37.2}
\expandafter\def\csname num@aob-checklist-clean\endcsname{84.1}
\expandafter\def\csname num@aob-checklist-correct\endcsname{16.2}
\expandafter\def\csname num@aob-checklist-unknown\endcsname{11.5}
\expandafter\def\csname num@aob-checklist-wrong\endcsname{72.3}
\expandafter\def\csname num@aob-enforce2-clean\endcsname{86.3}
\expandafter\def\csname num@aob-enforce2-correct\endcsname{3.0}
\expandafter\def\csname num@aob-enforce2-unknown\endcsname{88.2}
\expandafter\def\csname num@aob-enforce2-wrong\endcsname{8.8}
\expandafter\def\csname num@aob-gate2-clean\endcsname{84.9}
\expandafter\def\csname num@aob-gate2-correct\endcsname{5.2}
\expandafter\def\csname num@aob-gate2-unknown\endcsname{61.9}
\expandafter\def\csname num@aob-gate2-wrong\endcsname{32.9}
\expandafter\def\csname num@aob-judgecheck-clean\endcsname{86.5}
\expandafter\def\csname num@aob-judgecheck-correct\endcsname{17.4}
\expandafter\def\csname num@aob-judgecheck-unknown\endcsname{11.4}
\expandafter\def\csname num@aob-judgecheck-wrong\endcsname{71.2}
\expandafter\def\csname num@aob-none-clean\endcsname{84.9}
\expandafter\def\csname num@aob-none-correct\endcsname{16.4}
\expandafter\def\csname num@aob-none-unknown\endcsname{9.6}
\expandafter\def\csname num@aob-none-wrong\endcsname{74.0}
\expandafter\def\csname num@aob-om-clean\endcsname{86.1}
\expandafter\def\csname num@aob-om-correct\endcsname{16.6}
\expandafter\def\csname num@aob-om-unknown\endcsname{11.2}
\expandafter\def\csname num@aob-om-wrong\endcsname{72.2}
\expandafter\def\csname num@aob-pub-wrong-max\endcsname{73.8}
\expandafter\def\csname num@aob-pub-wrong-min\endcsname{71.2}
\expandafter\def\csname num@aob-repair3-clean\endcsname{83.3}
\expandafter\def\csname num@aob-repair3-correct\endcsname{57.3}
\expandafter\def\csname num@aob-repair3-unknown\endcsname{39.7}
\expandafter\def\csname num@aob-repair3-wrong\endcsname{3.0}
\expandafter\def\csname num@aob-requery-clean\endcsname{84.7}
\expandafter\def\csname num@aob-requery-correct\endcsname{17.0}
\expandafter\def\csname num@aob-requery-unknown\endcsname{8.1}
\expandafter\def\csname num@aob-requery-wrong\endcsname{74.9}
\expandafter\def\csname num@aob-retry2-clean\endcsname{88.3}
\expandafter\def\csname num@aob-retry2-correct\endcsname{17.3}
\expandafter\def\csname num@aob-retry2-unknown\endcsname{8.3}
\expandafter\def\csname num@aob-retry2-wrong\endcsname{74.4}
\expandafter\def\csname num@aob-reviewer-clean\endcsname{80.4}
\expandafter\def\csname num@aob-reviewer-correct\endcsname{15.8}
\expandafter\def\csname num@aob-reviewer-unknown\endcsname{10.4}
\expandafter\def\csname num@aob-reviewer-wrong\endcsname{73.8}
\expandafter\def\csname num@aob-selfverify-clean\endcsname{76.6}
\expandafter\def\csname num@aob-selfverify-correct\endcsname{14.2}
\expandafter\def\csname num@aob-selfverify-unknown\endcsname{29.6}
\expandafter\def\csname num@aob-selfverify-wrong\endcsname{56.2}
\expandafter\def\csname num@aob-sgp-clean\endcsname{84.9}
\expandafter\def\csname num@aob-sgp-correct\endcsname{2.4}
\expandafter\def\csname num@aob-sgp-unknown\endcsname{95.8}
\expandafter\def\csname num@aob-sgp-vs-checklist-wrong-fewer\endcsname{7}
\expandafter\def\csname num@aob-sgp-vs-checklist-wrong-n\endcsname{7}
\expandafter\def\csname num@aob-sgp-vs-checklist-wrong-p\endcsname{0.016}
\expandafter\def\csname num@aob-sgp-vs-enforce2-wrong-fewer\endcsname{6}
\expandafter\def\csname num@aob-sgp-vs-enforce2-wrong-n\endcsname{7}
\expandafter\def\csname num@aob-sgp-vs-enforce2-wrong-p\endcsname{0.031}
\expandafter\def\csname num@aob-sgp-vs-judge-check-wrong-fewer\endcsname{7}
\expandafter\def\csname num@aob-sgp-vs-judge-check-wrong-n\endcsname{7}
\expandafter\def\csname num@aob-sgp-vs-judge-check-wrong-p\endcsname{0.016}
\expandafter\def\csname num@aob-sgp-vs-none-wrong-fewer\endcsname{7}
\expandafter\def\csname num@aob-sgp-vs-none-wrong-n\endcsname{7}
\expandafter\def\csname num@aob-sgp-vs-none-wrong-p\endcsname{0.016}
\expandafter\def\csname num@aob-sgp-vs-om-wrong-fewer\endcsname{7}
\expandafter\def\csname num@aob-sgp-vs-om-wrong-n\endcsname{7}
\expandafter\def\csname num@aob-sgp-vs-om-wrong-p\endcsname{0.016}
\expandafter\def\csname num@aob-sgp-vs-repair3-wrong-fewer\endcsname{4}
\expandafter\def\csname num@aob-sgp-vs-repair3-wrong-n\endcsname{7}
\expandafter\def\csname num@aob-sgp-vs-repair3-wrong-p\endcsname{0.438}
\expandafter\def\csname num@aob-sgp-vs-retry2-wrong-fewer\endcsname{7}
\expandafter\def\csname num@aob-sgp-vs-retry2-wrong-n\endcsname{7}
\expandafter\def\csname num@aob-sgp-vs-retry2-wrong-p\endcsname{0.016}
\expandafter\def\csname num@aob-sgp-vs-reviewer-wrong-fewer\endcsname{7}
\expandafter\def\csname num@aob-sgp-vs-reviewer-wrong-n\endcsname{7}
\expandafter\def\csname num@aob-sgp-vs-reviewer-wrong-p\endcsname{0.016}
\expandafter\def\csname num@aob-sgp-wrong\endcsname{1.7}
\expandafter\def\csname num@aob-skeptic-clean\endcsname{86.1}
\expandafter\def\csname num@aob-skeptic-correct\endcsname{16.4}
\expandafter\def\csname num@aob-skeptic-unknown\endcsname{10.3}
\expandafter\def\csname num@aob-skeptic-wrong\endcsname{73.3}
\expandafter\def\csname num@api-guard-cleancorrect\endcsname{91.1}
\expandafter\def\csname num@api-guard-correct\endcsname{25.7}
\expandafter\def\csname num@api-guard-wrong\endcsname{17.8}
\expandafter\def\csname num@api-none-cleancorrect\endcsname{92.1}
\expandafter\def\csname num@api-none-correct\endcsname{3.5}
\expandafter\def\csname num@api-none-wrong\endcsname{89.8}
\expandafter\def\csname num@api-retry2-cleancorrect\endcsname{93.0}
\expandafter\def\csname num@api-retry2-correct\endcsname{3.2}
\expandafter\def\csname num@api-retry2-wrong\endcsname{89.5}
\expandafter\def\csname num@api-schema-cleancorrect\endcsname{92.1}
\expandafter\def\csname num@api-schema-correct\endcsname{70.8}
\expandafter\def\csname num@api-schema-wrong\endcsname{21.0}
\expandafter\def\csname num@api-sgp-cleancorrect\endcsname{92.4}
\expandafter\def\csname num@api-sgp-correct\endcsname{69.5}
\expandafter\def\csname num@api-sgp-wrong\endcsname{21.3}
\expandafter\def\csname num@cost-enforce-calls\endcsname{5.0}
\expandafter\def\csname num@cost-enforce-secs\endcsname{29.1}
\expandafter\def\csname num@cost-enforce-tokens\endcsname{11{,}745}
\expandafter\def\csname num@cost-judge-calls\endcsname{3.2}
\expandafter\def\csname num@cost-judge-secs\endcsname{47.2}
\expandafter\def\csname num@cost-judge-tokens\endcsname{6{,}287}
\expandafter\def\csname num@cost-none-calls\endcsname{2.0}
\expandafter\def\csname num@cost-none-secs\endcsname{9.2}
\expandafter\def\csname num@cost-none-tokens\endcsname{3{,}183}
\expandafter\def\csname num@cost-reviewer-calls\endcsname{1.9}
\expandafter\def\csname num@cost-reviewer-secs\endcsname{25.0}
\expandafter\def\csname num@cost-reviewer-tokens\endcsname{9{,}276}
\expandafter\def\csname num@cost-sgp-calls\endcsname{3.8}
\expandafter\def\csname num@cost-sgp-secs\endcsname{18.9}
\expandafter\def\csname num@cost-sgp-tokens\endcsname{9{,}055}
\expandafter\def\csname num@e3n-apis\endcsname{15}
\expandafter\def\csname num@e3n-maxinfl\endcsname{691}
\expandafter\def\csname num@e3n-silent\endcsname{10}
\expandafter\def\csname num@ob-none-correct\endcsname{84}
\expandafter\def\csname num@ob-none-unknown\endcsname{0}
\expandafter\def\csname num@ob-none-wrong\endcsname{132}
\expandafter\def\csname num@ob-rep-correct\endcsname{79}
\expandafter\def\csname num@ob-rep-unknown\endcsname{35}
\expandafter\def\csname num@ob-rep-wrong\endcsname{102}
\expandafter\def\csname num@ob-sgp-correct\endcsname{41}
\expandafter\def\csname num@ob-sgp-unknown\endcsname{122}
\expandafter\def\csname num@ob-sgp-wrong\endcsname{53}
\expandafter\def\csname num@ob-vbw\endcsname{49}
\expandafter\def\csname num@ra-blindr-perrun\endcsname{2.36}
\expandafter\def\csname num@ra-blindr-share\endcsname{88.1}
\expandafter\def\csname num@ra-ours-perrun\endcsname{0.24}
\expandafter\def\csname num@ra-ours-share\endcsname{22.6}
\expandafter\def\csname num@ret-advisory-clean\endcsname{35.7}
\expandafter\def\csname num@ret-advisory-correct\endcsname{7.7}
\expandafter\def\csname num@ret-advisory-wrong\endcsname{71.5}
\expandafter\def\csname num@ret-advisoryhint-clean\endcsname{36.9}
\expandafter\def\csname num@ret-advisoryhint-correct\endcsname{8.9}
\expandafter\def\csname num@ret-advisoryhint-wrong\endcsname{59.6}
\expandafter\def\csname num@ret-enforce2-clean\endcsname{33.8}
\expandafter\def\csname num@ret-enforce2-correct\endcsname{6.5}
\expandafter\def\csname num@ret-enforce2-wrong\endcsname{12.3}
\expandafter\def\csname num@ret-none-clean\endcsname{36.2}
\expandafter\def\csname num@ret-none-correct\endcsname{7.2}
\expandafter\def\csname num@ret-none-wrong\endcsname{78.6}
\expandafter\def\csname num@ret-repair3-clean\endcsname{35.7}
\expandafter\def\csname num@ret-repair3-correct\endcsname{30.5}
\expandafter\def\csname num@ret-repair3-wrong\endcsname{15.0}
\expandafter\def\csname num@ret-sgp-clean\endcsname{38.5}
\expandafter\def\csname num@ret-sgp-correct\endcsname{7.1}
\expandafter\def\csname num@ret-sgp-wrong\endcsname{11.0}
\expandafter\def\csname num@tm-blindr-clean\endcsname{54.2}
\expandafter\def\csname num@tm-blindr-p3\endcsname{60.4}
\expandafter\def\csname num@tm-blindr-p4\endcsname{27.0}
\expandafter\def\csname num@tm-blindr-silent\endcsname{43.7}
\expandafter\def\csname num@tm-checklist-clean\endcsname{55.5}
\expandafter\def\csname num@tm-checklist-loud\endcsname{38.3}
\expandafter\def\csname num@tm-checklist-p3\endcsname{26.8}
\expandafter\def\csname num@tm-checklist-p4\endcsname{19.5}
\expandafter\def\csname num@tm-checklist-silent\endcsname{23.2}
\expandafter\def\csname num@tm-om-clean\endcsname{57.1}
\expandafter\def\csname num@tm-om-loud\endcsname{40.7}
\expandafter\def\csname num@tm-om-p3\endcsname{36.3}
\expandafter\def\csname num@tm-om-p4\endcsname{24.8}
\expandafter\def\csname num@tm-om-silent\endcsname{30.5}
\expandafter\def\csname num@tm-omcl-clean\endcsname{54.8}
\expandafter\def\csname num@tm-omcl-loud\endcsname{39.8}
\expandafter\def\csname num@tm-omcl-p3\endcsname{39.9}
\expandafter\def\csname num@tm-omcl-p4\endcsname{25.6}
\expandafter\def\csname num@tm-omcl-silent\endcsname{32.7}
\expandafter\def\csname num@tm-ours-clean\endcsname{52.5}
\expandafter\def\csname num@tm-ours-loud\endcsname{39.8}
\expandafter\def\csname num@tm-ours-p3\endcsname{47.6}
\expandafter\def\csname num@tm-ours-p4\endcsname{24.4}
\expandafter\def\csname num@tm-ours-silent\endcsname{36.0}
\expandafter\def\csname num@tmtwo-blindr-p3\endcsname{56.0}
\expandafter\def\csname num@tmtwo-n\endcsname{7}
\expandafter\def\csname num@tmtwo-omcl-p3\endcsname{32.5}
\expandafter\def\csname num@tmtwo-ours-p3\endcsname{46.4}
\expandafter\def\csname num@tmtwo-oursminusblindr-diff\endcsname{-9.6}
\expandafter\def\csname num@tmtwo-oursminusblindr-p\endcsname{0.016}
\expandafter\def\csname num@tmtwo-oursminusblindr-wins\endcsname{0}
\expandafter\def\csname num@tmtwo-oursminusomcl-diff\endcsname{13.9}
\expandafter\def\csname num@tmtwo-oursminusomcl-p\endcsname{0.016}
\expandafter\def\csname num@tmtwo-oursminusomcl-wins\endcsname{7}
\expandafter\def\csname num@toolmaze34-ours-vs-blindr-diff\endcsname{-7.7}
\expandafter\def\csname num@toolmaze34-ours-vs-blindr-n\endcsname{14}
\expandafter\def\csname num@toolmaze34-ours-vs-blindr-p\endcsname{0.000}
\expandafter\def\csname num@toolmaze34-ours-vs-blindr-wins\endcsname{0}
\expandafter\def\csname num@toolmaze34-ours-vs-checklist-diff\endcsname{12.8}
\expandafter\def\csname num@toolmaze34-ours-vs-checklist-n\endcsname{14}
\expandafter\def\csname num@toolmaze34-ours-vs-checklist-p\endcsname{0.000}
\expandafter\def\csname num@toolmaze34-ours-vs-checklist-wins\endcsname{14}
\expandafter\def\csname num@toolmaze34-ours-vs-om-diff\endcsname{5.5}
\expandafter\def\csname num@toolmaze34-ours-vs-om-n\endcsname{14}
\expandafter\def\csname num@toolmaze34-ours-vs-om-p\endcsname{0.011}
\expandafter\def\csname num@toolmaze34-ours-vs-om-wins\endcsname{12}
\expandafter\def\csname num@toolmaze34-ours-vs-omcl-diff\endcsname{3.4}
\expandafter\def\csname num@toolmaze34-ours-vs-omcl-n\endcsname{14}
\expandafter\def\csname num@toolmaze34-ours-vs-omcl-p\endcsname{0.031}
\expandafter\def\csname num@toolmaze34-ours-vs-omcl-wins\endcsname{11}

\expandafter\def\csname num@aob-judgeblock-clean\endcsname{81.7}
\expandafter\def\csname num@aob-judgeblock-wrong\endcsname{6.1}
\expandafter\def\csname num@aob-judgeblockplain-clean\endcsname{83.7}
\expandafter\def\csname num@aob-judgeblockplain-wrong\endcsname{70.8}
\expandafter\def\csname num@aob-omblock-clean\endcsname{86.5}
\expandafter\def\csname num@aob-omblock-wrong\endcsname{70.9}
\expandafter\def\csname num@api-first4-none-correct\endcsname{2.9}
\expandafter\def\csname num@api-first4-none-wrong\endcsname{87.9}
\expandafter\def\csname num@api-first4-sgp-correct\endcsname{50.0}
\expandafter\def\csname num@api-first4-sgp-wrong\endcsname{38.6}
\expandafter\def\csname num@api-later5-none-correct\endcsname{4.0}
\expandafter\def\csname num@api-later5-none-wrong\endcsname{91.4}
\expandafter\def\csname num@api-later5-sgp-correct\endcsname{85.1}
\expandafter\def\csname num@api-later5-sgp-wrong\endcsname{7.4}
\expandafter\def\csname num@api-typed-wrong\endcsname{28.6}
\expandafter\def\csname num@api-typedsgp-wrong\endcsname{28.3}
\expandafter\def\csname num@bf-models\endcsname{7}
\expandafter\def\csname num@bf-none-cleancorrect\endcsname{88.6}
\expandafter\def\csname num@bf-none-correct\endcsname{0.0}
\expandafter\def\csname num@bf-none-unknown\endcsname{2.9}
\expandafter\def\csname num@bf-none-wrong\endcsname{97.1}
\expandafter\def\csname num@bf-schema-correct\endcsname{0.0}
\expandafter\def\csname num@bf-schema-unknown\endcsname{0.0}
\expandafter\def\csname num@bf-schema-wrong\endcsname{100.0}
\expandafter\def\csname num@bf-sgp-correct\endcsname{2.9}
\expandafter\def\csname num@bf-sgp-unknown\endcsname{77.1}
\expandafter\def\csname num@bf-sgp-wrong\endcsname{20.0}
\expandafter\def\csname num@bf-sgp-wrong-max\endcsname{80}
\expandafter\def\csname num@boot-tm-hi\endcsname{4.9}
\expandafter\def\csname num@boot-tm-lo\endcsname{1.8}
\expandafter\def\csname num@bud-all-calls\endcsname{5.2}
\expandafter\def\csname num@bud-all-caught\endcsname{97.2}
\expandafter\def\csname num@bud-all-lie\endcsname{2.8}
\expandafter\def\csname num@bud-all-wrong\endcsname{0.0}
\expandafter\def\csname num@bud-clean-fa\endcsname{0.0}
\expandafter\def\csname num@bud-dropfield-all\endcsname{83.3}
\expandafter\def\csname num@bud-dropfield-r1\endcsname{33.3}
\expandafter\def\csname num@bud-last-calls\endcsname{2.0}
\expandafter\def\csname num@bud-last-caught\endcsname{88.0}
\expandafter\def\csname num@bud-last-lie\endcsname{0.0}
\expandafter\def\csname num@bud-last-wrong\endcsname{8.8}
\expandafter\def\csname num@bud-none-wrong\endcsname{80.6}
\expandafter\def\csname num@bud-r1-calls\endcsname{2.0}
\expandafter\def\csname num@bud-r1-caught\endcsname{88.4}
\expandafter\def\csname num@bud-r1-lie\endcsname{0.9}
\expandafter\def\csname num@bud-r1-wrong\endcsname{8.6}
\expandafter\def\csname num@bud-r2-calls\endcsname{4.0}
\expandafter\def\csname num@bud-r2-caught\endcsname{94.4}
\expandafter\def\csname num@bud-r2-lie\endcsname{1.9}
\expandafter\def\csname num@bud-r2-wrong\endcsname{2.8}
\expandafter\def\csname num@bud-seq-calls\endcsname{2.3}
\expandafter\def\csname num@bud-seq-caught\endcsname{97.2}
\expandafter\def\csname num@bud-seq-lie\endcsname{2.8}
\expandafter\def\csname num@bud-seq-wrong\endcsname{0.0}
\expandafter\def\csname num@bv-fa\endcsname{0}
\expandafter\def\csname num@bv-fired\endcsname{15}
\expandafter\def\csname num@bv-ignored\endcsname{15}
\expandafter\def\csname num@bv-n\endcsname{15}
\expandafter\def\csname num@chk-apis\endcsname{13}
\expandafter\def\csname num@chk-exact\endcsname{61}
\expandafter\def\csname num@chk-exact-pct\endcsname{73}
\expandafter\def\csname num@chk-n\endcsname{84}
\expandafter\def\csname num@cost-judge-tokens\endcsname{6{,}287}
\expandafter\def\csname num@cost-none-tokens\endcsname{3{,}183}
\expandafter\def\csname num@cost-sgp-tokens\endcsname{9{,}055}
\expandafter\def\csname num@eg-enf-requery\endcsname{130}
\expandafter\def\csname num@eg-enf-wrong\endcsname{266}
\expandafter\def\csname num@eg-sgp-requery\endcsname{13}
\expandafter\def\csname num@eg-sgp-wrong\endcsname{52}
\expandafter\def\csname num@enf-max\endcsname{24.8}
\expandafter\def\csname num@enf-min\endcsname{0.5}
\expandafter\def\csname num@fk-any\endcsname{11}
\expandafter\def\csname num@fk-bad\endcsname{3}
\expandafter\def\csname num@fk-cap\endcsname{6}
\expandafter\def\csname num@fk-n\endcsname{15}
\expandafter\def\csname num@fk-reach\endcsname{14}
\expandafter\def\csname num@fr-enforce-clean\endcsname{92.6}
\expandafter\def\csname num@fr-enforce-clean-max\endcsname{100.0}
\expandafter\def\csname num@fr-enforce-clean-min\endcsname{87.5}
\expandafter\def\csname num@fr-enforce-wrong\endcsname{33.9}
\expandafter\def\csname num@fr-enforce-wrong-max\endcsname{36.1}
\expandafter\def\csname num@fr-enforce-wrong-min\endcsname{31.5}
\expandafter\def\csname num@fr-luna-enforce-clean\endcsname{87.5}
\expandafter\def\csname num@fr-luna-enforce-wrong\endcsname{36.1}
\expandafter\def\csname num@fr-luna-none-clean\endcsname{84.7}
\expandafter\def\csname num@fr-luna-none-wrong\endcsname{59.0}
\expandafter\def\csname num@fr-luna-permission-clean\endcsname{73.6}
\expandafter\def\csname num@fr-luna-permission-wrong\endcsname{45.1}
\expandafter\def\csname num@fr-luna-selfverify-clean\endcsname{16.7}
\expandafter\def\csname num@fr-luna-selfverify-wrong\endcsname{6.9}
\expandafter\def\csname num@fr-luna-sgp-clean\endcsname{95.8}
\expandafter\def\csname num@fr-luna-sgp-wrong\endcsname{1.4}
\expandafter\def\csname num@fr-luna-warning-clean\endcsname{69.4}
\expandafter\def\csname num@fr-luna-warning-wrong\endcsname{63.2}
\expandafter\def\csname num@fr-none-clean\endcsname{92.1}
\expandafter\def\csname num@fr-none-clean-max\endcsname{98.6}
\expandafter\def\csname num@fr-none-clean-min\endcsname{84.7}
\expandafter\def\csname num@fr-none-wrong\endcsname{70.2}
\expandafter\def\csname num@fr-none-wrong-max\endcsname{89.4}
\expandafter\def\csname num@fr-none-wrong-min\endcsname{59.0}
\expandafter\def\csname num@fr-permission-clean\endcsname{78.7}
\expandafter\def\csname num@fr-permission-clean-max\endcsname{86.1}
\expandafter\def\csname num@fr-permission-clean-min\endcsname{73.6}
\expandafter\def\csname num@fr-permission-wrong\endcsname{27.8}
\expandafter\def\csname num@fr-permission-wrong-max\endcsname{45.1}
\expandafter\def\csname num@fr-permission-wrong-min\endcsname{2.3}
\expandafter\def\csname num@fr-selfverify-clean\endcsname{20.4}
\expandafter\def\csname num@fr-selfverify-clean-max\endcsname{29.2}
\expandafter\def\csname num@fr-selfverify-clean-min\endcsname{15.3}
\expandafter\def\csname num@fr-selfverify-wrong\endcsname{4.5}
\expandafter\def\csname num@fr-selfverify-wrong-max\endcsname{6.9}
\expandafter\def\csname num@fr-selfverify-wrong-min\endcsname{2.5}
\expandafter\def\csname num@fr-sgp-clean\endcsname{98.1}
\expandafter\def\csname num@fr-sgp-clean-max\endcsname{100.0}
\expandafter\def\csname num@fr-sgp-clean-min\endcsname{95.8}
\expandafter\def\csname num@fr-sgp-wrong\endcsname{0.5}
\expandafter\def\csname num@fr-sgp-wrong-max\endcsname{1.4}
\expandafter\def\csname num@fr-sgp-wrong-min\endcsname{0.0}
\expandafter\def\csname num@fr-sol-enforce-clean\endcsname{100.0}
\expandafter\def\csname num@fr-sol-enforce-wrong\endcsname{31.5}
\expandafter\def\csname num@fr-sol-none-clean\endcsname{98.6}
\expandafter\def\csname num@fr-sol-none-wrong\endcsname{89.4}
\expandafter\def\csname num@fr-sol-permission-clean\endcsname{86.1}
\expandafter\def\csname num@fr-sol-permission-wrong\endcsname{2.3}
\expandafter\def\csname num@fr-sol-selfverify-clean\endcsname{29.2}
\expandafter\def\csname num@fr-sol-selfverify-wrong\endcsname{2.5}
\expandafter\def\csname num@fr-sol-sgp-clean\endcsname{100.0}
\expandafter\def\csname num@fr-sol-sgp-wrong\endcsname{0.0}
\expandafter\def\csname num@fr-sol-warning-clean\endcsname{77.8}
\expandafter\def\csname num@fr-sol-warning-wrong\endcsname{11.3}
\expandafter\def\csname num@fr-terra-enforce-clean\endcsname{90.3}
\expandafter\def\csname num@fr-terra-enforce-wrong\endcsname{34.0}
\expandafter\def\csname num@fr-terra-none-clean\endcsname{93.1}
\expandafter\def\csname num@fr-terra-none-wrong\endcsname{62.3}
\expandafter\def\csname num@fr-terra-permission-clean\endcsname{76.4}
\expandafter\def\csname num@fr-terra-permission-wrong\endcsname{35.9}
\expandafter\def\csname num@fr-terra-selfverify-clean\endcsname{15.3}
\expandafter\def\csname num@fr-terra-selfverify-wrong\endcsname{3.9}
\expandafter\def\csname num@fr-terra-sgp-clean\endcsname{98.6}
\expandafter\def\csname num@fr-terra-sgp-wrong\endcsname{0.2}
\expandafter\def\csname num@fr-terra-warning-clean\endcsname{68.1}
\expandafter\def\csname num@fr-terra-warning-wrong\endcsname{53.9}
\expandafter\def\csname num@fr-warning-clean\endcsname{71.8}
\expandafter\def\csname num@fr-warning-clean-max\endcsname{77.8}
\expandafter\def\csname num@fr-warning-clean-min\endcsname{68.1}
\expandafter\def\csname num@fr-warning-wrong\endcsname{42.8}
\expandafter\def\csname num@fr-warning-wrong-max\endcsname{63.2}
\expandafter\def\csname num@fr-warning-wrong-min\endcsname{11.3}
\expandafter\def\csname num@frapi-none-clean\endcsname{100.0}
\expandafter\def\csname num@frapi-none-correct\endcsname{42.2}
\expandafter\def\csname num@frapi-none-wrong\endcsname{57.8}
\expandafter\def\csname num@frapi-sgp-correct\endcsname{90.4}
\expandafter\def\csname num@frapi-sgp-wrong\endcsname{8.9}
\expandafter\def\csname num@gain-aob-clean\endcsname{3.4}
\expandafter\def\csname num@gain-aob-wrong\endcsname{54.5}
\expandafter\def\csname num@gain-realapi-clean\endcsname{0.6}
\expandafter\def\csname num@gain-realapi-correct\endcsname{1.3}
\expandafter\def\csname num@gain-realapi-wrong\endcsname{0.3}
\expandafter\def\csname num@gain-retail-clean\endcsname{2.3}
\expandafter\def\csname num@gain-retail-correct\endcsname{23.4}
\expandafter\def\csname num@gain-retail-wrong\endcsname{1.3}
\expandafter\def\csname num@gain-toolmaze-clean\endcsname{4.6}
\expandafter\def\csname num@gain-toolmaze-loud\endcsname{0.9}
\expandafter\def\csname num@gain-toolmaze-silent\endcsname{7.7}
\expandafter\def\csname num@gainp-aob-clean\endcsname{0.750}
\expandafter\def\csname num@gainp-aob-wrong\endcsname{0.016}
\expandafter\def\csname num@gainp-realapi-clean\endcsname{0.938}
\expandafter\def\csname num@gainp-realapi-correct\endcsname{0.500}
\expandafter\def\csname num@gainp-realapi-wrong\endcsname{1.000}
\expandafter\def\csname num@gainp-retail-clean\endcsname{0.625}
\expandafter\def\csname num@gainp-retail-correct\endcsname{0.031}
\expandafter\def\csname num@gainp-retail-wrong\endcsname{0.469}
\expandafter\def\csname num@gainp-toolmaze-clean\endcsname{0.016}
\expandafter\def\csname num@gainp-toolmaze-loud\endcsname{0.531}
\expandafter\def\csname num@gainp-toolmaze-silent\endcsname{0.016}
\expandafter\def\csname num@gate-decisions\endcsname{2{,}448}
\expandafter\def\csname num@gate-maxsupport\endcsname{52}
\expandafter\def\csname num@gate-safe-tau\endcsname{1.85}
\expandafter\def\csname num@ig-adv-conf\endcsname{0.43}
\expandafter\def\csname num@ig-adv-mention\endcsname{40}
\expandafter\def\csname num@ig-clean-conf\endcsname{1.00}
\expandafter\def\csname num@ig-judge-high\endcsname{42}
\expandafter\def\csname num@ig-judge-mention\endcsname{26}
\expandafter\def\csname num@l1-gap7-seen\endcsname{2.1}
\expandafter\def\csname num@l1-gap7-unseen\endcsname{0.2}
\expandafter\def\csname num@lam-aob-none\endcsname{0.2}
\expandafter\def\csname num@lam-aob-repair\endcsname{43.6}
\expandafter\def\csname num@lam-listseen-none\endcsname{0.6}
\expandafter\def\csname num@lam-listseen-repair\endcsname{4.9}
\expandafter\def\csname num@lam-listunseen-none\endcsname{1.9}
\expandafter\def\csname num@lam-listunseen-repair\endcsname{5.7}
\expandafter\def\csname num@lam-unseen-none\endcsname{0.5}
\expandafter\def\csname num@lam-unseen-repair\endcsname{0.8}
\expandafter\def\csname num@learned-warn\endcsname{33.5}
\expandafter\def\csname num@learned-withhold\endcsname{32.1}
\expandafter\def\csname num@live-id-exact\endcsname{25}
\expandafter\def\csname num@live-id-n\endcsname{25}
\expandafter\def\csname num@open7-enforce-wrong\endcsname{8.8}
\expandafter\def\csname num@open7-none-clean\endcsname{84.9}
\expandafter\def\csname num@open7-permission-wrong\endcsname{18.2}
\expandafter\def\csname num@open7-selfverify-clean\endcsname{76.6}
\expandafter\def\csname num@pm-aob-n\endcsname{7}
\expandafter\def\csname num@pm-aob-p\endcsname{0.016}
\expandafter\def\csname num@pm-aob-wins\endcsname{7}
\expandafter\def\csname num@pm-tm-n\endcsname{7}
\expandafter\def\csname num@pm-tm-p\endcsname{0.125}
\expandafter\def\csname num@pm-tm-wins\endcsname{5}
\expandafter\def\csname num@prereg-held\endcsname{26}
\expandafter\def\csname num@prereg-n\endcsname{39}
\expandafter\def\csname num@rc-advisorynonumabstain-cov\endcsname{72}
\expandafter\def\csname num@rc-advisorynonumabstain-risk\endcsname{11.3}
\expandafter\def\csname num@rc-enforce2-cov\endcsname{70}
\expandafter\def\csname num@rc-enforce2-risk\endcsname{6.7}
\expandafter\def\csname num@rc-judgeblock-cov\endcsname{66}
\expandafter\def\csname num@rc-judgeblock-risk\endcsname{6.1}
\expandafter\def\csname num@rc-judgecheck-cov\endcsname{91}
\expandafter\def\csname num@rc-judgecheck-risk\endcsname{24.0}
\expandafter\def\csname num@rc-none-cov\endcsname{90}
\expandafter\def\csname num@rc-none-risk\endcsname{24.6}
\expandafter\def\csname num@rc-repair3-cov\endcsname{80}
\expandafter\def\csname num@rc-repair3-risk\endcsname{3.9}
\expandafter\def\csname num@rc-sgp-cov\endcsname{67}
\expandafter\def\csname num@rc-sgp-risk\endcsname{3.3}
\expandafter\def\csname num@rc10-advisorynonumabstain-cov\endcsname{82}
\expandafter\def\csname num@rc10-advisorynonumabstain-risk\endcsname{7.5}
\expandafter\def\csname num@rc10-enforce2-cov\endcsname{82}
\expandafter\def\csname num@rc10-enforce2-risk\endcsname{4.8}
\expandafter\def\csname num@rc10-judgeblock-cov\endcsname{78}
\expandafter\def\csname num@rc10-judgeblock-risk\endcsname{4.7}
\expandafter\def\csname num@rc10-judgecheck-cov\endcsname{92}
\expandafter\def\csname num@rc10-judgecheck-risk\endcsname{13.0}
\expandafter\def\csname num@rc10-none-cov\endcsname{90}
\expandafter\def\csname num@rc10-none-risk\endcsname{13.0}
\expandafter\def\csname num@rc10-repair3-cov\endcsname{84}
\expandafter\def\csname num@rc10-repair3-risk\endcsname{3.8}
\expandafter\def\csname num@rc10-sgp-cov\endcsname{79}
\expandafter\def\csname num@rc10-sgp-risk\endcsname{2.9}
\expandafter\def\csname num@rc50-advisorynonumabstain-cov\endcsname{56}
\expandafter\def\csname num@rc50-advisorynonumabstain-risk\endcsname{20.6}
\expandafter\def\csname num@rc50-enforce2-cov\endcsname{51}
\expandafter\def\csname num@rc50-enforce2-risk\endcsname{12.0}
\expandafter\def\csname num@rc50-judgeblock-cov\endcsname{47}
\expandafter\def\csname num@rc50-judgeblock-risk\endcsname{10.0}
\expandafter\def\csname num@rc50-judgecheck-cov\endcsname{90}
\expandafter\def\csname num@rc50-judgecheck-risk\endcsname{42.4}
\expandafter\def\csname num@rc50-none-cov\endcsname{90}
\expandafter\def\csname num@rc50-none-risk\endcsname{43.8}
\expandafter\def\csname num@rc50-repair3-cov\endcsname{73}
\expandafter\def\csname num@rc50-repair3-risk\endcsname{4.2}
\expandafter\def\csname num@rc50-sgp-cov\endcsname{46}
\expandafter\def\csname num@rc50-sgp-risk\endcsname{4.5}
\expandafter\def\csname num@rec-full-clean\endcsname{84.8}
\expandafter\def\csname num@rec-full-seen-correct\endcsname{21.1}
\expandafter\def\csname num@rec-full-seen-wrong\endcsname{9.1}
\expandafter\def\csname num@rec-full-unseen-correct\endcsname{7.1}
\expandafter\def\csname num@rec-full-unseen-wrong\endcsname{11.5}
\expandafter\def\csname num@rec-none-clean\endcsname{80.5}
\expandafter\def\csname num@rec-none-seen-correct\endcsname{43.7}
\expandafter\def\csname num@rec-none-seen-wrong\endcsname{49.3}
\expandafter\def\csname num@rec-none-unseen-correct\endcsname{56.7}
\expandafter\def\csname num@rec-none-unseen-wrong\endcsname{37.8}
\expandafter\def\csname num@rec-rep-seen-wrong\endcsname{14.6}
\expandafter\def\csname num@rec-rep-unseen-correct\endcsname{37.5}
\expandafter\def\csname num@rec-rep-unseen-wrong\endcsname{16.9}
\expandafter\def\csname num@rec-sg-seen-wrong\endcsname{11.2}
\expandafter\def\csname num@rec-sg-unseen-correct\endcsname{7.5}
\expandafter\def\csname num@rec-sg-unseen-wrong\endcsname{11.7}
\expandafter\def\csname num@rp-apis\endcsname{6}
\expandafter\def\csname num@rp-correct\endcsname{49}
\expandafter\def\csname num@rp-correct-fires\endcsname{0}
\expandafter\def\csname num@rp-fires\endcsname{8}
\expandafter\def\csname num@rp-holds\endcsname{38}
\expandafter\def\csname num@rp-one\endcsname{14}
\expandafter\def\csname num@rp-wrong\endcsname{60}
\expandafter\def\csname num@sva-adv-wrong\endcsname{64.4}
\expandafter\def\csname num@sva-judge-clean\endcsname{13.9}
\expandafter\def\csname num@sva-judge-flag\endcsname{82.4}
\expandafter\def\csname num@sva-judge-wrong\endcsname{77.4}
\expandafter\def\csname num@sva-om-clean\endcsname{6.2}
\expandafter\def\csname num@sva-om-fault\endcsname{3.5}
\expandafter\def\csname num@tmx-blindr-clean\endcsname{54.2}
\expandafter\def\csname num@tmx-blindr-p3\endcsname{60.4}
\expandafter\def\csname num@tmx-blindr-p4\endcsname{27.0}
\expandafter\def\csname num@tmx-omcl-clean\endcsname{54.8}
\expandafter\def\csname num@tmx-omcl-p3\endcsname{39.9}
\expandafter\def\csname num@tmx-omcl-p4\endcsname{25.6}
\expandafter\def\csname num@tmx-ours-clean\endcsname{52.5}
\expandafter\def\csname num@tmx-ours-p3\endcsname{47.6}
\expandafter\def\csname num@tmx-ours-p4\endcsname{24.4}
\expandafter\def\csname num@tmx-retrygain-p4\endcsname{1.4}
\expandafter\def\csname num@un-none-correct\endcsname{38.9}
\expandafter\def\csname num@un-none-wrong\endcsname{61.1}
\expandafter\def\csname num@un-sgp-correct\endcsname{19.0}
\expandafter\def\csname num@un-sgp-wrong\endcsname{24.5}
\expandafter\def\csname num@vs-caught\endcsname{26}
\expandafter\def\csname num@vs-n\endcsname{28}
\expandafter\def\csname num@x6-abst\endcsname{18.2}
\expandafter\def\csname num@x6-abst-cap\endcsname{59}
\expandafter\def\csname num@x6-abst-clean\endcsname{83.5}
\expandafter\def\csname num@x6-abst-max\endcsname{59}
\expandafter\def\csname num@x6-abst-min\endcsname{0}
\expandafter\def\csname num@x6-abst1\endcsname{12.1}
\expandafter\def\csname num@x6-abst1-cap\endcsname{23}
\expandafter\def\csname num@x6-abst1-clean\endcsname{83.9}
\expandafter\def\csname num@x6-abst1-max\endcsname{23}
\expandafter\def\csname num@x6-abst1-min\endcsname{5}
\expandafter\def\csname num@x6-adv\endcsname{70.1}
\expandafter\def\csname num@x6-adv-cap\endcsname{87}
\expandafter\def\csname num@x6-adv-clean\endcsname{86.3}
\expandafter\def\csname num@x6-adv-max\endcsname{87}
\expandafter\def\csname num@x6-adv-min\endcsname{19}
\expandafter\def\csname num@x6-adv1\endcsname{63.1}
\expandafter\def\csname num@x6-adv1-cap\endcsname{92}
\expandafter\def\csname num@x6-adv1-clean\endcsname{85.5}
\expandafter\def\csname num@x6-adv1-max\endcsname{91}
\expandafter\def\csname num@x6-adv1-min\endcsname{17}
\expandafter\def\csname num@x6-jap\endcsname{71.7}
\expandafter\def\csname num@x6-jap-cap\endcsname{85}
\expandafter\def\csname num@x6-jap-clean\endcsname{80.7}
\expandafter\def\csname num@x6-jap-max\endcsname{84}
\expandafter\def\csname num@x6-jap-min\endcsname{20}
\expandafter\def\csname num@x6-jb\endcsname{6.1}
\expandafter\def\csname num@x6-jb-cap\endcsname{19}
\expandafter\def\csname num@x6-jb-clean\endcsname{81.7}
\expandafter\def\csname num@x6-jb-max\endcsname{19}
\expandafter\def\csname num@x6-jb-min\endcsname{2}
\expandafter\def\csname num@x6-jbp\endcsname{70.8}
\expandafter\def\csname num@x6-jbp-cap\endcsname{81}
\expandafter\def\csname num@x6-jbp-clean\endcsname{83.7}
\expandafter\def\csname num@x6-jbp-max\endcsname{81}
\expandafter\def\csname num@x6-jbp-min\endcsname{22}
\expandafter\def\csname num@x6-jc\endcsname{71.2}
\expandafter\def\csname num@x6-jc-cap\endcsname{84}
\expandafter\def\csname num@x6-jc-clean\endcsname{86.5}
\expandafter\def\csname num@x6-jc-max\endcsname{84}
\expandafter\def\csname num@x6-jc-min\endcsname{25}
\expandafter\def\csname num@x6-none\endcsname{74.0}
\expandafter\def\csname num@x6-none-cap\endcsname{87}
\expandafter\def\csname num@x6-none-clean\endcsname{84.9}
\expandafter\def\csname num@x6-none-max\endcsname{87}
\expandafter\def\csname num@x6-none-min\endcsname{32}
\expandafter\def\csname num@x6-sgp\endcsname{1.7}
\expandafter\def\csname num@x6-sgp-cap\endcsname{4}
\expandafter\def\csname num@x6-sgp-clean\endcsname{84.9}
\expandafter\def\csname num@x6-sgp-max\endcsname{4}
\expandafter\def\csname num@x6-sgp-min\endcsname{0}

\newcommand{\pct}[1]{\num{#1}\%}
\newcommand{\inputrows}[1]{\csname @@input\endcsname #1 }
\newcommand{\sgp}{SilentGuard}

\title{Invariant-Aware Harness Enforcement to Protect\\LLM Agents from Silent API Failures}
\hypersetup{pdftitle={Invariant-Aware Harness Enforcement to Protect LLM Agents from Silent API Failures},
  pdfauthor={Sajib Acharjee Dip, Brian Ouyang, Lala Motlhabi}}
\author{
  \textbf{Sajib Acharjee Dip}\textsuperscript{1,2},
  \textbf{Brian Ouyang}\textsuperscript{2},
  \textbf{Lala Motlhabi}\textsuperscript{2}
  \\[2pt]
  \textsuperscript{1}Virginia Tech, Blacksburg, VA, USA \\
  \textsuperscript{2}Genentech, South San Francisco, CA, USA
  \\[2pt]
  \small{\textbf{Correspondence:}
    \href{mailto:sajibacharjeedip@vt.edu}{sajibacharjeedip@vt.edu},
    \href{mailto:motlhabi.lala@gene.com}{motlhabi.lala@gene.com}}
  \\[4pt]
  {\normalsize\faGithub~\href{https://github.com/Sajib-006/silentguard}{\textbf{Code}}
    \qquad \raisebox{-0.3ex}{\includegraphics[height=1.2em]{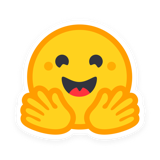}}~\href{https://huggingface.co/datasets/Sajib-006/AgentOrchBench}{\textbf{AgentOrchBench Dataset}}}
}

\begin{document}
\maketitle

\begin{abstract}
LLM agents answer questions by calling APIs and trusting the results. An API can fail silently. It
ignores part of a request, such as a misnamed filter, yet returns HTTP 200 and a plausible but wrong
result. Current fixes, such as checklists, reviewer models and LLM judges, ask the model to catch it.
But models often see the problem and answer anyway. We present \sgp{}, which moves the decision from
the model to the harness running tool calls. It tests each result against invariants that any
correct API must satisfy and withholds values that break one. We find silent failures in
\num{fk-any} of \num{fk-n} public research APIs, with counts inflated up to \num{e3n-maxinfl} times.
Current fixes leave over 70\% of answers wrong, and agents report \pct{sva-judge-wrong} of counts a
judge flags. Across 7 open models, \sgp{} cuts wrong answers on its target faults from
\pct{aob-none-wrong} to \pct{aob-sgp-wrong}, with no loss on fault-free tasks. On three frontier
models they fall from \pct{fr-none-wrong} to \pct{fr-sgp-wrong}. A warning from the same check
leaves \pct{x6-adv} wrong, so enforcement is the key. On new fault types, the harness alone cuts
wrong answers from \pct{un-none-wrong} to \pct{un-sgp-wrong}. It avoids wrong answers by abstaining
rather than recovering.
\end{abstract}

\section{Introduction}

LLM agents now do much of their work through tools. They search literature databases, look up genes
and proteins, query sample catalogs and book travel by calling APIs, and they build their answers
from what those APIs return. The agent's answer can only be as good as the tool results it reads.

Most of the time, a broken tool call is easy to see, because the API returns an error, times out or
sends malformed data. Some failures leave no such trace. Consider an agent that asks a catalog API how
many samples match two conditions. The API's documentation names one filter slightly wrong, so the API
ignores that filter and counts the samples that match the other condition alone. The response is well
formed, the status is 200, and the number is wrong. We call such a response a \emph{silent API
failure}. It looks successful but does not answer the request that was sent. We call its cause a
\emph{silent fault}. Neither is rare. Among
\num{e3n-apis} public research APIs, \num{fk-any} answer at least one plausible mistake this way, by
ignoring a misnamed filter, dropping a filter value with a typo or quietly cutting a page short
(\S\ref{sec:res-realapis}). Earlier work shows that agents miss silent tool errors
\citep{sun2024toolsfail,zhu2026whentoolsfail} and that production APIs fail silently when a
constraint is stated only in prose \citep{li2026silentprobe}. In our faults the response is well
formed and consistent with itself, so no check on the response alone can show the fault. It takes
extra queries to the same API.

The fixes in use today put the model in charge. A checklist prompt asks it to inspect each result
\citep{sun2024toolsfail}. A reviewer model checks each call \citep{ta2026reinforced}. Outcome
Monitors attach a warning to suspect results \citep{panthi2026outcome}, and an LLM judge rates them. In our tests, each of these leaves wrong answers close to the \pct{aob-none-wrong} of an
unguarded agent. The surprising part is why. The problem does not go unseen. A judge that is shown
the results of a consistency check flags \pct{sva-judge-flag} of faulted runs (and \pct{sva-judge-clean} of
clean runs), and a plain warning
from the check tells the agent that the result failed. The agent then reports the flagged count anyway, in
\pct{sva-judge-wrong} of the runs the judge flagged and in \pct{sva-adv-wrong} of the runs that got a
warning. Agents are not blind to the flag, since their stated confidence drops and many mention the
warning. They answer anyway.

Controlled comparisons on 7 open models separate three ingredients (Table~\ref{tab:twobytwo}).
The first is \emph{evidence}, a test whose result shows that the response cannot be right. Without
it, a judge leaves about 70\% of answers wrong, whether its verdict is advice (\pct{x6-jap}) or is
enforced by blocking (\pct{x6-jbp}). The second is \emph{permission}. Agents default to giving a
number, and a short prompt that allows the answer ``cannot be determined'' cuts wrong answers after a
warning from \pct{x6-adv} to \pct{x6-abst}. The cut is uneven
(\num{x6-abst-min}--\num{x6-abst-max}\% wrong by model), and the same prompt without evidence does
nothing. The third is \emph{enforcement}. When the harness (the code that runs the agent's tool
calls) decides what happens to a failed result, the outcome depends far less on the model. A harness
that withholds the value leaves \pct{x6-sgp} wrong, at most \pct{x6-sgp-cap} on any model.

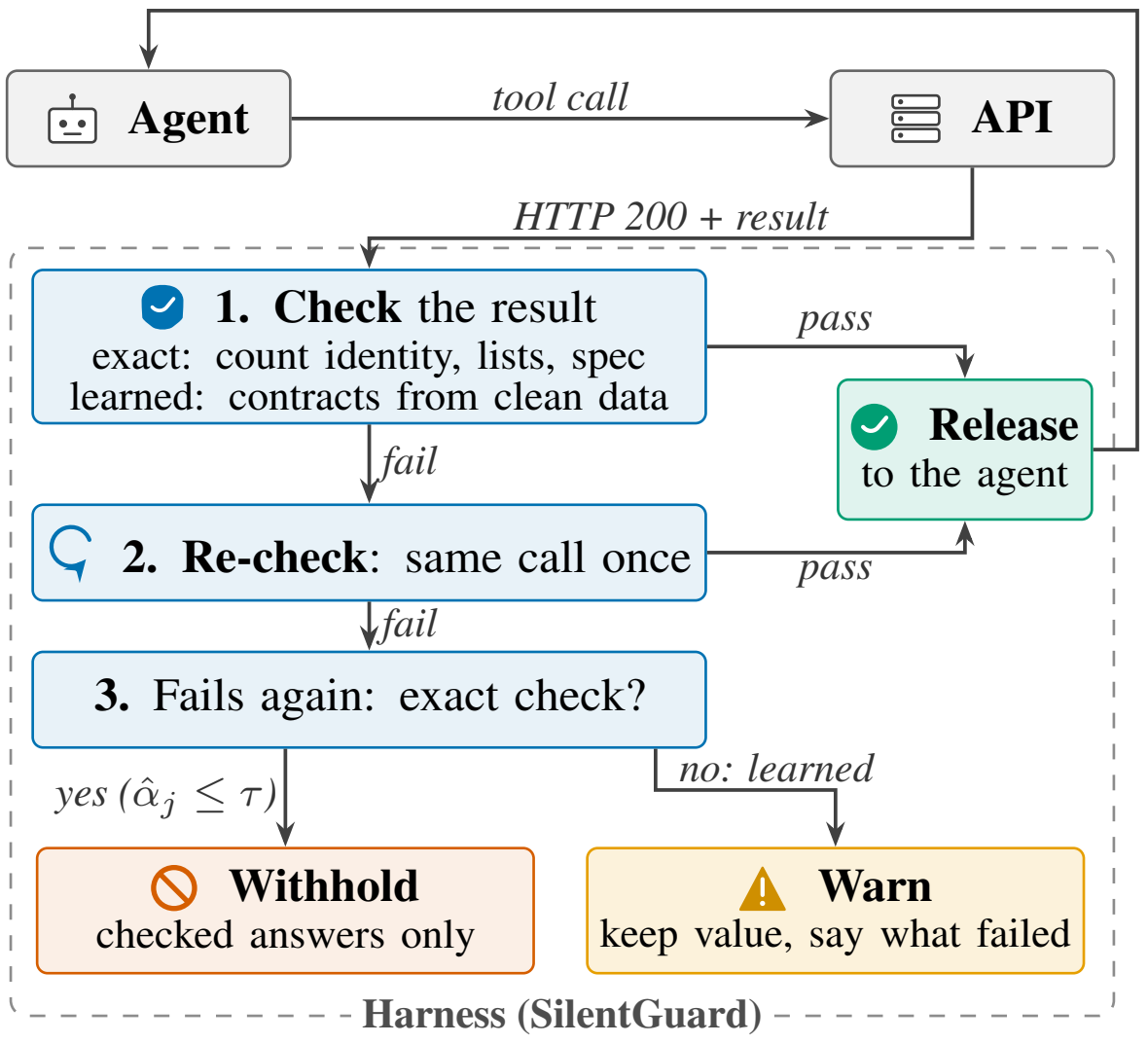
\begin{figure}[t]
\centering
\definecolor{sgblue}{HTML}{0072B2}
\definecolor{sggreen}{HTML}{009E73}
\definecolor{sgred}{HTML}{D55E00}
\definecolor{sgamber}{HTML}{E69F00}
\newcommand{\ico}[1]{\raisebox{-0.22em}{\tikz[x=1em,y=1em,line cap=round,line join=round]{#1}}}
\newcommand{\icoAgent}{\ico{\draw[black!75,line width=0.5pt,fill=white,rounded corners=0.8pt] (0,0) rectangle (1,0.72);
  \fill[black!75] (0.3,0.4) circle (0.085) (0.7,0.4) circle (0.085); \draw[black!75,line width=0.5pt] (0.33,0.18)--(0.67,0.18) (0.5,0.72)--(0.5,0.95);
  \fill[black!75] (0.5,0.98) circle (0.07);}}
\newcommand{\icoAPI}{\ico{\foreach \y in {0,0.36,0.72}{\draw[black!75,line width=0.5pt,fill=white,rounded corners=0.6pt] (0,\y) rectangle (1,\y+0.27);
  \fill[black!75] (0.17,\y+0.135) circle (0.055);}}}
\newcommand{\icoShield}{\ico{\fill[sgblue] (0.5,1.02) -- (0.95,0.86) -- (0.95,0.5) .. controls (0.93,0.22) and (0.7,0.06) .. (0.5,-0.04)
  .. controls (0.3,0.06) and (0.07,0.22) .. (0.05,0.5) -- (0.05,0.86) -- cycle;
  \draw[white,line width=0.7pt] (0.28,0.5)--(0.45,0.33)--(0.74,0.68);}}
\newcommand{\icoRedo}{\ico{\draw[sgblue,line width=0.8pt,-{Stealth[length=1.5mm,width=1.4mm]}] (0.86,0.62) arc (20:320:0.42);}}
\newcommand{\icoOK}{\ico{\fill[sggreen] (0.5,0.48) circle (0.5); \draw[white,line width=0.8pt] (0.26,0.48)--(0.44,0.3)--(0.76,0.68);}}
\newcommand{\icoBan}{\ico{\draw[sgred,line width=1pt] (0.5,0.48) circle (0.44) (0.2,0.78)--(0.8,0.18);}}
\newcommand{\icoWarn}{\ico{\fill[sgamber!90!black,rounded corners=0.5pt] (0.5,1.0)--(1.02,0.02)--(-0.02,0.02)--cycle;
  \draw[white,line width=0.8pt] (0.5,0.7)--(0.5,0.36); \fill[white] (0.5,0.18) circle (0.055);}}
\begin{tikzpicture}[font=\footnotesize, >={Stealth[length=1.7mm]},
  box/.style={draw, rounded corners=2pt, align=center, inner sep=2.5pt, line width=0.5pt},
  ent/.style={box, fill=black!5, draw=black!60, minimum height=6.5mm},
  stp/.style={box, fill=sgblue!9, draw=sgblue, text width=4.4cm, minimum height=6.5mm},
  outc/.style={box, text width=3.2cm, minimum height=8.5mm, anchor=north},
  arr/.style={->, line width=0.7pt, draw=black!75},
  lab/.style={font=\fontsize{8}{9}\selectfont\itshape, text=black!80, inner sep=1.2pt}]
\node[ent, text width=1.75cm] (agent) at (-2.8, 0) {\icoAgent\ \ \textbf{Agent}};
\node[ent, text width=1.75cm] (api) at (2.8, 0) {\icoAPI\ \ \textbf{API}};
\draw[arr] (agent) -- node[lab, above] {tool call} (api);
\node[stp] (check) at (-1.3, -1.55) {\icoShield\ \ \textbf{1. Check} the result\\[-1pt]
  {\fontsize{8}{9}\selectfont exact: count identity, lists, spec}\\[-2pt]{\fontsize{8}{9}\selectfont learned: contracts from clean data}};
\node[stp] (re) at (-1.3, -2.95) {\icoRedo\ \ \textbf{2. Re-check}: same call once};
\node[stp] (gate) at (-1.3, -3.95) {\textbf{3.} Fails again: exact check?};
\node[box, fill=sggreen!12, draw=sggreen, text width=1.55cm, minimum height=9.5mm] (rel) at (2.75, -2.25)
  {\icoOK\ \ \textbf{Release}\\[-1pt]{\fontsize{8}{9}\selectfont to the agent}};
\node[outc, fill=sgred!10, draw=sgred] (wh) at (-1.87, -4.95)
  {\icoBan\ \ \textbf{Withhold}\\[-1pt]{\fontsize{8}{9}\selectfont checked answers only}};
\node[outc, fill=sgamber!14, draw=sgamber] (wa) at (1.87, -4.95)
  {\icoWarn\ \ \textbf{Warn}\\[-1pt]{\fontsize{8}{9}\selectfont keep value, say what failed}};
\coordinate (pad) at ([yshift=-4pt]wh.south);
\begin{scope}[on background layer]
\node[draw=black!45, dashed, rounded corners=3pt, line width=0.5pt, inner sep=4pt,
      fit=(check)(re)(gate)(rel)(wh)(wa)(pad)] (frame) {};
\end{scope}
\node[font=\fontsize{8}{9}\selectfont\bfseries, text=black!70, fill=white, inner sep=2pt] at (frame.south) {Harness (SilentGuard)};
\draw[arr] (api.south) |- ++(0,-0.48) -| node[lab, pos=0.25, above] {HTTP 200 + result} (check.north);
\draw[arr] (check.east) -| node[lab, pos=0.25, above] {pass} (rel.north);
\draw[arr] (check) -- node[lab, right, xshift=1pt] {fail} (re);
\draw[arr] (re.east) -| node[lab, pos=0.25, below] {pass} (rel.south);
\draw[arr] (re) -- node[lab, right, xshift=1pt] {fail} (gate);
\draw[arr] (gate.south -| wh.north) -- node[lab, left] {yes ($\hat\alpha_j\le\tau$)} (wh.north);
\draw[arr] (gate.south east) ++(-0.35,0) -- ++(0,-0.28) -| node[lab, pos=0.25, above, xshift=6pt] {no: learned} (wa.north);
\draw[arr] (rel.east) -- ++(0.3,0) |- ([yshift=4mm]api.north) -| (agent.north);
\end{tikzpicture}
\caption{\textbf{SilentGuard in the harness.} Every tool result is checked, and a result that fails is
re-checked once. If it fails again, the harness withholds it when the failed check is exact and warns
when the check is learned and may itself be wrong (Eq.~\ref{eq:policy}). The model never decides
alone whether a result can be trusted.}
\label{fig:overview}
\end{figure}

\sgp{} (Figure~\ref{fig:overview}) applies these findings as invariant-aware harness enforcement. It produces evidence
with \emph{invariants}, properties that must hold on a correct API. For a filtered count, the count
with a condition plus the count with its negation must equal the count without it. For a list, every
returned record must meet the filter. Every request must use parameters the API documents. Where no
exact invariant exists it uses contracts learned from clean responses. A result that fails is
re-checked once. What the harness does next depends on how certain the failed check is. A broken
exact invariant proves that the API did not answer the request, so the harness withholds the value
and accepts only checked answers. A learned check can itself be wrong, so the harness only warns. No extra model call is needed. Across 7 open models \sgp{} cuts wrong answers from
\pct{aob-none-wrong} to \pct{aob-sgp-wrong} with no loss on fault-free (clean) tasks, and on three
frontier models from \pct{fr-none-wrong} to \pct{fr-sgp-wrong} (\S\ref{sec:res-frontier}). On nine
real APIs, given documentation in which we changed one filter name, wrong answers fall from
\pct{api-none-wrong} to \pct{api-sgp-wrong}. A check against the API's published parameters does as
well alone (\pct{api-schema-wrong}).

\sgp{} prevents wrong answers by abstaining, not by recovering. Under faults it mostly says the count
cannot be determined, and it cannot see a fault that keeps its checks true. So we report risk
against coverage (Figure~\ref{fig:riskcov}). At a 25\% fault rate \sgp{} answers \pct{rc-sgp-cov} of
tasks with \pct{rc-sgp-risk} of its answers wrong, while methods without evidence, or with advice
alone, stay near the \pct{rc-none-risk} risk of no guard. \sgp{} also has limits. Blind retry does
better on faults that clear within one or two calls. The parameter check cannot see a filter written
inside a query string.

\paragraph{Contributions.} (1) A measurement of silent failures in \num{e3n-apis} real APIs, covering
three kinds of fault. (2) Controlled comparisons showing that evidence is necessary, that most agents
do not act on it unless told they may abstain, and that only enforcement by the harness is reliable
across models. (3) \sgp{}, a set of invariant checks enforced by the harness, which withholds results
that fail an exact check and only warns on results that fail a learned check. (4) AgentOrchBench, our benchmark with exact answers, and an evaluation
with predictions written down in advance. The evaluation covers four benchmarks, count and list answers, published
methods and controls, 7 open models, and 3 frontier models added later. Code and tasks are at
\url{https://github.com/Sajib-006/silentguard}.

\section{Related Work}

\paragraph{Tool failures and agent robustness.} Recent benchmarks inject failures into tool
responses. They find that observation handling is where agents break \citep{zheng2026toolrobustbench},
that implicit failures hurt more than explicit errors \citep{zhu2026whentoolsfail}, and that
reliability drops under production-like stress
\citep{gupta2026reliabilitybench,tian2026beyondfunction,wang2026agentnoisebench}. Agents also claim
success they did not achieve \citep{advani2026confidentclosing}, stronger models can trust tool
outputs more \citep{wang2026tooldecides}, and agents adopt corrupted tool returns even when they
question them \citep{yang2026overreliance}. Production APIs often accept queries they do not
understand and still return HTTP 200 \citep{li2026silentprobe}. We study what agents do with such
results and how the harness can catch them.

\paragraph{Fixes for unreliable tools.} Warning the model that tools may fail and giving it a
checklist helps on simple tool errors \citep{sun2024toolsfail}. A second model can review each tool
call before it runs \citep{ta2026reinforced}. Outcome Monitors learn contracts from clean traces
and attach a non-binding warning to results that break them \citep{panthi2026outcome}. Skeptical
prompting and re-querying give mixed gains \citep{zhang2026dontblindlytrust}, and runtime
layers check or wrap tool use \citep{zhao2026agenttether,yan2026trustnotool,mansoor2026verifiedtoolcalls}.
Code outside the model can also decide. Guardrail frameworks \citep{rebedea2023nemo}, secure agent
designs \citep{debenedetti2025camel} and rule-based enforcement \citep{wang2025agentspec,shi2025progent}
control which actions an agent may take. ToolGate commits a tool result only when a hand-written
postcondition holds \citep{liu2026toolgate}, deterministic gates block writes that break a policy
\citep{reddy2026reasonless}, and a final verifier can withhold a whole episode
\citep{zhang2026harnessvalue}. These check actions, hand-written conditions or whole episodes. We
check each read result with exact identities that the harness tests by sending extra queries to the
same API, and we show in controlled tests why the decision must leave the model.

\paragraph{Checks without ground truth.} Our count check is a metamorphic relation
\citep{chen1998metamorphic,segura2016survey}, a two-way form of the query partition used to find bugs
in database engines \citep{rigger2020tlp}. Metamorphic relations have been used to test REST APIs
offline \citep{segura2018restful,khan2026armeta}, and fuzzers such as RESTler
\citep{atlidakis2019restler} look for crashes, not wrong answers. We run the check online, per call.
Learned contracts follow specification and invariant mining \citep{ernst2007daikon,panthi2026outcome}.

\paragraph{How models use evidence.} LLMs often fail to correct their own answers without outside
feedback \citep{huang2024cannot}. When the context contradicts what they believe, they may follow it
or ignore it \citep{xie2024adaptive}. We study the reverse case, in which the outside feedback is
right and most models do not act on it unless told they may abstain. Models guess rather than
abstain, because training and scoring reward a guess \citep{kalai2025hallucinate,kirichenko2025abstentionbench}.
Agents invent values mainly when a tool failure is not signalled \citep{sethi2026fabrication}.
Our faults are never signalled, and a prompt that allows abstaining does nothing without evidence. Saying ``cannot be determined''
is a form of abstention \citep{wen2025abstention} and of selective prediction
\citep{geifman2017selective,kamath2020selective}. Here the harness, not the model, decides when to
abstain.

\section{Methods}
\label{sec:methods}

\paragraph{Silent API failures.} A tool call is a request $q$ and a response $r$. A silent failure is
a response with a success status whose content does not answer $q$. For example, a filter is ignored, a
condition is misread, or a total is capped. Nothing in $r$ marks it.

\paragraph{Checks.} \sgp{} uses two kinds of check, depending on what the tool allows
(App.~\ref{app:impl}).
\emph{Exact checks} hold on any API that applies its filters correctly, so a failure shows that the
request was not answered as sent. For count queries we test, after every filter
query, the count identity
\begin{equation}
N(F) + N(F_{<k} \wedge \neg C_k) \;=\; N(F_{<k}),
\label{eq:partition}
\end{equation}
where $N(F)$ is the count the API returns for filter $F = C_1\wedge\dots\wedge C_k$ and $F_{<k}$ drops
the last condition. The harness issues the two extra queries itself. For every position of the condition list we repeat the test, so an
ignored filter anywhere is caught. When the answer is a list of records, the harness also checks
that every returned record meets the filter, that no record repeats, and that the list size matches
the reported total.
When an API publishes its parameter list, a request with a
parameter outside it is rejected before it runs (the parameter check, or \emph{spec check}). \emph{Learned checks} come
from clean traces and cover field presence and type, echoed arguments, positive values, categorical values,
linear relations between fields and date order, mined as in \citet{panthi2026outcome}.
Table~\ref{tab:checks} lists the checks by kind of tool operation, including ones we did not test.

\begin{table}[t]
\centering\footnotesize
\setlength{\tabcolsep}{2.6pt}
\begin{tabular}{@{}>{\raggedright\arraybackslash}p{1.6cm}>{\raggedright\arraybackslash}p{2.15cm}>{\raggedright\arraybackslash}p{2.5cm}c@{}}
\toprule
\rowcolor{head}\textbf{Operation} & \textbf{Silent failure} & \textbf{Check that must hold} & \textbf{Tested}\\
\midrule
\multicolumn{4}{@{}l}{\textit{Exact checks}}\\
Count, filtered & filter ignored, joined with OR, capped & count identity at every position (Eq.~\ref{eq:partition}) & \checkmark\\
\addlinespace[1.5pt]
List, filtered & records outside the filter, repeats & each record meets the filter, no repeats, size = total & \checkmark\\
\addlinespace[1.5pt]
Documented API & unknown parameter accepted & parameter is in the published spec & \checkmark\\
\addlinespace[1.5pt]
One filter & filter ignored & value swap (a different value changes the result) & \checkmark\\
\addlinespace[1.5pt]
Pages, ranges, sorting, writes & records skipped, or bound, order or update ignored & pages do not overlap, results in bound and in order, a read returns the write & \\
\midrule
\multicolumn{4}{@{}l}{\textit{Learned checks}}\\
Any tool & wrong type, echo, range, relation & contracts mined from clean responses & \checkmark\\
\bottomrule
\end{tabular}
\caption{Checks by kind of tool operation. Exact checks are invariants that hold on every correct API. Learned checks are
mined from clean responses and can raise false alarms. Of
\num{chk-n} documented operations in \num{chk-apis} public APIs, \num{chk-exact} allow an exact check
and all allow the spec check (Table~\ref{tab:app-checkability}).}
\label{tab:checks}
\end{table}

\paragraph{Detection and action.} \sgp{} separates finding a problem from acting on it. Each check
$j$ maps a tool result $r$ to pass or fail and has an estimated false-alarm rate
$\hat\alpha_j = 1/(n_j+2)$, where $n_j$ is the number of clean examples it held on (rule of
succession). Exact checks have $n_j=\infty$, so $\hat\alpha_j=0$. After one re-check, the harness action is
\begin{equation}
\pi(r)=\begin{cases}
\text{release} & \text{all checks pass,}\\
\text{withhold} & \hat\alpha_j \le \tau \text{ for a failed } j,\\
\text{warn} & \text{otherwise,}
\end{cases}
\label{eq:policy}
\end{equation}
with cutoff $\tau=1\%$. New checks plug into $\pi$ without changing it.

\paragraph{\sgp{}.} For every tool result the harness runs the checks. (1) A result that passes is
released unchanged. (2) A result that fails is re-checked once. The harness sends the same call
again and releases the new result only if it passes. (3) If the result fails again, the action
depends on the failed check (Eq.~\ref{eq:policy}). If a failed check is exact
($\hat\alpha_j\le\tau$), the failed values are
withheld and the agent is told which values could not be verified and which other tool could redo
the step. On count answers this becomes an answer rule. The harness accepts only a count that
passed a check, rejects any other count twice, and then answers that the count cannot be
determined. If a failed check is learned from few examples, the data stays visible with a warning
naming the failed property and the recovery tools, as in \citet{panthi2026outcome}. (4) Results that
already carry an error pass through unchanged. A result checklist \citep{sun2024toolsfail} is added
to the system prompt. \sgp{} never repairs a value on its own. Recovery is always the agent's own
call with a listed tool (\S\ref{sec:res-repair} shows why).

\section{Experimental Setup}
\label{sec:setup}

\paragraph{AgentOrchBench (ours).} We built AgentOrchBench to measure silent failures where the
right answer is known exactly. The agent works from a shell and calls eight local REST services with
\texttt{curl}, and all services are reset before each task. Each task asks how many of the 400 catalog
records match two or three conditions, and the answer is a single number computed from the seed data,
so grading is exact and needs no judge. The catalog returns only the count, so the agent cannot check
a filter by reading the records. Faults are injected inside the server. The reply is HTTP 200, and
the fault repeats on every call, so asking again returns the same wrong number. Each fault type is
modeled on a kind of API bug (Table~\ref{tab:faults}, App.~\ref{app:aob}). One of them, an ignored
parameter, is the failure we measure in real public APIs (\S\ref{sec:res-realapis}). We ran the six
fault types in the top part of Table~\ref{tab:faults}. The two below them keep the count identity true
by design and are used in the list tasks and the harness-only tests (App.~\ref{app:lists}
and~\ref{app:budget}).
A retail orders API with four fault types tests whether results carry over to a
different schema. We ran 36 tasks with 2 seeds under each fault type and with no fault, for every
model and method variant (arm).

\paragraph{Other benchmarks.} The \emph{real public APIs} are 9 APIs (NCBI, UniProt, Europe PMC, GBIF,
Zenodo, ChEMBL, arXiv, DataCite, iNaturalist) with 5 count tasks each and correct or slightly wrong
documentation. That gives 45 runs per model, arm and documentation version.
\emph{ToolMaze} \citep{zhu2026whentoolsfail} has 400 multi-step tool workflows with silent faults that
hit the first call (P3) or every call (P4), explicit errors (P1, P2) and clean runs (P0), scored by
its own judge (App.~\ref{app:toolmaze}). We used 180 workflows never used in method design.

\paragraph{Methods compared.} The published methods are a checklist prompt \citep{sun2024toolsfail},
a reviewer \citep{ta2026reinforced} and Outcome Monitors (OM) \citep{panthi2026outcome}. The reviewer
is the agent's own model, which checks each call before it runs for up to five loops. OM uses its
contract miner and warning text, with rules learned from clean responses of other tasks. On ToolMaze
we also run the checklist plus OM. The baselines and controls are no guard, a caution prompt, an LLM
judge in the sense of \citet{zheng2023judging}, self-verification, blind retry (every call sent
twice, the second response released) and re-query (both responses shown). The judge is a fixed
model (gpt-oss-120b) that also sees the check results, meaning the harness's consistency queries. The
ablations of the harness action are warning, warning with a recovery tip, answer gate, Enforce
(checked counts only, failed count left visible) and auto-repair (the harness recounts by another
route). Prompts are in App.~\ref{app:baselines}.

\paragraph{Models.} The main runs use 7 open-weight (open) models (Phi-4, gpt-oss-120b, Gemma-3-27B,
Llama-3.3-70B, Llama-4-Scout, Llama-4-Maverick and Nemotron-3-Ultra). Three
frontier models (GPT-5.6 Sol, Terra and Luna) ran six arms of AgentOrchBench and two arms of the
real-API study (\S\ref{sec:res-frontier}). The frontier AgentOrchBench runs and the Nemotron-3-Ultra control arms ran on
Genentech's internal platform with the same code, tasks, seeds and grader. Company policy bars
releasing their logs, so we report their answer counts. All models use at most 2,000 output tokens per turn, at
most 16 turns (25 on ToolMaze) and the same prompts. The open models run at temperature 0, and the
frontier models use the provider default (App.~\ref{app:models}).

\paragraph{Protocol.} Every planned experiment was written down with its predictions before it
ran. The code was frozen by hash, and runs refuse to start if the code changed. The method went
through three versions on ToolMaze. Each version was tested only on unseen workflows,
and the final \sgp{} was frozen before all planned runs (App.~\ref{app:history}).

\paragraph{Grading and statistics.} We grade each answer as \emph{correct} (it equals the ground
truth), \emph{unknown} (it gives no count) or \emph{wrong} (any other count). Confidence values and
identifiers are not read as counts. ToolMaze uses its own judge. We report the mean over models with
its standard error (SE, $\pm$) and 95\% task-cluster bootstrap intervals for the key differences. We
also report how many of the 7 models show the difference, with an exact sign-flip test (smallest
$p$ 0.016, App.~\ref{app:models}).

\section{Results}
\label{sec:results}

\subsection{Silent Failures Are Common in Real APIs}
\label{sec:res-realapis}

For each API we sent an unfiltered query, the documented filtered query, and two or three plausible
wrong filter names (for example \texttt{organism=human} on NCBI instead of \texttt{human[orgn]} in the
search term). Of \num{e3n-apis} APIs, \num{e3n-silent} return HTTP 200 with exactly the unfiltered
count and no warning for at least one wrong name, up to \num{e3n-maxinfl}$\times$ too large
(all probes in Table~\ref{tab:realapis} and Figure~\ref{fig:inflation}, App.~\ref{app:realapis}). The
other APIs reject unknown parameters with an error. The value-swap check, which flags a filter whose count does not change with its
value, caught \num{vs-caught} of the \num{vs-n} ignored filter names. Misnamed filters are not the only silent fault. When asked for
more records per page than they allow, \num{fk-cap} of the \num{fk-reach} APIs we could test return
HTTP 200 with fewer records and no error. Three APIs silently drop a filter whose value has
a typo, such as a bad year or date. The spec check cannot see such a typo, because the parameter name
is valid. The count identity and the value-swap check flagged it in \num{bv-fired} of \num{bv-n}
probes, with no false alarm (Table~\ref{tab:app-badvalue}). In all, \num{fk-any} of \num{fk-n} APIs
show at least one silent fault (Table~\ref{tab:app-faultkinds}).

\subsection{Evidence Is Necessary and Enforcement Makes It Reliable}
\label{sec:res-main}

\begin{table}[t]
\centering
\small
\setlength{\tabcolsep}{2.5pt}
\begin{tabular}{@{}lcc@{}}
\toprule
\rowcolor{head}
\textbf{(a) One LLM judge} & \textbf{Advises} & \textbf{Blocks}\\
\midrule
Without the check's results & 71.7 & 70.8\\
With the check's results & 71.2 & \cellcolor{ours}\textbf{6.1}\\
\midrule
\rowcolor{head}\textbf{(b) Acting on the check} & \textbf{Wrong$\downarrow$} & \textbf{Clean$\uparrow$}\\
\midrule
No guard (no check) & 74.0 (32--87) & 84.9\\
Warning & 70.1 (19--87) & 86.3\\
Warning + may abstain & 18.2 (0--59) & 83.5\\
\rowcolor{ours}SilentGuard (withholds) & \textbf{1.7} (0.0--3.7) & 84.9\\
\bottomrule
\end{tabular}
\caption{\textbf{Evidence is necessary, and enforcement makes it reliable.} \% wrong answers under silent faults on AgentOrchBench. The same 7 open models and 2 seeds in every cell. (a) One judge (gpt-oss-120b) in four settings. It does or does not see the results of the count check's extra queries, and its verdict is advice to the agent (advises) or is enforced by the harness (blocks). (b) Three ways to act on the full count check. May abstain: the prompt allows the answer that the count cannot be determined. In brackets: lowest and highest model. Clean: \% correct with no fault. All methods: Table~\ref{tab:main}.}
\label{tab:twobytwo}
\end{table}

Table~\ref{tab:twobytwo} separates the ingredients on all 7 open models. With no guard, agents report
a wrong count in \pct{x6-none} of faulted runs. Methods that rely on the model's own reading of the
result change little (checklist \pct{aob-checklist-wrong}, reviewer \pct{aob-reviewer-wrong}, Outcome
Monitors \pct{aob-om-wrong}, Table~\ref{tab:main}).

\paragraph{Evidence is necessary.} Part (a) tests one judge in four settings, with or without the
check's results and advising or blocking. Without the count check's extra queries,
the judge's verdict leaves \pct{x6-jap} wrong as advice and \pct{x6-jbp} when the harness blocks on it. Outcome
Monitors with blocking leave \pct{aob-omblock-wrong}, because their contracts rarely fire on a wrong
count (\pct{sva-om-fault} of faulted runs, \pct{sva-om-clean} of clean ones). A silently wrong count
gives a detector nothing to see.

\paragraph{Evidence is not enough.} With the check's queries, the judge flags a result as suspect in
\pct{sva-judge-flag} of faulted runs (and \pct{sva-judge-clean} of clean runs), yet as advice it still
leaves \pct{x6-jc} wrong. In
\pct{sva-judge-wrong} of flagged runs the agent reports the flagged count anyway. After a warning
from an earlier version of the check, it does so in \pct{sva-adv-wrong} of warned runs.
The agents do notice the flag. After a warning their
stated confidence falls from \num{ig-clean-conf} on clean runs to \num{ig-adv-conf}, and
\num{ig-adv-mention}\% of the wrong answers mention the warning.
Still, agents default to giving a number. A short prompt that permits the answer ``cannot be
determined'' cuts wrong answers after a warning from \pct{x6-adv} to \pct{x6-abst} (part b), so agents
can act on evidence. But wrong answers still range from \num{x6-abst-min}\% to \num{x6-abst-max}\%
across models, and the same permission without evidence changes nothing (caution prompt,
\pct{aob-skeptic-wrong}).

\paragraph{Enforcement makes it reliable.} When the harness acts on the evidence, the result depends
far less on the model and the prompt. The judge with evidence and blocking leaves \pct{x6-jb} wrong.
\sgp{} withholds the failed value and accepts only checked counts. It leaves \pct{x6-sgp} wrong (at
most \pct{x6-sgp-cap} on any model), needs no extra model call and keeps clean accuracy
(\pct{x6-sgp-clean}, against \pct{x6-none-clean} with no guard). \sgp{} has fewer wrong answers than
no guard, each published method and blind retry on 7 of 7 models (Tables~\ref{tab:main},
\ref{tab:app-permodel} and~\ref{tab:permodel}, with $p=\num{pm-aob-p}$ against no guard, the smallest
possible with 7 models). How the harness acts matters. Enforce, which accepts only
checked counts but leaves the failed count visible, gives \num{enf-min}\% to \num{enf-max}\% wrong by
model, because agents re-query until some other query passes the check (App.~\ref{app:full}).

\begin{table}[t]
\centering
\footnotesize
\setlength{\tabcolsep}{2.6pt}
\begin{tabular}{@{}lcccc@{}}
\toprule
\rowcolor{head}
\textbf{Method} & Wrong$\downarrow$ & Corr.$\uparrow$ & Unk. & Clean$\uparrow$\\
\midrule
\multicolumn{5}{@{}l}{\textit{Baselines and controls}}\\
No guard & 74.0\se{7.1} & 16.4 & 9.6 & 84.9\se{10.6}\\
Caution prompt & 73.3\se{7.3} & 16.4 & 10.3 & 86.1\se{9.1}\\
Self-verification & 56.2\se{9.2} & 14.2 & 29.6 & 76.6\se{12.8}\\
Re-query & 74.9\se{5.0} & 17.0 & 8.1 & 84.7\se{8.6}\\
Blind retry & 74.4\se{5.1} & \underline{17.3} & 8.3 & \textbf{88.3}\se{7.4}\\
LLM judge & 71.2\se{7.8} & \textbf{17.4} & 11.4 & \underline{86.5}\se{8.5}\\
\midrule
\multicolumn{5}{@{}l}{\textit{Published methods}}\\
Checklist & 72.3\se{8.1} & 16.2 & 11.5 & 84.1\se{10.6}\\
Reviewer & 73.8\se{5.7} & 15.8 & 10.4 & 80.4\se{10.0}\\
Outcome Monitors & 72.2\se{8.1} & 16.6 & 11.2 & 86.1\se{8.8}\\
\midrule
\multicolumn{5}{@{}l}{\textit{Evidence, permission, enforcement}}\\
Judge, no evidence & 71.7\se{8.7} & 16.3 & 11.9 & 80.7\se{12.3}\\
\quad + blocking & 70.8\se{8.2} & 16.7 & 12.5 & 83.7\se{10.6}\\
Judge + blocking & \underline{6.1}\se{2.2} & 3.5 & 90.4 & 81.7\se{9.3}\\
OM + blocking & 70.9\se{8.5} & 16.6 & 12.5 & \underline{86.5}\se{9.4}\\
Warning + may abstain & 18.2\se{9.3} & 5.0 & 76.8 & 83.5\se{11.2}\\
\midrule
\rowcolor{ours}\textbf{SilentGuard (ours)} & \textbf{1.7}\se{0.5} & 2.4 & 95.8 & 84.9\se{11.3}\\
\bottomrule
\end{tabular}
\caption{\textbf{SilentGuard gives the fewest wrong answers on AgentOrchBench.} Mean over 7 open models ($\pm$ standard error), \% of runs with a silent fault (Wrong, Corr., Unk.)\ or with no fault (Clean). Unk.: the answer says the count cannot be determined. Bold is best and underlined is second. The LLM judge sees the check results and advises. Published: checklist \citep{sun2024toolsfail}, reviewer \citep{ta2026reinforced}, Outcome Monitors (OM) \citep{panthi2026outcome}. Blocking: a flagged result is withheld and only accepted counts may be reported. No evidence: the judge does not see the check results.}
\label{tab:main}
\end{table}

\begin{figure}[t]
\centering
\includegraphics[width=\columnwidth]{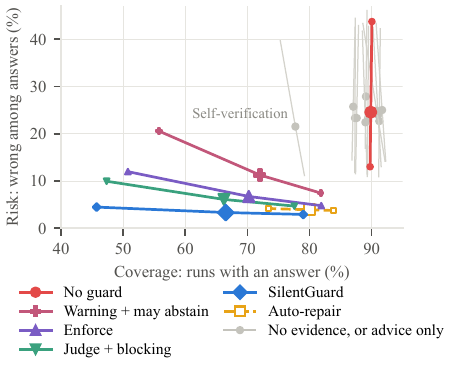}
\caption{\textbf{Risk against coverage} on AgentOrchBench with clean and faulted calls mixed. Lower is
safer and further right answers more. Each line runs from a 10\% fault rate (right end) to 50\% (left
end, higher risk), with a large marker at 25\%. Mean over 7 open models. Gray lines are methods
without evidence of the fault, or with advice alone.}
\label{fig:riskcov}
\end{figure}

\paragraph{Abstaining, not recovering.} Under faults \sgp{} mostly answers that the count cannot be
determined, and its correct answers fall from \pct{aob-none-correct} to \pct{aob-sgp-correct}. The
\pct{aob-sgp-wrong} applies to the six fault types the count check can see. On six other fault types,
written after the method was frozen, wrong answers fall only from \pct{un-none-wrong} to
\pct{un-sgp-wrong}, and correct answers fall from \pct{un-none-correct} to \pct{un-sgp-correct}
(harness alone, Table~\ref{tab:repair}). So we also compare methods as selective prediction
does \citep{geifman2017selective}, by risk (the share of answers that are wrong) against coverage
(the share of runs that get an answer). Figure~\ref{fig:riskcov} mixes each method's measured clean
and faulted outcomes at fault rates from 10\% to 50\%. Methods without evidence, or with advice
alone, sit with no guard. At a 25\% fault rate the LLM judge answers \pct{rc-judgecheck-cov} of tasks
at \pct{rc-judgecheck-risk} risk, and no guard answers \pct{rc-none-cov} at \pct{rc-none-risk}. The
harness arms sit far below. \sgp{} has the lowest risk at fault rates up to 25\% (\pct{rc-sgp-risk} at
25\%). At 50\% auto-repair is slightly lower (\pct{rc50-repair3-risk} against \pct{rc50-sgp-risk}).
\sgp{} is also among the lowest in coverage (\pct{rc-sgp-cov}, against \pct{rc-judgeblock-cov} for
judge + blocking). Auto-repair is the other choice. On the fault types it was built for it
answers \pct{rc-repair3-cov} at \pct{rc-repair3-risk} risk, but on new fault types it releases wrong
answers marked as verified (\S\ref{sec:res-repair}, with the cost trade-off in Table~\ref{tab:app-risk}).

\begin{table}[t]
\centering
\footnotesize
\setlength{\tabcolsep}{3pt}
\begin{tabular}{@{}lccc@{}}
\toprule
\rowcolor{head}\textbf{Retail domain} & Wrong$\downarrow$ & Correct$\uparrow$ & Clean$\uparrow$\\
\midrule
No guard & 78.6\se{9.9} & \underline{7.2}\se{2.4} & \underline{36.2}\se{11.5}\\
Auto-repair & 15.0\se{4.0} & \textbf{30.5}\se{10.0} & 35.7\se{11.8}\\
Enforce & \underline{12.3}\se{3.8} & 6.5\se{2.3} & 33.8\se{11.5}\\
\rowcolor{ours}\textbf{SilentGuard (ours)} & \textbf{11.0}\se{2.8} & 7.1\se{2.0} & \textbf{38.5}\se{11.5}\\
\midrule
\rowcolor{head}\textbf{Real APIs (9)} & Wrong$\downarrow$ & Correct$\uparrow$ & Right docs$\uparrow$\\
\midrule
No guard & 89.8\se{5.3} & 3.5\se{2.8} & 92.1\se{6.5}\\
Blind retry & 89.5\se{5.8} & 3.2\se{2.5} & \textbf{93.0}\se{5.5}\\
Spec check alone & \textbf{21.0}\se{4.6} & \textbf{70.8}\se{6.8} & 92.1\se{6.5}\\
Typed calls & 28.6\se{4.6} & 66.3\se{5.6} & 90.8\se{5.4}\\
Typed calls + ours & 28.3\se{5.0} & 65.4\se{5.5} & 92.1\se{4.7}\\
\rowcolor{ours}\textbf{SilentGuard (ours)} & \underline{21.3}\se{3.8} & \underline{69.5}\se{7.5} & \underline{92.4}\se{7.3}\\
\midrule
\rowcolor{head}\textbf{ToolMaze} & Silent$\uparrow$ & Loud$\uparrow$ & Clean$\uparrow$\\
\midrule
Checklist & 23.2\se{3.3} & 38.3\se{4.2} & \underline{55.5}\se{4.1}\\
Outcome Monitors & 30.5\se{1.3} & \textbf{40.7}\se{2.2} & \textbf{57.1}\se{3.6}\\
Checklist + OM & 32.7\se{2.2} & \underline{39.8}\se{2.6} & 54.8\se{4.1}\\
Blind retry & \textbf{43.7}\se{1.7} & -- & 54.2\se{2.6}\\
\rowcolor{ours}\textbf{SilentGuard (ours)} & \underline{36.0}\se{2.5} & \underline{39.8}\se{2.9} & 52.5\se{4.2}\\
\bottomrule
\end{tabular}
\caption{\textbf{The other benchmarks}, each with the methods run on it (mean over 7 open models, $\pm$ standard error, \%). Retail: our count benchmark on a second schema. Real APIs: agents given slightly wrong documentation. Right docs: \% correct with the right documentation. Typed: tool calls with a strict parameter schema, as in function calling. ToolMaze \citep{zhu2026whentoolsfail}: \% task success under silent faults, loud (explicit) errors and no fault. OM: Outcome Monitors. Blind retry was not run on loud errors. Bold: best. Underlined: second.}
\label{tab:other}
\end{table}

\paragraph{Retail and real APIs.} Wrong answers also fall on the retail domain (Table~\ref{tab:other}),
from \pct{ret-none-wrong} with no guard to \pct{ret-sgp-wrong} with \sgp{}. On real public APIs,
agents given slightly wrong documentation report a wrong count \pct{api-none-wrong} of the time with
no guard and \pct{api-sgp-wrong} with \sgp{}, and correct answers rise from \pct{api-none-correct} to
\pct{api-sgp-correct}. With the right documentation an unguarded agent is correct
\pct{api-none-cleancorrect} of the time. On these tasks the gain comes from the spec check, not from
the count check. The spec check alone does as well (\pct{api-schema-wrong} wrong), because it rejects
the misnamed filter. Typed tool calls with a strict schema,
as in function calling, do worse (\pct{api-typed-wrong}). The gain is uneven (\pct{api-first4-sgp-wrong} wrong on the four
APIs chosen before freezing, \pct{api-later5-sgp-wrong} on the five added later). Most of the errors left are on APIs whose filters live
inside a query string (NCBI, UniProt, Europe PMC, arXiv), which neither a schema nor the value-swap
check can see. When the documentation instead gives a wrong value format for a valid parameter
(iNaturalist dates), the spec check cannot help (\pct{bf-schema-wrong} wrong, against
\pct{bf-none-wrong} with no guard). The value-swap check cuts wrong answers to \pct{bf-sgp-wrong},
mostly by abstaining (Table~\ref{tab:app-badformat}).

\paragraph{A benchmark built by others.} On ToolMaze, scored by its own judge, only learned checks
exist, so \sgp{} warns and never withholds. So this benchmark tests the re-check, not
enforcement. Under silent faults \sgp{}'s task success is slightly above the published methods
(\pct{tm-ours-silent}, against \pct{tm-omcl-silent} for checklist + OM, on
\num{pm-tm-wins} of \num{pm-tm-n} models, 95\% interval [\num{boot-tm-lo}, \num{boot-tm-hi}] points).
The lead comes from one-time faults (not lasting ones), and clean success is
\num{gain-toolmaze-clean} points below OM.

\paragraph{List answers.} When the answer is a list of record IDs, the same rule gives the fewest
wrong lists (from \pct{rec-none-unseen-wrong} to \pct{rec-full-unseen-wrong} on unseen faults),
and correct lists fall too (\pct{rec-none-unseen-correct} to \pct{rec-full-unseen-correct},
App.~\ref{app:lists}).

\subsection{Frontier Models Fall for Silent Faults Too}
\label{sec:res-frontier}

\begin{table}[t]
\centering
\footnotesize
\setlength{\tabcolsep}{2.6pt}
\begin{tabular}{@{}lcccccc@{}}
\toprule
\rowcolor{head}
\textbf{Model} & None & Self-v. & Warn & W+A & Enf. & \textbf{Ours}\\
\midrule
\multicolumn{7}{@{}l}{\textit{Wrong answers under silent faults (\%, $\downarrow$)}}\\
Open models & 74.0 & 56.2 & 70.1 & 18.2 & 8.8 & \cellcolor{ours}\textbf{1.7}\\
GPT-5.6 Sol & 89.4 & 2.5 & 11.3 & 2.3 & 31.5 & \cellcolor{ours}\textbf{0.0}\\
GPT-5.6 Terra & 62.3 & 3.9 & 53.9 & 35.9 & 34.0 & \cellcolor{ours}\textbf{0.2}\\
GPT-5.6 Luna & 59.0 & 6.9 & 63.2 & 45.1 & 36.1 & \cellcolor{ours}\textbf{1.4}\\
\midrule
\multicolumn{7}{@{}l}{\textit{Correct on clean tasks (\%, $\uparrow$)}}\\
Open models & 84.9 & 76.6 & \textbf{86.3} & 83.5 & \textbf{86.3} & \cellcolor{ours}84.9\\
GPT-5.6 Sol & 98.6 & 29.2 & 77.8 & 86.1 & \textbf{100.0} & \cellcolor{ours}\textbf{100.0}\\
GPT-5.6 Terra & 93.1 & 15.3 & 68.1 & 76.4 & 90.3 & \cellcolor{ours}\textbf{98.6}\\
GPT-5.6 Luna & 84.7 & 16.7 & 69.4 & 73.6 & 87.5 & \cellcolor{ours}\textbf{95.8}\\
\bottomrule
\end{tabular}
\caption{\textbf{Frontier models fall for silent faults too. SilentGuard cuts wrong answers the most and keeps clean accuracy.} AgentOrchBench, six methods on three frontier models (2 seeds) and the mean of the 7 open models. Self-v.: self-verification, the model runs the count check itself. Warn: the harness warns. W+A: warning plus permission to abstain. Enf.: Enforce. None: no guard. Ours: SilentGuard. Bold: best in the row. Means are computed before rounding.}
\label{tab:frontier}
\end{table}

Table~\ref{tab:frontier} adds three frontier models on six arms. With no guard they report a wrong
count in \num{fr-none-wrong-min}--\num{fr-none-wrong-max}\% of faulted runs, and GPT-5.6
Sol is the highest. Told to run the check themselves, they almost stop giving wrong answers
(\pct{fr-selfverify-wrong}), but their clean accuracy falls from \pct{fr-none-clean} to
\pct{fr-selfverify-clean}, which suggests that they mostly refuse. A warning from the harness is enough for Sol
(\pct{fr-sol-warning-wrong} wrong, and \pct{fr-sol-permission-wrong} with permission to abstain) and
not for Terra or Luna (\pct{fr-terra-permission-wrong} and \pct{fr-luna-permission-wrong} even with
permission). Both cost clean accuracy (\pct{fr-warning-clean} with the warning and
\pct{fr-permission-clean} with permission, against \pct{fr-none-clean} with no guard). Enforce does worse
here than on the open models (\pct{fr-enforce-wrong} against \pct{open7-enforce-wrong}). Without logs we cannot test why. On the open models, about half of Enforce's wrong
answers come from re-querying (App.~\ref{app:full}). \sgp{} leaves
\num{fr-sgp-wrong-min}--\num{fr-sgp-wrong-max}\% wrong, with clean accuracy at \pct{fr-sgp-clean}. On
the real APIs it cuts the frontier models' wrong answers from \pct{frapi-none-wrong} to
\pct{frapi-sgp-wrong} (Table~\ref{tab:app-realapi-models}). Silent faults are not only a small-model
problem.

\paragraph{Blind retry.}\label{sec:res-retry} Sending every call twice helps only when the fault
clears. On ToolMaze's one-time faults blind retry reaches \pct{tmx-blindr-p3} task success against
\pct{tmx-ours-p3} for \sgp{}, against our prediction (Table~\ref{tab:retry}). On AgentOrchBench it
leaves \pct{aob-retry2-wrong} wrong and on real APIs \pct{api-retry2-wrong}, and it repeats a
booking, payment or message \num{ra-blindr-perrun} times per run (\num{ra-ours-perrun} for \sgp{}).

\paragraph{Learned checks.}\label{sec:res-precision} Learned contracts raise false alarms that
repeat on re-check, so we warn on them. On 80 fresh ToolMaze workflows, withholding on learned checks
gave \pct{learned-withhold} task success against \pct{learned-warn} for warning, a small difference.
No learned check reaches the cutoff $\tau$, so in our runs the rule comes down to withholding on exact
checks and warning on learned ones (Tables~\ref{tab:ablation} and~\ref{tab:app-gate}).

\subsection{Auto-Repair Returns Wrong Answers Marked as Verified}
\label{sec:res-repair}

\begin{table}[t]
\centering
\small
\setlength{\tabcolsep}{3pt}
\resizebox{\columnwidth}{!}{%
\begin{tabular}{@{}lcccc@{}}
\toprule
\rowcolor{head}
\textbf{Unseen fault} & \textbf{Fired} & \textbf{No guard} & \textbf{SilentGuard} & \textbf{Auto-repair}\\
\midrule
Word-stem match & 0/36 & 32/4/0 & 32/4/0 & 32/4/0\\
NOT drops empty & 0/36 & 0/36/0 & 0/36/0 & 0/36/0\\
Partial shard & 36/36 & 20/16/0 & 0/0/36 & 34/2/0\\
Tag-join double & 15/36 & 36/0/0 & 21/0/15 & 36/0/0\\
Cursor skips ties & 36/36 & 9/27/0 & 0/0/36 & 0/36/0\\
Lossy index & 35/36 & 35/1/0 & 0/1/35 & 0/1/35\\
\midrule
Total & & 132/84/0 & \textbf{53/41/122} & 102/79/35\\
\bottomrule
\end{tabular}}
\caption{\textbf{Auto-repair returns wrong answers marked as verified.} Six fault types written after the method was frozen, run through the harness alone (36 count tasks each, cells: wrong / correct / unknown). Fired: tasks where the count check fired. When its check fires, SilentGuard (bold) never releases a wrong answer.}
\label{tab:repair}
\end{table}

On the fault types it was designed for, auto-repair recovers many true counts
(\pct{aob-repair3-correct} correct, Table~\ref{tab:ablation}). To test it on faults it was not
designed for, we wrote six new fault types after the method was frozen and ran the harness alone,
with no model (Table~\ref{tab:repair}). Where its check fired, the \sgp{} answer rule never released
a wrong count. Its \num{ob-sgp-wrong} wrong answers all come from faults the check cannot see, which
every method misses. Auto-repair returned more correct counts (\num{ob-rep-correct} against
\num{ob-sgp-correct}) but also \num{ob-rep-wrong} wrong ones, and \num{ob-vbw} of its repairs were
wrong while marked as verified. A user has no way to tell them apart from correct answers.

\paragraph{Cost and fine-tuning.}\label{sec:res-cost}
\sgp{} makes no extra model call. Per faulted run it takes less time than the LLM judge
(\num{cost-sgp-secs}\,s against \num{cost-judge-secs}\,s) but more tokens. A small model fine-tuned
to refuse also refuses on unseen fault types, but only when it is shown the harness check
(App.~\ref{app:cost}).

\section{Conclusion}

Silent API failures are common in real services and fool every agent we tested. The fixes in use
today leave the decision to the model, which by default reports the number it was given, even when it
is told the number is suspect. Evidence of the fault is necessary, and permission to abstain lets some
models act on it. Only enforcement by the harness works for every model. Harness builders should give the harness the final say over
results that fail an exact check, and repair only fault types the check can confirm. API builders
should reject unknown parameters and say when a page was cut. Benchmarks should include faults
that repeat.

\section*{Limitations}

\paragraph{Checks have blind spots.} A fault that is consistent with itself (prefix matching) and an
API that knows the check (the consistent lie) both pass count checks, and a filter the agent never
sends cannot be checked. \sgp{} leaves these wrong.
\paragraph{What ``exact'' assumes.} The count identity assumes two-valued filters. Fields with missing
values need the three-way partition of \citet{rigger2020tlp}, which we do not use, and totals that are
estimated or change between queries can break the identity. The identity held on all \num{live-id-n} live queries we
tried, which is a small sample. The value-swap check compares live counts too. In the value-format
study, four wrong answers passed because the total changed slightly between the request and the
check. An exact check can also fire when the reported value is right (the ignore-NOT fault breaks the
check's own queries), and those runs are scored as faulted, not clean.
\paragraph{Saying unknown.} Under faults, \sgp{} mostly answers that the count cannot be determined.
This is the safe answer when the API cannot give a correct one, but it gives the user no number.
\paragraph{Models.} Seven open-weight models ran every agent experiment. Three frontier models from one vendor ran
six arms of the main benchmark and two of the real-API study. These frontier runs and the control arms
of Nemotron-3-Ultra ran on Genentech's internal platform, so we have their answer counts but cannot
release or recount their logs, and we have no hash records for them. Those runs were added after the
planned experiments, with no predictions. Seven models limit the power of model-level tests.
\paragraph{Short faults and false alarms.} Blind retry beats \sgp{} on faults that clear within one
or two calls (\S\ref{sec:res-retry}), and learned checks cost some clean success on ToolMaze.
\paragraph{Method design.} The method went through three versions, and each was tested on fresh data.
The 1\% cutoff $\tau$ was chosen after seeing earlier results, but any cutoff below
\pct{gate-safe-tau} gives the same decisions.
\paragraph{Benchmarks.} Our fault injectors run on our own APIs. The real-API probes show that
ignored filters, silently capped pages and ignored filter values occur in practice. The other fault
types are not measured in real APIs. On a further benchmark ($\tau$-bench) \sgp{} tied
the published methods (App.~\ref{app:tau}).

\section*{Ethics Statement}
The real-API study sent read-only GET requests, about one per second, with no credentials and no
personal data. Our fault injectors run only on local test services. All benchmark records are
synthetic, and no company or patient data was used in this work.

\section*{Code and Data Availability}
Code, fault injectors, the AgentOrchBench and retail services with their synthetic data, all tasks,
the prediction log and the code hashes are at \url{https://github.com/Sajib-006/silentguard}
(Apache-2.0). The AgentOrchBench tasks are also at
\url{https://huggingface.co/datasets/Sajib-006/AgentOrchBench} (CC-BY-4.0). The logs of all runs
outside Genentech's internal platform are available from the authors on request. ToolMaze and
$\tau$-bench are public benchmarks from their authors. The real-API study used only public, read-only
endpoints.

\section*{Acknowledgments}
We thank the Genentech CS CoE DDC team for their support and feedback, and the 2026 Genentech
summer internship program, where this work began. We also thank Nick Pierce, formerly of Genentech, for MCP server
development and integration with enterprise data and processing systems. His MCP servers were
the starting point that this project adopted.

\bibliographystyle{acl_natbib}
\bibliography{silentguard}

\begin{thebibliography}{53}
\providecommand{\natexlab}[1]{#1}

\bibitem[{Abdin et~al.(2024)Abdin, Aneja, Behl, Bubeck, Eldan
  et~al.}]{abdin2024phi4}
Marah Abdin, Jyoti Aneja, Harkirat Behl, S{\'e}bastien Bubeck, Ronen Eldan,
  et~al. 2024.
\newblock {Phi-4 Technical Report}.
\newblock \emph{arXiv preprint arXiv:2412.08905}.

\bibitem[{Abouelenin et~al.(2025)Abouelenin, Ashfaq, Atkinson
  et~al.}]{abouelenin2025phi4mini}
Abdelrahman Abouelenin, Atabak Ashfaq, Adam Atkinson, et~al. 2025.
\newblock {Phi-4-Mini Technical Report: Compact yet Powerful Multimodal
  Language Models via Mixture-of-LoRAs}.
\newblock \emph{arXiv preprint arXiv:2503.01743}.

\bibitem[{Advani(2026)}]{advani2026confidentclosing}
Laksh Advani. 2026.
\newblock {From Confident Closing to Silent Failure: Characterizing False
  Success in LLM Agents}.
\newblock \emph{arXiv preprint arXiv:2606.09863}.

\bibitem[{Atlidakis et~al.(2019)Atlidakis, Godefroid, and
  Polishchuk}]{atlidakis2019restler}
Vaggelis Atlidakis, Patrice Godefroid, and Marina Polishchuk. 2019.
\newblock {RESTler: Stateful REST API Fuzzing}.
\newblock In \emph{Proceedings of the 41st International Conference on Software
  Engineering ({ICSE})}, pages 748--758.

\bibitem[{Chen et~al.(1998)Chen, Cheung, and Yiu}]{chen1998metamorphic}
T.~Y. Chen, S.~C. Cheung, and S.~M. Yiu. 1998.
\newblock {Metamorphic Testing: A New Approach for Generating Next Test Cases}.
\newblock Technical Report HKUST-CS98-01, Hong Kong University of Science and
  Technology.

\bibitem[{Debenedetti et~al.(2025)Debenedetti, Shumailov, Fan, Hayes, Carlini,
  Fabian, Kern, Shi, Terzis, and Tramèr}]{debenedetti2025camel}
Edoardo Debenedetti, Ilia Shumailov, Tianqi Fan, Jamie Hayes, Nicholas Carlini,
  Daniel Fabian, Christoph Kern, Chongyang Shi, Andreas Terzis, and Florian
  Tramèr. 2025.
\newblock {Defeating Prompt Injections by Design}.
\newblock \emph{arXiv preprint arXiv:2503.18813}.

\bibitem[{Dettmers et~al.(2023)Dettmers, Pagnoni, Holtzman, and
  Zettlemoyer}]{dettmers2023qlora}
Tim Dettmers, Artidoro Pagnoni, Ari Holtzman, and Luke Zettlemoyer. 2023.
\newblock {QLoRA: Efficient Finetuning of Quantized LLMs}.
\newblock In \emph{Advances in Neural Information Processing Systems},
  volume~36, pages 10088--10115.

\bibitem[{Ernst et~al.(2007)Ernst, Perkins, Guo, McCamant, Pacheco, Tschantz,
  and Xiao}]{ernst2007daikon}
Michael~D. Ernst, Jeff~H. Perkins, Philip~J. Guo, Stephen McCamant, Carlos
  Pacheco, Matthew~S. Tschantz, and Chen Xiao. 2007.
\newblock {The Daikon System for Dynamic Detection of Likely Invariants}.
\newblock \emph{Science of Computer Programming}, 69(1--3):35--45.

\bibitem[{Geifman and El-Yaniv(2017)}]{geifman2017selective}
Yonatan Geifman and Ran El-Yaniv. 2017.
\newblock {Selective Classification for Deep Neural Networks}.
\newblock In \emph{Advances in Neural Information Processing Systems},
  volume~30, pages 4878--4887.

\bibitem[{Gupta(2026)}]{gupta2026reliabilitybench}
Aayush Gupta. 2026.
\newblock {ReliabilityBench: Evaluating LLM Agent Reliability Under
  Production-Like Stress Conditions}.
\newblock \emph{arXiv preprint arXiv:2601.06112}.

\bibitem[{Hu et~al.(2022)Hu, Shen, Wallis, Allen-Zhu, Li, Wang, Wang, and
  Chen}]{hu2021lora}
Edward~J. Hu, Yelong Shen, Phillip Wallis, Zeyuan Allen-Zhu, Yuanzhi Li, Shean
  Wang, Lu~Wang, and Weizhu Chen. 2022.
\newblock {LoRA: Low-Rank Adaptation of Large Language Models}.
\newblock In \emph{The Tenth International Conference on Learning
  Representations}.

\bibitem[{Huang et~al.(2024)Huang, Chen, Mishra, Zheng, Yu, Song, and
  Zhou}]{huang2024cannot}
Jie Huang, Xinyun Chen, Swaroop Mishra, Huaixiu~Steven Zheng, Adams~Wei Yu,
  Xinying Song, and Denny Zhou. 2024.
\newblock {Large Language Models Cannot Self-Correct Reasoning Yet}.
\newblock In \emph{The Twelfth International Conference on Learning
  Representations}.

\bibitem[{Jiang et~al.(2023)Jiang, Sablayrolles, Mensch, Bamford, Chaplot,
  de~las Casas, Bressand, Lengyel, Lample, Saulnier, Lavaud, Lachaux, Stock,
  Scao, Lavril, Wang, Lacroix, and Sayed}]{jiang2023mistral}
Albert~Q. Jiang, Alexandre Sablayrolles, Arthur Mensch, Chris Bamford,
  Devendra~Singh Chaplot, Diego de~las Casas, Florian Bressand, Gianna Lengyel,
  Guillaume Lample, Lucile Saulnier, L{\'e}lio~Renard Lavaud, Marie-Anne
  Lachaux, Pierre Stock, Teven~Le Scao, Thibaut Lavril, Thomas Wang,
  Timoth{\'e}e Lacroix, and William~El Sayed. 2023.
\newblock {Mistral 7B}.
\newblock \emph{arXiv preprint arXiv:2310.06825}.

\bibitem[{Kalai et~al.(2025)Kalai, Nachum, Vempala, and
  Zhang}]{kalai2025hallucinate}
Adam~Tauman Kalai, Ofir Nachum, Santosh~S. Vempala, and Edwin Zhang. 2025.
\newblock {Why Language Models Hallucinate}.
\newblock \emph{arXiv preprint arXiv:2509.04664}.

\bibitem[{Kamath et~al.(2025)Kamath, Ferret, Pathak, Vieillard, Merhej
  et~al.}]{kamath2025gemma3}
Aishwarya Kamath, Johan Ferret, Shreya Pathak, Nino Vieillard, Ramona Merhej,
  et~al. 2025.
\newblock {Gemma 3 Technical Report}.
\newblock \emph{arXiv preprint arXiv:2503.19786}.

\bibitem[{Kamath et~al.(2020)Kamath, Jia, and Liang}]{kamath2020selective}
Amita Kamath, Robin Jia, and Percy Liang. 2020.
\newblock {Selective Question Answering under Domain Shift}.
\newblock In \emph{Proceedings of the 58th Annual Meeting of the Association
  for Computational Linguistics}, pages 5684--5696.

\bibitem[{Khan et~al.(2026)Khan, Mughees, Sudheerbabu, Ahmad, and
  Truscan}]{khan2026armeta}
Shehroz Khan, Abdullah Mughees, Gaadha Sudheerbabu, Tanwir Ahmad, and Dragos
  Truscan. 2026.
\newblock {Multi-Agent LLM-based Metamorphic Testing for REST APIs}.
\newblock \emph{arXiv preprint arXiv:2605.28321}.

\bibitem[{Kirichenko et~al.(2025)Kirichenko, Ibrahim, Chaudhuri, and
  Bell}]{kirichenko2025abstentionbench}
Polina Kirichenko, Mark Ibrahim, Kamalika Chaudhuri, and Samuel~J. Bell. 2025.
\newblock {AbstentionBench: Reasoning LLMs Fail on Unanswerable Questions}.
\newblock \emph{arXiv preprint arXiv:2506.09038}.

\bibitem[{Kwon et~al.(2023)Kwon, Li, Zhuang, Sheng, Zheng, Yu, Gonzalez, Zhang,
  and Stoica}]{kwon2023vllm}
Woosuk Kwon, Zhuohan Li, Siyuan Zhuang, Ying Sheng, Lianmin Zheng, Cody~Hao Yu,
  Joseph~E. Gonzalez, Hao Zhang, and Ion Stoica. 2023.
\newblock {Efficient Memory Management for Large Language Model Serving with
  PagedAttention}.
\newblock In \emph{Proceedings of the 29th Symposium on Operating Systems
  Principles}, pages 611--626.

\bibitem[{Li et~al.(2026)Li, Ye, Guo, and Dang}]{li2026silentprobe}
Zongrong Li, Shengkun Ye, Feiyou Guo, and Zuoyou Dang. 2026.
\newblock {SilentProbe: Measuring Silent Failure in Production APIs Used as
  Agent Tools}.
\newblock \emph{arXiv preprint arXiv:2609.00035}.

\bibitem[{Liu et~al.(2026)Liu, Peng, Cao, Wang, Deng, Chen, Yin, and
  Zhang}]{liu2026toolgate}
Yanming Liu, Xinyue Peng, Jiannan Cao, Xinyi Wang, Songhang Deng, Jintao Chen,
  Jianwei Yin, and Xuhong Zhang. 2026.
\newblock {ToolGate: Contract-Grounded and Verified Tool Execution for LLMs}.
\newblock \emph{arXiv preprint arXiv:2601.04688}.

\bibitem[{Mansoor et~al.(2026)Mansoor, Phadke, and
  Rana}]{mansoor2026verifiedtoolcalls}
Isham~Kalappurackal Mansoor, Abhishek Phadke, and Pratip Rana. 2026.
\newblock {Verified Tool Calls Improve LLM Agent Reliability Under Non-Atomic
  Failures}.
\newblock \emph{arXiv preprint arXiv:2608.02645}.

\bibitem[{{Meta}(2024)}]{meta2024llama33}
{Meta}. 2024.
\newblock {Llama 3.3 Model Card}.
\newblock \url{https://huggingface.co/meta-llama/Llama-3.3-70B-Instruct}.

\bibitem[{{Meta}(2025)}]{meta2025llama4}
{Meta}. 2025.
\newblock {Llama 4 Model Card}.
\newblock
  \url{https://huggingface.co/meta-llama/Llama-4-Scout-17B-16E-Instruct}.

\bibitem[{{NVIDIA}(2026)}]{nvidia2026nemotron3}
{NVIDIA}. 2026.
\newblock {NVIDIA Nemotron 3 Ultra Model Card}.
\newblock
  \url{https://huggingface.co/nvidia/NVIDIA-Nemotron-3-Ultra-550B-A55B-BF16}.

\bibitem[{{OpenAI}(2025)}]{openai2025gptoss}
{OpenAI}. 2025.
\newblock {gpt-oss-120b \& gpt-oss-20b Model Card}.
\newblock \emph{arXiv preprint arXiv:2508.10925}.

\bibitem[{{OpenAI}(2026)}]{openai2026gpt56}
{OpenAI}. 2026.
\newblock {GPT-5.6 System Card}.
\newblock \url{https://deploymentsafety.openai.com/gpt-5-6}.

\bibitem[{Panthi and Abdelfattah(2026)}]{panthi2026outcome}
Sugam Panthi and Rabab Abdelfattah. 2026.
\newblock {Outcome Monitors: Recovery Affordances for Silent Tool Failures}.
\newblock \emph{arXiv preprint arXiv:2608.19303}.

\bibitem[{Rebedea et~al.(2023)Rebedea, Dinu, Sreedhar, Parisien, and
  Cohen}]{rebedea2023nemo}
Traian Rebedea, Razvan Dinu, Makesh~Narsimhan Sreedhar, Christopher Parisien,
  and Jonathan Cohen. 2023.
\newblock {NeMo Guardrails: A Toolkit for Controllable and Safe LLM
  Applications with Programmable Rails}.
\newblock In \emph{Proceedings of the 2023 Conference on Empirical Methods in
  Natural Language Processing: System Demonstrations}, pages 431--445.

\bibitem[{Reddy et~al.(2026)Reddy, Challaram, and Basu}]{reddy2026reasonless}
Vikas Reddy, Sumanth~Reddy Challaram, and Abhishek Basu. 2026.
\newblock {Reason Less, Verify More: Deterministic Gates Recover a Silent
  Policy-Violation Failure Mode in Tool-Using LLM Agents}.
\newblock \emph{arXiv preprint arXiv:2607.07405}.

\bibitem[{Rigger and Su(2020)}]{rigger2020tlp}
Manuel Rigger and Zhendong Su. 2020.
\newblock \href {https://doi.org/10.1145/3428279} {{Finding Bugs in Database
  Systems via Query Partitioning}}.
\newblock \emph{Proceedings of the ACM on Programming Languages},
  4(OOPSLA):1--30.

\bibitem[{Segura et~al.(2016)Segura, Fraser, Sanchez, and
  Ruiz-Cort{\'{e}}s}]{segura2016survey}
Sergio Segura, Gordon Fraser, Ana~B. Sanchez, and Antonio Ruiz-Cort{\'{e}}s.
  2016.
\newblock {A Survey on Metamorphic Testing}.
\newblock \emph{{IEEE} Transactions on Software Engineering}, 42(9):805--824.

\bibitem[{Segura et~al.(2018)Segura, Parejo, Troya, and
  Ruiz-Cort{\'e}s}]{segura2018restful}
Sergio Segura, Jos{\'e}~A. Parejo, Javier Troya, and Antonio Ruiz-Cort{\'e}s.
  2018.
\newblock {Metamorphic Testing of RESTful Web APIs}.
\newblock \emph{IEEE Transactions on Software Engineering}, 44(11):1083--1099.

\bibitem[{Sethi et~al.(2026)Sethi, Kenzhebayev, Paturi, Raina, Raina, and
  Sheth}]{sethi2026fabrication}
Arham Sethi, Arsen Kenzhebayev, Saanvi Paturi, Vatsal Raina, Vyas Raina, and
  Ivaxi Sheth. 2026.
\newblock {Fabrication After Tool Failure: Tool-Augmented Agents Assert Values
  Their Tools Did Not Return}.
\newblock \emph{arXiv preprint arXiv:2609.14758}.

\bibitem[{Shi et~al.(2025)Shi, He, Wang, Li, Wu, Guo, and
  Song}]{shi2025progent}
Tianneng Shi, Jingxuan He, Zhun Wang, Hongwei Li, Linyu Wu, Wenbo Guo, and Dawn
  Song. 2025.
\newblock {Progent: Securing AI Agents with Privilege Control}.
\newblock \emph{arXiv preprint arXiv:2504.11703}.

\bibitem[{Sun et~al.(2024)Sun, Min, Chang, and Bisk}]{sun2024toolsfail}
Jimin Sun, So~Yeon Min, Yingshan Chang, and Yonatan Bisk. 2024.
\newblock \href {https://doi.org/10.18653/v1/2024.emnlp-main.790} {{Tools Fail:
  Detecting Silent Errors in Faulty Tools}}.
\newblock In \emph{Proceedings of the 2024 Conference on Empirical Methods in
  Natural Language Processing}, pages 14272--14289.

\bibitem[{Ta et~al.(2026)Ta, Zhu, and Shayandeh}]{ta2026reinforced}
Anh Ta, Junjie Zhu, and Shahin Shayandeh. 2026.
\newblock \href {https://doi.org/10.18653/v1/2026.gem-main.13} {{Reinforced
  Agent: Inference-Time Feedback for Tool-Calling Agents}}.
\newblock In \emph{Proceedings of the Fifth Workshop on Generation, Evaluation
  and Metrics ({GEM})}, pages 136--147.

\bibitem[{Tian et~al.(2026)Tian, Shi, Zhou, and Zhao}]{tian2026beyondfunction}
Yang Tian, Zhengpeng Shi, Yu~Zhou, and Bo~Zhao. 2026.
\newblock {Beyond Function Calling: Benchmarking Tool-Using Agents under
  Tool-Environment Unreliability}.
\newblock \emph{arXiv preprint arXiv:2606.25819}.

\bibitem[{Wang et~al.(2025)Wang, Poskitt, and Sun}]{wang2025agentspec}
Haoyu Wang, Christopher~M. Poskitt, and Jun Sun. 2025.
\newblock {AgentSpec: Customizable Runtime Enforcement for Safe and Reliable
  LLM Agents}.
\newblock \emph{arXiv preprint arXiv:2503.18666}.

\bibitem[{Wang et~al.(2026)Wang, Chen, Wang, Wu, Fang, Cai, Gu, Su, Zhang,
  Wang, Cai, and Chua}]{wang2026agentnoisebench}
Ruipeng Wang, Yuxin Chen, Yukai Wang, Chang Wu, Junfeng Fang, Xiaodong Cai,
  Qi~Gu, Hui Su, An~Zhang, Xiang Wang, Xunliang Cai, and Tat-Seng Chua. 2026.
\newblock {AgentNoiseBench: Benchmarking Robustness of Tool-Using LLM Agents
  Under Noisy Condition}.
\newblock \emph{arXiv preprint arXiv:2602.11348}.

\bibitem[{Wang and Vemuri(2026)}]{wang2026tooldecides}
Zhongyuan Wang and Pratyusha Vemuri. 2026.
\newblock {When the Tool Decides: LLM Agents Defer Blindly to Graph Neural
  Network Tools, and Stronger Backbones Defer More}.
\newblock \emph{arXiv preprint arXiv:2606.14476}.

\bibitem[{Wen et~al.(2025)Wen, Yao, Feng, Xu, Tsvetkov, Howe, and
  Wang}]{wen2025abstention}
Bingbing Wen, Jihan Yao, Shangbin Feng, Chenjun Xu, Yulia Tsvetkov, Bill Howe,
  and Lucy~Lu Wang. 2025.
\newblock {Know Your Limits: A Survey of Abstention in Large Language Models}.
\newblock \emph{Transactions of the Association for Computational Linguistics},
  13:529--556.

\bibitem[{Xie et~al.(2024)Xie, Zhang, Chen, Lou, and Su}]{xie2024adaptive}
Jian Xie, Kai Zhang, Jiangjie Chen, Renze Lou, and Yu~Su. 2024.
\newblock {Adaptive Chameleon or Stubborn Sloth: Revealing the Behavior of
  Large Language Models in Knowledge Conflicts}.
\newblock In \emph{The Twelfth International Conference on Learning
  Representations}.

\bibitem[{Yan et~al.(2026)Yan, Li, Han, Li, Wang, Wang, Lyu, and
  Chen}]{yan2026trustnotool}
Lecheng Yan, Ruizhe Li, Xicheng Han, Wenxi Li, Binwu Wang, Longyue Wang,
  Chenyang Lyu, and Guanhua Chen. 2026.
\newblock {Trust No Tool: Evaluating and Defending LLM Agents under Untrusted
  Tool Feedback}.
\newblock \emph{arXiv preprint arXiv:2605.17453}.

\bibitem[{Yang et~al.(2026)Yang, Song, Kim, Song, Park, and
  Jo}]{yang2026overreliance}
Hoyeol Yang, Woojung Song, Taewon Kim, Jonghyun Song, Seoyeon Park, and Yohan
  Jo. 2026.
\newblock {Agents' Overreliance on Unreliable Tools}.
\newblock \emph{arXiv preprint arXiv:2609.05587}.

\bibitem[{Yao et~al.(2025)Yao, Shinn, Razavi, and Narasimhan}]{yao2024tau}
Shunyu Yao, Noah Shinn, Pedram Razavi, and Karthik Narasimhan. 2025.
\newblock {$\tau$-bench: A Benchmark for Tool-Agent-User Interaction in
  Real-World Domains}.
\newblock In \emph{The Thirteenth International Conference on Learning
  Representations}.

\bibitem[{Yu et~al.(2025)Yu, Zhang, Zhu, Yuan, Zuo, Yue, Dai, Fan, Liu, Liu,
  Liu, Lin, Lin, Ma, Sheng, Tong, Zhang, Zhang, Zhang, Zhu, Zhu, Chen, Chen,
  Wang, Yu, Song, Wei, Zhou, Liu, Ma, Zhang, Yan, Qiao, Wu, and
  Wang}]{yu2025dapo}
Qiying Yu, Zheng Zhang, Ruofei Zhu, Yufeng Yuan, Xiaochen Zuo, Yu~Yue, Weinan
  Dai, Tiantian Fan, Gaohong Liu, Lingjun Liu, Xin Liu, Haibin Lin, Zhiqi Lin,
  Bole Ma, Guangming Sheng, Yuxuan Tong, Chi Zhang, Mofan Zhang, Wang Zhang,
  and 16 others. 2025.
\newblock {DAPO: An Open-Source LLM Reinforcement Learning System at Scale}.
\newblock \emph{arXiv preprint arXiv:2503.14476}.

\bibitem[{Zhang et~al.(2026{\natexlab{a}})Zhang, Wan, Yu, Zhou, Zhao, Wu, Zhou,
  and Tsang}]{zhang2026dontblindlytrust}
Chubin Zhang, Zhenglin Wan, Xingrui Yu, Pengfei Zhou, Wangbo Zhao, Jingxuan Wu,
  Yaxin Zhou, and Ivor Tsang. 2026{\natexlab{a}}.
\newblock {Don't Blindly Trust It: How Unreliable Feedback Breaks Tool-Using
  LLM Agents}.
\newblock \emph{arXiv preprint arXiv:2606.21409}.

\bibitem[{Zhang et~al.(2026{\natexlab{b}})Zhang, Xu, and
  Chen}]{zhang2026harnessvalue}
Yukun Zhang, Kemu Xu, and Yishen Chen. 2026{\natexlab{b}}.
\newblock {How Do Agent Harnesses Create Value? Planning Information and
  Release Control in Stateful LLM Agents}.
\newblock \emph{arXiv preprint arXiv:2609.20474}.

\bibitem[{Zhao et~al.(2026)Zhao, Zhang, Gu, Sun, Pei, Bansal, Rajmohan, and
  Ma}]{zhao2026agenttether}
Chenyu Zhao, Shenglin Zhang, Wenwei Gu, Yongqian Sun, Dan Pei, Chetan Bansal,
  Saravan Rajmohan, and Minghua Ma. 2026.
\newblock {AgentTether: Graph-Guided Diagnosis and Runtime Intervention for
  Reliable LLM Agent Operation}.
\newblock \emph{arXiv preprint arXiv:2607.06273}.

\bibitem[{Zheng et~al.(2023)Zheng, Chiang, Sheng, Zhuang, Wu, Zhuang, Lin, Li,
  Li, Xing, Zhang, Gonzalez, and Stoica}]{zheng2023judging}
Lianmin Zheng, Wei-Lin Chiang, Ying Sheng, Siyuan Zhuang, Zhanghao Wu, Yonghao
  Zhuang, Zi~Lin, Zhuohan Li, Dacheng Li, Eric~P. Xing, Hao Zhang, Joseph~E.
  Gonzalez, and Ion Stoica. 2023.
\newblock {Judging LLM-as-a-Judge with MT-Bench and Chatbot Arena}.
\newblock In \emph{Advances in Neural Information Processing Systems},
  volume~36, pages 46595--46623.

\bibitem[{Zheng et~al.(2026)Zheng, Wu, and Chang}]{zheng2026toolrobustbench}
YiShan Zheng, Yuan Wu, and Yi~Chang. 2026.
\newblock {ToolRobustBench: Stage-Wise Perturbation Evaluation and Failure
  Diagnosis for Tool-Calling Agents}.
\newblock \emph{arXiv preprint arXiv:2608.23635}.

\bibitem[{Zhu et~al.(2026)Zhu, Ma, Shen, Li, Zhao, Wang, Yan, and
  Yin}]{zhu2026whentoolsfail}
Dongsheng Zhu, Xuchen Ma, Yucheng Shen, Xiang Li, Yukun Zhao, Shuaiqiang Wang,
  Lingyong Yan, and Dawei Yin. 2026.
\newblock {When Tools Fail: Benchmarking Dynamic Replanning and Anomaly
  Recovery in LLM Agents}.
\newblock \emph{arXiv preprint arXiv:2606.05806}.

\end{thebibliography}

\appendix
\renewcommand{\topfraction}{0.9}\renewcommand{\dbltopfraction}{0.9}\renewcommand{\bottomfraction}{0.6}
\renewcommand{\textfraction}{0.05}\renewcommand{\floatpagefraction}{0.75}\renewcommand{\dblfloatpagefraction}{0.75}
\setcounter{topnumber}{4}\setcounter{dbltopnumber}{3}\setcounter{bottomnumber}{2}\setcounter{totalnumber}{6}

\section{Benchmark Details}
\label{app:bench}

This section gives the tasks, faults and grading of each benchmark.

\subsection{AgentOrchBench}
\label{app:aob}

\paragraph{Agent protocol.} The agent solves one task per run. In each turn it writes exactly one of
\texttt{RUN: <one curl command>} or \texttt{ANSWER: <answer>}, and after each \texttt{RUN} it receives
the HTTP status and the response body. Pipes, other shell commands and file access are not allowed.
All eight local services (among them a sample registry, a metadata catalog and a task tracker, on
ports 8001 to 8008) are reset before every task, so runs do not affect each other.

\paragraph{Tasks.} The 36 count tasks each ask how many of the 400 catalog records match two
(15 tasks) or three (21 tasks) conditions, for example \emph{``How many catalog assets have
QC\_Status=qc-pass AND Sample\_Type=tumor? Return: count=<n>''}. The ground truth is computed from
the seed data. The records are synthetic, so no model can know a count from its training data.

\paragraph{Fault injection.} Faults are switched on inside the catalog server for a whole run. The
server echoes the fault mode back, and the runner refuses to start if the echo does not match, so a
run can never use the wrong fault. Every fault returns HTTP 200 and a well-formed body, and repeats
on every call. The catalog returns only the count (\texttt{total\_count}), as a real aggregation
endpoint does. With full records returned, a model could check the filter by reading them.
Table~\ref{tab:faults} lists the fault types and the kind of bug each is modeled on.

\begin{table}[h]
\centering\small
\setlength{\tabcolsep}{2.4pt}
\begin{tabular}{@{}lllc@{}}
\toprule
\rowcolor{head}\textbf{Fault} & \textbf{API behavior} & \textbf{Modeled on} & \textbf{Check}\\
\midrule
Drop last & drops last filter & wrong filter name & \checkmark \\
Cap total & reports at most 5 & page size as total & \checkmark \\
AND$\to$OR & uses OR & query builder & \checkmark \\
Ignore field & ignores any filter & unindexed field & \checkmark \\
Ignore NOT & drops negation & unsupported NOT & FA \\
Off by one & count $+1$ & range bug & \checkmark \\
\midrule
Prefix match & prefix, not equal & loose matching & -- \\
Consistent lie & lies everywhere & adversarial API & -- \\
\bottomrule
\end{tabular}
\caption{Silent fault types in AgentOrchBench and the kind of API bug each is modeled on. The first is the
failure we measure in real APIs (\S\ref{sec:res-realapis}). The Check column says whether the exact count
check (Eq.~\ref{eq:partition}) fires (\checkmark), fires although the count may be right (FA, false alarm),
or cannot see the fault (--), because the fault keeps the identity true.}
\label{tab:faults}
\end{table}

\paragraph{Grading.} We read the number after \texttt{count=} or in the \texttt{ANSWER} line. An
answer is \emph{correct} if it equals the ground truth, \emph{unknown} if it states that the
count cannot be determined or gives no count, and \emph{wrong} otherwise. Confidence values, task
identifiers and numbers quoted from the question are not read as counts. The grader is the same
for every method and was frozen before the planned runs.

\paragraph{List tasks.} The 30 list tasks ask for the \texttt{item\_id} of every matching record (``List the
item\_id of every catalog asset that has Batch\_ID=batch\_1 AND File\_Format=h5ad. Return:
ids=<comma-separated item\_ids>''), graded as a set. The faults are six types used in design (drop last,
ignore field, AND$\to$OR, prefix, consistent lie, cap total) and five written after the method was
frozen (word-stem text match, tag-join double counting, partial shard timeout, lossy index, cursor
ties). An empty list counts as no answer, not as a wrong one.

\paragraph{Retail domain.} The retail domain is a second schema, an orders API with four services (ports 8001 to 8004),
30 count tasks and four fault types (drop last, cap total, AND$\to$OR, ignore field), with the same
protocol, grading and clean runs.

\subsection{Real Public APIs}
\label{app:realapis}

\begin{figure}[h]
\centering
\includegraphics[width=\columnwidth]{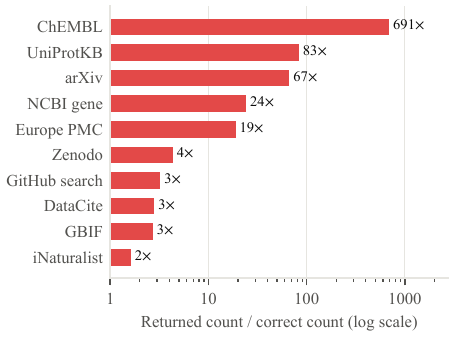}
\caption{How many times too large the returned count is when a real API silently ignores a
misnamed filter (log scale). The overcount spans more than two orders of magnitude.}
\label{fig:inflation}
\end{figure}

\paragraph{Probes.} For each of \num{e3n-apis} public research APIs we sent an unfiltered query,
the documented filtered query, and two or three plausible wrong filter names, all as read-only GET
requests at about one request per second (Table~\ref{tab:realapis}). Figure~\ref{fig:inflation} shows
the overcounts.

\begin{table}[h]
\centering
\footnotesize
\setlength{\tabcolsep}{4pt}
\begin{tabular}{@{}lrrrc@{}}
\toprule
\rowcolor{head}
\textbf{API} & \textbf{Returned} & \textbf{Correct} & \textbf{Over} & \textbf{Caught}\\
\midrule
\multicolumn{5}{@{}l}{\textit{Misnamed filter ignored (HTTP 200)}}\\
ChEMBL & 2{,}921{,}148 & 4{,}225 & 691$\times$ & 3/3\\
UniProtKB & 55{,}460 & 666 & 83$\times$ & 3/3\\
arXiv & 186{,}698 & 2{,}799 & 67$\times$ & 2/2\\
NCBI gene & 37{,}885 & 1{,}567 & 24$\times$ & 3/3\\
Europe PMC & 291{,}095 & 15{,}436 & 19$\times$ & 3/3\\
Zenodo & 110{,}913 & 25{,}549 & 4$\times$ & 2/3\\
GitHub search & 55{,}661 & 17{,}225 & 3$\times$ & 2/2\\
DataCite & 609{,}191 & 216{,}660 & 3$\times$ & 2/3\\
GBIF & 31{,}660 & 11{,}681 & 3$\times$ & 3/3\\
iNaturalist & 18{,}185 & 11{,}212 & 2$\times$ & 3/3\\
\midrule
\multicolumn{5}{@{}l}{\textit{Misnamed filter rejected (error)}}\\
ClinicalTrials.gov & 24{,}424 & 1{,}963 & -- & --\\
OpenAlex & 611{,}853 & 37{,}622 & -- & --\\
Crossref & 89{,}634 & 4{,}268 & -- & --\\
openFDA & 617{,}937 & 435{,}397 & -- & --\\
Semantic Scholar & 214{,}821 & -- & -- & --\\
\bottomrule
\end{tabular}
\caption{Real-API probes. Of 15 public APIs, 10 answer a request with a plausibly misnamed filter with HTTP 200 and the unfiltered count (ignored). The others reject the unknown parameter with an error (rejected). Returned: the count with the misnamed filter, or the unfiltered count when the API rejected it. Correct: the count with the documented filter. Over: returned / correct. Caught: misnamed filters flagged by the value-swap check (swap the filter's value, an ignored filter returns the same count). --: not applicable, or not measured (rate limit).}
\label{tab:realapis}
\end{table}

\paragraph{Agent study.} On nine of these APIs agents answer five count tasks each, given either the
right documentation or documentation with one filter name slightly wrong. Every wrong name was measured
to be silently ignored. The ground truth is the documented query's count, taken right before and right
after each run. The first four APIs (NCBI, UniProt, Europe PMC, GBIF) were chosen before the method
was frozen. The other five (Zenodo, ChEMBL, arXiv, DataCite, iNaturalist) were added later, with the
method unchanged. Zenodo, ChEMBL and DataCite place the total after the record list, and the agent sees
the first 1,500 characters of a response, so for these three the JSON is shown with the object that
holds the total first (nothing removed or changed). Table~\ref{tab:app-realapi-per} gives each API
and Table~\ref{tab:app-realapi-models} each model, with the three frontier models.

\begin{table}[h]
\centering
\small
\setlength{\tabcolsep}{3pt}
\begin{tabular}{@{}lccccc@{}}
\toprule
\rowcolor{head}
 & \multicolumn{3}{c}{\textbf{Wrong}$\downarrow$} & \multicolumn{2}{c}{\textbf{Correct}$\uparrow$}\\
\rowcolor{head}\cmidrule(lr){2-4}\cmidrule(l){5-6}
\rowcolor{head}\textbf{API} & None & Retry & Ours & None & Ours\\
\midrule
NCBI gene & 86 & 91 & \textbf{57} & 6 & \textbf{29}\\
UniProtKB & 80 & 83 & \textbf{40} & 6 & \textbf{51}\\
Europe PMC & 91 & 97 & \textbf{51} & 0 & \textbf{26}\\
GBIF & 94 & 91 & \textbf{6} & 0 & \textbf{94}\\
\midrule
Zenodo & 100 & 91 & \textbf{3} & 0 & \textbf{83}\\
ChEMBL & 100 & 94 & \textbf{0} & 0 & \textbf{100}\\
arXiv & 71 & 80 & \textbf{29} & 20 & \textbf{63}\\
DataCite & 97 & 86 & \textbf{6} & 0 & \textbf{80}\\
iNaturalist & 89 & 91 & \textbf{0} & 0 & \textbf{100}\\
\bottomrule
\end{tabular}
\caption{Real-API agent study per API, wrong documentation (\% of runs, 7 models pooled, 35 runs per API and arm). The four APIs above the line were chosen before the method was frozen. The five below were added later, with the method unchanged. On NCBI, UniProt, Europe PMC and arXiv the correct filter goes inside a query-language string that the wrong documentation never mentions, so after the spec check rejects the misnamed parameter the agent must find that syntax itself. On the others the correct parameter is in the list of allowed parameters that the spec check returns, which is the most likely reason for the gap. None: no guard. Retry: blind retry. Ours: SilentGuard (bold).}
\label{tab:app-realapi-per}
\end{table}

\begin{table}[h]
\centering
\footnotesize
\setlength{\tabcolsep}{2.5pt}
\begin{tabular}{@{}lcccc@{}}
\toprule
\rowcolor{head}
 & \textbf{No guard} & \multicolumn{2}{c}{\textbf{SilentGuard}} & \textbf{Right docs}\\
\rowcolor{head}\cmidrule(lr){3-4}
\rowcolor{head}\textbf{Model} & Wrong$\downarrow$ & Wrong$\downarrow$ & Corr.$\uparrow$ & Corr.$\uparrow$\\
\midrule
Llama-4-Scout & 97.8 & 35.6 & 48.9 & 97.8\\
Llama-4-Maverick & 93.3 & 6.7 & 93.3 & 100.0\\
Llama-3.3-70B & 100.0 & 28.9 & 68.9 & 97.8\\
gpt-oss-120b & 80.0 & 11.1 & 88.9 & 100.0\\
Nemotron-3-Ultra & 100.0 & 20.0 & 80.0 & 100.0\\
Gemma-3-27B & 62.2 & 22.2 & 40.0 & 53.3\\
Phi-4 & 95.6 & 24.4 & 66.7 & 95.6\\
\rowcolor{ours}\textit{Mean, open} & 89.8 & 21.3 & 69.5 & 92.1\\
\midrule
GPT-5.6 Sol & 48.9 & 8.9 & 91.1 & 100.0\\
GPT-5.6 Terra & 48.9 & 8.9 & 91.1 & 100.0\\
GPT-5.6 Luna & 75.6 & 8.9 & 88.9 & 100.0\\
\rowcolor{ours}\textit{Mean, frontier} & 57.8 & 8.9 & 90.4 & 100.0\\
\bottomrule
\end{tabular}
\caption{Agents on nine real public APIs, per model (45 runs per model and cell, \%). No guard and SilentGuard: agents given slightly wrong documentation. Right docs: \% correct with no guard and the right documentation. The frontier models ran these two arms only.}
\label{tab:app-realapi-models}
\end{table}

\paragraph{Other silent faults.} We also probed every API for two more fault kinds, a typo in the
value of a valid filter and a request for more records per page than allowed
(Table~\ref{tab:app-faultkinds}). On the three APIs that silently ignore a bad value we then tested, with
no model, whether a check fires on the query an agent would send (Table~\ref{tab:app-badvalue}).

\begin{table}[h]
\centering
\footnotesize
\setlength{\tabcolsep}{4pt}
\begin{tabular}{@{}lccc@{}}
\toprule
\rowcolor{head}
 & \textbf{Misnamed} & \textbf{Bad filter} & \textbf{Page over}\\
\rowcolor{head}\textbf{API} & \textbf{filter} & \textbf{value} & \textbf{limit}\\
\midrule
NCBI gene & \textbf{\textcolor{loss}{ignored}} & applied & honored\\
UniProtKB & \textbf{\textcolor{loss}{ignored}} & rejected & rejected\\
ClinicalTrials.gov & rejected & rejected & \textbf{\textcolor{loss}{capped}}\\
Europe PMC & \textbf{\textcolor{loss}{ignored}} & \textbf{\textcolor{loss}{ignored}} & rejected\\
OpenAlex & rejected & rejected & rejected\\
Crossref & rejected & rejected & rejected\\
openFDA & rejected & rejected & rejected\\
DataCite & \textbf{\textcolor{loss}{ignored}} & applied & \textbf{\textcolor{loss}{capped}}\\
Zenodo & \textbf{\textcolor{loss}{ignored}} & applied & rejected\\
GitHub search & \textbf{\textcolor{loss}{ignored}} & \textbf{\textcolor{loss}{ignored}} & \textbf{\textcolor{loss}{capped}}\\
ChEMBL & \textbf{\textcolor{loss}{ignored}} & rejected & \textbf{\textcolor{loss}{capped}}\\
arXiv & \textbf{\textcolor{loss}{ignored}} & applied & honored\\
GBIF & \textbf{\textcolor{loss}{ignored}} & applied & \textbf{\textcolor{loss}{capped}}\\
iNaturalist & \textbf{\textcolor{loss}{ignored}} & \textbf{\textcolor{loss}{ignored}} & \textbf{\textcolor{loss}{capped}}\\
Semantic Scholar & rejected & -- & --\\
\bottomrule
\end{tabular}
\caption{Three kinds of silent fault in real public APIs (read-only probes). Misnamed filter: a plausibly misnamed filter parameter. Rejected means the API returned an error. Bad filter value: a typo in the value of a valid filter (for example a bad year or date). Ignored means the count equals the unfiltered one, applied means the value was used as given (usually zero results), which is correct. Page over limit: a request for more records per page than the API allows. Capped means HTTP 200 with fewer records than asked and no error. Honored means the API returned the larger page. Bold red: silent. --: rate-limited, not measured.}
\label{tab:app-faultkinds}
\end{table}

\begin{table}[h]
\centering
\footnotesize
\setlength{\tabcolsep}{2.3pt}
\begin{tabular}{@{}llcccc@{}}
\toprule
\rowcolor{head}
\textbf{API} & \textbf{Typo in} & \textbf{Ignored} & \textbf{Over} & \textbf{Fired} & \textbf{FA}\\
\midrule
Europe PMC & query string & 5/5 & 19--40$\times$ & 5/5 & 0/5\\
GitHub search & query string & 5/5 & 3.2--4.4$\times$ & 5/5 & 0/5\\
iNaturalist & parameter & 5/5 & 1.1--1.5$\times$ & 5/5 & 0/5\\
\bottomrule
\end{tabular}
\caption{A typo in the \emph{value} of a valid filter, on the three APIs that silently ignore it (five probes each, read-only, no model). The parameter name is valid, so the spec check passes. Ignored: the API returned the unfiltered count. Over: returned / correct. Fired: the count identity (Europe PMC, GitHub, where the value sits inside the query string) or the value-swap check (iNaturalist) flagged the result. FA: the same check fired on the correctly spelled query.}
\label{tab:app-badvalue}
\end{table}

\paragraph{Wrong value format.} To test a real fault that the spec check cannot see, we gave agents
iNaturalist documentation with a valid parameter (\texttt{d2}, observed on or before a date) but the
wrong date format (DD/MM/YYYY). The API silently ignores such a date and returns the unfiltered count.
The harness was the frozen one. With no guard and with the spec check alone almost every answer was
wrong (\pct{bf-none-wrong} and \pct{bf-schema-wrong}). \sgp{} left \pct{bf-sgp-wrong} wrong, \pct{bf-sgp-unknown} unknown and only
\pct{bf-sgp-correct} correct. That is less recovery than we predicted (Table~\ref{tab:app-badformat},
and V4 in Table~\ref{tab:app-prereg}). Of its seven wrong answers, four passed the value-swap check
because the live total changed slightly between the request and the check, and three came from one
model that, after the withhold, switched to a different valid filter (the year). For this study the
backend of Gemma-3-27B was named explicitly, because automatic routing had stopped selecting one.

\begin{table}[h]
\centering
\footnotesize
\setlength{\tabcolsep}{4pt}
\begin{tabular}{@{}lccc@{}}
\toprule
\rowcolor{head}
\textbf{Method} & Wrong$\downarrow$ & Correct$\uparrow$ & Unknown\\
\midrule
No guard & 97.1 & 0.0 & 2.9\\
Spec check alone & 100.0 & 0.0 & 0.0\\
\rowcolor{ours}\textbf{SilentGuard} & 20.0 & 2.9 & 77.1\\
\bottomrule
\end{tabular}
\caption{A wrong value format in the documentation of a real API (iNaturalist date filter, 5 tasks, mean over 7 open models, \% of runs). The documentation gives the date as DD/MM/YYYY, which the API silently ignores. The parameter name is valid, so the spec check passes. With the right documentation an unguarded agent is correct 88.6\% of the time. Predictions for this run are V1 to V5 in Table~\ref{tab:app-prereg}.}
\label{tab:app-badformat}
\end{table}

\paragraph{Count identity on final queries.} We also replayed, with no model, the final query of the
\sgp{} runs of the agent study on the \num{rp-apis} APIs that take a query string. We could replay
\num{rp-wrong} wrong answers. Adding the count identity would have caught \num{rp-fires} of them. In
\num{rp-one} the final query had a single term, so the agent had dropped the second condition. In
\num{rp-holds} the identity held, because the agent had written a different query that is consistent
with itself, for example plain words in place of a field. No check built from the API's own answers
can see those. On \num{rp-correct} correct answers with two or more terms the identity raised
\num{rp-correct-fires} false alarms.

\paragraph{Which operations allow which check.} From the documentation of \num{chk-apis} public APIs,
we classified \num{chk-n} operations by the checks they allow (Table~\ref{tab:app-checkability}). Every
search or list operation allows an exact check, while lookups by ID mostly allow only the spec check.

\begin{table}[h]
\centering
\small
\setlength{\tabcolsep}{4pt}
\begin{tabular}{@{}lcc@{}}
\toprule
\rowcolor{head}
\textbf{Operation kind} & \textbf{Operations} & \textbf{Exact check}\\
\midrule
Search or list & 43 & 43\\
Get by ID & 16 & 5\\
Other read & 9 & 4\\
Aggregate & 6 & 5\\
Write & 5 & 0\\
Pagination & 4 & 3\\
Count & 1 & 1\\
\midrule
Total & 84 & 61\\
\bottomrule
\end{tabular}
\caption{Which checks real API operations allow, read from the documentation of 13 public APIs. Exact check: the operation allows a record-membership check (47 operations), the count identity (25) or a value swap (56). All 84 have a published parameter list, so the spec check applies to every one. The APIs are those of Table~\ref{tab:realapis} without GitHub and openFDA. Write operations were not counted as allowing the write-then-read check of Table~\ref{tab:checks}. These are well-documented APIs. The share will be lower for undocumented or free-text tools.}
\label{tab:app-checkability}
\end{table}

\subsection{ToolMaze}
\label{app:toolmaze}

ToolMaze \citep{zhu2026whentoolsfail} has multi-step tool workflows at four complexity levels, with five
modes. These are no fault (P0), explicit errors (P1, P2), a silent fault on the first call to a tool (P3) and a
silent fault on every call (P4). We used its tools, tasks, fault injector and trace judge unchanged,
and ported every method into its agent loop. Method design used separate workflows. All ToolMaze
numbers in the paper come from 180 workflows that were never used in design, drawn
with a fixed seed. The first 80 were the test set of the second method version, whose behavior on
ToolMaze (re-check, then warn) is the same as that of the final version and was fixed before they
were run. The other 100 were first used to test the final version. For the two-call test we changed one line of the fault injector so the P3 fault
lasts the first two calls to the tool. Tasks, judge and arms are otherwise the same.

\paragraph{Harness fix.} Models sometimes wrote several tool calls in one reply. ToolMaze's loop runs
only the first but stored the whole reply, so the model later saw calls it believed had run. We stored
each reply only up to its first tool call. The fix applies to every method, and all affected runs
were deleted and run again.

\section{Implementation of SilentGuard}
\label{app:impl}

This section gives the checks, the cutoff rule and the exact messages the agent sees.

\paragraph{Count check.} For a filter query $F = C_1\wedge\dots\wedge C_k$ that returns $N$, the
harness checks each position $i$. It queries the other conditions without $C_i$ (count $T_i$) and
with $\neg C_i$ (count $M_i$), and tests $N + M_i = T_i$. That is $2k$ extra queries per filter call
(at most 6 here). On any API that applies AND and NOT correctly the identity holds, so the check
cannot fire on a correct count. A fault that keeps the identity true (prefix matching, a consistent
lie) passes it. A fault that breaks the negation queries themselves (ignore NOT) makes the check fire
although the reported count may be right.

\paragraph{Spec check and value-swap check.} When an API publishes its parameter list, a
request with a parameter outside the list is stopped before it runs, and the agent gets an error
instead of a silently wrong result. For a parameter that is in the list, the harness re-issues
the query with the filter's value swapped. If the count does not change, the filter was ignored.

\paragraph{Learned checks.} On ToolMaze we used the Outcome Monitors contract miner
\citep{panthi2026outcome}, which covers presence and type, echoed arguments, positive values, categorical values,
linear relations and date order. Contracts are mined from clean responses of \emph{other} workflows
with 5-fold cross-fitting by task, so a contract is never mined on the workflow it checks.

\paragraph{Cutoff.} A check that held on $n$ clean examples gets the false-alarm estimate
$1/(n+2)$ (rule of succession). Exact identities and the spec check count as $n=\infty$. If any
failed check has an estimate of at most the cutoff $\tau=1\%$, the failed values are withheld.
Otherwise the result stays visible with a warning. With the few clean examples ToolMaze provides, no
learned contract reaches the cutoff, so on ToolMaze \sgp{} always warns.

\paragraph{Messages shown to the agent.} After a count fails the check twice, the agent sees the
response with the count removed and the following note.
\begin{quote}\small
total\_count withheld: this result failed an exact consistency check twice (the harness already
re-ran the request). Do not guess the count; re-query differently, or answer that the count cannot
be determined.
\end{quote}
If the agent then answers with a count that did not pass the check, the harness replies (at most
twice, after which it answers that the count cannot be determined) as follows.
\begin{quote}\small
HARNESS: answer rejected. The count you report is not the count of a query that passed the
consistency check. Re-query in a way that yields a verified count, or answer that the count cannot
be determined.
\end{quote}
A count is treated as verified only if it came from a query that passed the check and that has at
least as many conditions as any other query in the run. This stops a check passed on a looser query
from backing a count for the full filter.

\paragraph{Result checklist.} \sgp{} adds this checklist, written in the style of
\citet{sun2024toolsfail}, to the system prompt.
\begin{quote}\small
RESULT CHECKLIST. Before you use any tool result, check it against this list: 1. Presence and type:
every field you need is present and has the right type. 2. Agreement with your request: values that
repeat your request (IDs, names, dates, codes) match what you asked for. 3. Plausibility: numbers
and categories are sensible (for example, no negative prices or impossible values). 4. Consistency:
the result does not contradict itself and is about what you asked. If any check fails, treat the
result as incorrect: retry the tool, use another tool that can do the same step, or stop and report
which step failed.
\end{quote}

\section{Baselines and Published Methods}
\label{app:baselines}

All methods run in the same harness, with the same models, tasks, decoding settings and turn limits.
Each prompt follows the design of the cited paper and is adapted to our tasks. We say below which
text is the authors' own.

\paragraph{Checklist \citep{sun2024toolsfail}.} This arm adds the result checklist above to the system prompt. The
source uses task-specific checklists, and ours is a general one in the same style.

\paragraph{Reviewer \citep{ta2026reinforced}.} Before each tool call runs, the agent's own model
reviews it, and the agent may revise the call (up to five review loops, as in the source). The source
uses a separate reviewer model and its own prompts. We wrote the prompt below for our tasks, following
their design.
\begin{quote}\small
You are a reviewer agent. A primary assistant is carrying out a user's task with tools. Before its
provisional tool call is executed, decide whether the call is correct. Check the call against the
user request, the tool documentation, and the whole conversation so far, including earlier tool
results. Look for: a wrong tool, wrong or made-up arguments, arguments copied from a tool result that
is wrong or inconsistent, skipped or repeated steps, and continuing when the workflow should stop.
[...] Reply with JSON only.
\end{quote}

\paragraph{Outcome Monitors \citep{panthi2026outcome}.} Contracts are mined as above. A response that
breaks one gets the authors' non-binding advisory below, with ``the retail database'' replaced by
``this API''.
\begin{quote}\small
[OUTCOME CONTRACT ADVISORY] The observed result [...] is inconsistent with a known invariant of this
API. Violation: [...]. This advisory is nonbinding. Treat the result as unreliable and verify or
recover using the available documented tools and task context before relying on it.
\end{quote}
On AgentOrchBench, contracts are mined from clean responses of other tasks, 5-fold by task.

\paragraph{LLM judge.} This is our own baseline, an LLM judge in the sense of
\citet{zheng2023judging}. After each filter query, a fixed judge model (gpt-oss-120b, the same for
every agent) sees the task, the request filters, the response and the
results of the harness's own consistency queries, and returns TRUSTWORTHY or SUSPECT with a reason,
which is appended to the response. We first tried a smaller judge (GPT-4o-mini). It called correct
responses suspect even when given the passing check numbers, so we used the stronger judge to keep
the baseline fair.

\paragraph{Controls for evidence, permission and enforcement.} \emph{LLM judge, no check results}
uses the same judge and prompt, but the judge sees only the task, the request filters and the
response, not the harness's consistency queries. With \emph{+ blocking}, a result the judge calls
SUSPECT (or that breaks an Outcome Monitors contract) is withheld, and the final count must come from
a result the detector accepted, with the same two-rejection answer rule as \sgp{}. \emph{Warning + may
abstain} combines the warning of the full count check (the check \sgp{} uses, which names the failed
condition and shows no numbers) with the caution prompt below, which allows the answer that the count
cannot be determined. An earlier version used the last-condition check and a warning that shows the
check's numbers. It gave \pct{x6-adv1} wrong for the warning alone and \pct{x6-abst1} with permission,
but misses a filter ignored at another position. These ran on 7 models with 2 seeds (Nemotron-3-Ultra
on Genentech's internal platform with the same code and grader, counts only).

\paragraph{Caution prompt.} This text is added to the system prompt.
\begin{quote}\small
IMPORTANT: tool and API outputs are not always reliable. An API can return HTTP 200 with wrong data,
for example by silently ignoring part of your request. Do not blindly trust a response. Before
answering, check that each response is consistent with what you asked for; if you cannot confirm it,
say the count cannot be determined instead of reporting a number.
\end{quote}

\paragraph{Self-verification.} The model is told the same every-position check the harness runs and
must run it itself before reporting a count, or answer that the count cannot be determined. This
shows the effect of who runs the check, the model or the harness.

\paragraph{Re-query and blind retry.} Re-query sends the same query a second time and shows both
counts. Blind retry sends every request twice and shows only the second response. On ToolMaze it is
added to the checklist with Outcome Monitors, so it differs from that arm only by the retry.

\paragraph{Harness actions (ablations).} These differ in the action after a failed count check.
\emph{Warning} appends the check result, and \emph{warning + tip} also names the recovery tools. The
\emph{answer gate} rejects an answer that repeats a flagged count twice and then answers unknown.
\emph{Enforce} accepts only checked counts. \emph{Auto-repair} recomputes the count through separate
routes and shows it only if a route passed its own check. Three earlier arms, \emph{warning (last
condition)}, its variant with may abstain and the answer gate, check the last condition only. All
others use the full check.

\section{Models, Inference and Statistics}
\label{app:models}

This section gives the models, the inference settings and the statistical tests.

\paragraph{Models.} Llama-4-Scout (17B-16E) and Llama-4-Maverick (17B-128E, FP8)
\citep{meta2025llama4}, Llama-3.3-70B \citep{meta2024llama33}, Nemotron-3-Ultra (550B-A55B)
\citep{nvidia2026nemotron3}, Gemma-3-27B \citep{kamath2025gemma3} and Phi-4 (14B) \citep{abdin2024phi4}
were served through Hugging Face Inference Providers, with one backend pinned for Llama-3.3-70B.
The model gpt-oss-120b \citep{openai2025gptoss} was served by an endpoint hosted by Virginia Tech Advanced Research Computing (ARC). For the two-call
ToolMaze test, Gemma-3-27B and Phi-4 ran on ARC GPUs with vLLM \citep{kwon2023vllm} (same
weights) and Llama-3.3-70B on a second provider backend. GPT-5.6 Sol, Terra and Luna
\citep{openai2026gpt56} ran the real-API study on the OpenAI API and AgentOrchBench on Genentech's
internal platform. They do not accept temperature 0 and sample at the
provider default. Three older GPT models (GPT-4.1, GPT-4o, GPT-4o-mini) finished only part of their
runs because of API budget limits (Table~\ref{tab:app-permodel}).

\paragraph{Settings.} We used temperature 0 (the provider default for the GPT-5.6 models), at most
2,000 output tokens per turn, at most 16 turns (25 on ToolMaze) and identical prompts for every model.
Every API call has client-side rate limits and retries with backoff. A run whose calls failed at the
provider is marked, excluded from grading and run again. A few runs still failed after the rerun and
are left out (at most 7 of 504 for any model and method).

\paragraph{Statistics.} The unit of analysis is the model, because runs of one model share tasks and
are not independent. Tables report the mean over models with the standard error of that mean, and 95\%
confidence intervals ($t$ interval over models) are in Table~\ref{tab:app-ci}. Differences are tested
with an exact two-sided sign-flip test over models, which lists all $2^n$ sign patterns and
assumes nothing about the distribution. With 7 models the smallest possible $p$ is $2/128=0.016$, so
we read it as a count of how many models agree, not as a test that survives correction for
many comparisons. For the key differences we also report a task-cluster bootstrap. Each of 2{,}000
resamples draws the benchmark's tasks with replacement, the same draw for every model and arm, and
recomputes the mean over models (Table~\ref{tab:app-boot}).

\begin{table}[h]
\centering
\footnotesize
\setlength{\tabcolsep}{4pt}
\begin{tabular}{@{}llc@{}}
\toprule
\rowcolor{head}
\textbf{Measure} & \textbf{Against} & \textbf{Difference [95\% interval]}\\
\midrule
\multicolumn{3}{@{}l}{\textit{AgentOrchBench}}\\
wrong & No guard & $-72.3$ [$-74.0$, $-70.5$]\\
wrong & Self-verification & $-54.5$ [$-56.6$, $-52.3$]\\
wrong & LLM judge & $-69.5$ [$-71.3$, $-67.7$]\\
wrong & Outcome Monitors & $-70.4$ [$-72.3$, $-68.3$]\\
wrong & Blind retry & $-72.7$ [$-74.7$, $-70.5$]\\
wrong & Enforce & $-7.1$ [$-8.5$, $-5.6$]\\
correct & No guard & $-13.9$ [$-14.6$, $-13.2$]\\
correct & Auto-repair & $-54.8$ [$-56.6$, $-52.9$]\\
\midrule
\multicolumn{3}{@{}l}{\textit{Real APIs (9)}}\\
wrong & No guard & $-68.6$ [$-77.1$, $-59.4$]\\
wrong & Blind retry & $-68.3$ [$-75.2$, $-60.3$]\\
wrong & Spec check alone & $+0.3$ [$-2.5$, $+3.2$]\\
correct & No guard & $+66.0$ [$+57.1$, $+74.6$]\\
correct & Spec check alone & $-1.3$ [$-4.4$, $+1.9$]\\
\midrule
\multicolumn{3}{@{}l}{\textit{ToolMaze}}\\
success & Checklist + OM & $+3.3$ [$+1.8$, $+4.9$]\\
success & Outcome Monitors & $+5.5$ [$+3.5$, $+7.5$]\\
success & Checklist & $+12.9$ [$+10.1$, $+15.6$]\\
\bottomrule
\end{tabular}
\caption{Task-cluster bootstrap for the key differences (SilentGuard minus the other method, in percentage points, computed before rounding). We used 2{,}000 resamples of tasks, the same draw for every model and method, and the mean over 7 open models. An interval that excludes 0 means the difference holds across tasks, not only on average. OM: Outcome Monitors.}
\label{tab:app-boot}
\end{table}

\begin{table}[h]
\centering
\footnotesize
\setlength{\tabcolsep}{4pt}
\begin{tabular}{@{}lccc@{}}
\toprule
\rowcolor{head}
\textbf{Against} & \textbf{Other} & \textbf{SilentGuard} & $\lambda^{*}$\\
\midrule
\multicolumn{4}{@{}l}{\textit{AgentOrchBench, seen faults (7 models)}}\\
No guard & 16.4 / 74.0 & 2.4 / 1.7 & 0.2\\
Auto-repair & 57.3 / 3.0 & 2.4 / 1.7 & 43.6\\
\midrule
\multicolumn{4}{@{}l}{\textit{Lists, seen faults (7 models)}}\\
No guard & 43.7 / 49.3 & 21.1 / 9.1 & 0.6\\
Auto-repair & 47.9 / 14.6 & 21.1 / 9.1 & 4.9\\
\midrule
\multicolumn{4}{@{}l}{\textit{Lists, unseen faults (7 models)}}\\
No guard & 56.7 / 37.8 & 7.1 / 11.5 & 1.9\\
Auto-repair & 37.5 / 16.9 & 7.1 / 11.5 & 5.7\\
\midrule
\multicolumn{4}{@{}l}{\textit{Unseen count faults, harness only}}\\
No guard & 38.9 / 61.1 & 19.0 / 24.5 & 0.5\\
Auto-repair & 36.6 / 47.2 & 19.0 / 24.5 & 0.8\\
\bottomrule
\end{tabular}
\caption{Where withholding pays. Cells: \% correct / \% wrong of faulted runs. With score $=$ correct $-\ \lambda\times$ wrong, $\lambda^{*}$ is the cost of a wrong answer, relative to a correct one, above which SilentGuard scores higher than the other method (computed from unrounded values). Against no guard $\lambda^{*}$ is small everywhere. Against auto-repair on count tasks it is large on the fault types repair was built for and small on fault types no method was built for.}
\label{tab:app-risk}
\end{table}

\section{Full Results}
\label{app:full}

\begin{figure}[t]
\centering
\includegraphics[width=\columnwidth]{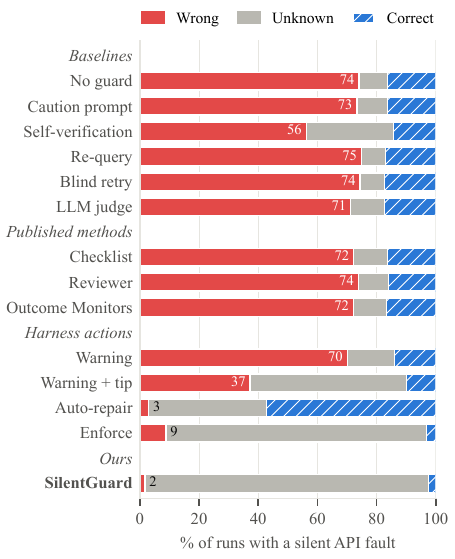}
\caption{All answers under silent faults on AgentOrchBench, mean over 7 open models. Prompts and
model-based methods leave most answers wrong. Harness actions that go beyond a warning move them out
of the wrong share, and under SilentGuard nearly all become unknown.}
\label{fig:outcomes}
\end{figure}

\begin{figure}[t]
\centering
\includegraphics[width=\columnwidth]{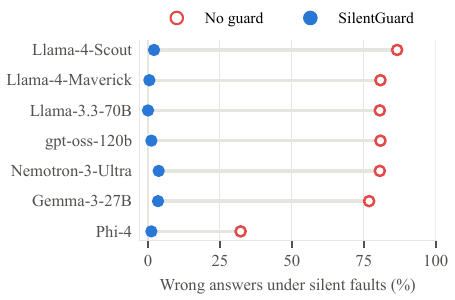}
\caption{Wrong answers per model on AgentOrchBench, no guard against \sgp{}. Every model falls to
below 4\%.}
\label{fig:permodel}
\end{figure}

\begin{figure}[t]
\centering
\includegraphics[width=\columnwidth]{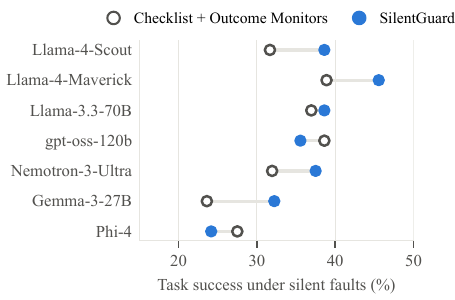}
\caption{ToolMaze task success under silent faults per model, for the checklist with Outcome Monitors
against \sgp{}. \sgp{} is ahead on 5 of 7 models.}
\label{fig:tmpermodel}
\end{figure}

\begin{figure}[t]
\centering
\includegraphics[width=\columnwidth]{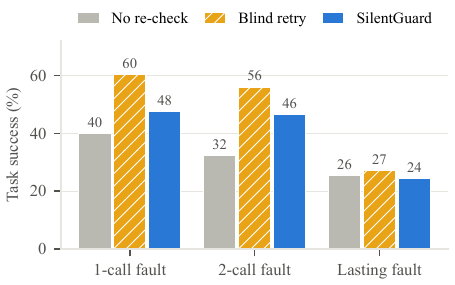}
\caption{ToolMaze task success by fault type for no re-check, blind retry and \sgp{}. Blind retry leads
while the fault clears and not when it lasts.}
\label{fig:retry}
\end{figure}

Table~\ref{tab:app-aob-full} gives every method on AgentOrchBench by fault type, with correct and
unknown shares, and Figure~\ref{fig:outcomes} shows the same outcomes. Table~\ref{tab:app-ci} gives
the confidence intervals of the main tables. Tables~\ref{tab:app-permodel} and~\ref{tab:permodel} and
Figures~\ref{fig:permodel} and~\ref{fig:tmpermodel} give the results per model, including the partial
runs of three older GPT models. Table~\ref{tab:app-toolmaze} covers all ToolMaze modes, and
Tables~\ref{tab:retry} and~\ref{tab:app-twocall} with Figure~\ref{fig:retry} compare blind retry with
the re-check. Table~\ref{tab:app-realapi} gives the real-API agent study with each check alone, and
Table~\ref{tab:ablation} the harness actions.

\paragraph{Withholding against Enforce.} \sgp{} and Enforce both accept only checked counts. Enforce leaves the failed
count in view, and agents then keep re-querying until some other query passes the check and report
its count. Of Enforce's \num{eg-enf-wrong} wrong answers, \num{eg-enf-requery} follow this pattern,
against \num{eg-sgp-requery} of \num{eg-sgp-wrong} for \sgp{}, which removes the value.

\begin{table*}[t]
\centering
\footnotesize
\setlength{\tabcolsep}{2.7pt}
\begin{tabular}{@{}lcccccccccc@{}}
\toprule
\rowcolor{head}
 & \multicolumn{6}{c}{\textbf{Wrong answers by fault type}$\downarrow$} & \multicolumn{4}{c}{\textbf{All faults}}\\
\rowcolor{head}\cmidrule(lr){2-7}\cmidrule(l){8-11}
\rowcolor{head} & Drop & Cap & AND & Ignore & Ignore & Off by & & & & \\
\rowcolor{head}\textbf{Method} & last & total & $\to$OR & field & NOT & one & Wrong$\downarrow$ & Correct$\uparrow$ & Unknown & Clean$\uparrow$\\
\midrule
No guard & 89.7 & 92.9 & 89.3 & 77.4 & 6.2 & 88.9 & 74.0\se{7.1} & 16.4\se{1.8} & 9.6\se{8.7} & 84.9\se{10.6}\\
Caution prompt & 88.3 & 91.3 & 87.7 & 75.4 & 4.6 & 92.9 & 73.3\se{7.3} & 16.4\se{2.2} & 10.3\se{9.3} & 86.1\se{9.1}\\
LLM judge & 88.7 & 91.7 & 84.7 & 73.0 & 6.2 & 82.9 & 71.2\se{7.8} & 17.4\se{2.1} & 11.4\se{9.7} & 86.5\se{8.5}\\
Checklist & 87.7 & 90.7 & 87.7 & 73.4 & 4.8 & 89.3 & 72.3\se{8.1} & 16.2\se{2.2} & 11.5\se{10.3} & 84.1\se{10.6}\\
Reviewer & 91.9 & 88.3 & 88.1 & 76.0 & 8.5 & 90.1 & 73.8\se{5.7} & 15.8\se{1.8} & 10.4\se{7.2} & 80.4\se{10.0}\\
Outcome Monitors & 89.3 & 90.3 & 87.1 & 72.0 & 6.3 & 87.9 & 72.2\se{8.1} & 16.6\se{1.6} & 11.2\se{9.5} & 86.1\se{8.8}\\
Blind retry & 92.3 & 91.3 & 91.7 & 76.0 & 3.4 & 92.1 & 74.4\se{5.1} & 17.3\se{1.4} & 8.3\se{6.5} & 88.3\se{7.4}\\
Re-query & 91.1 & 92.3 & 91.5 & 79.4 & 3.4 & 91.7 & 74.9\se{5.0} & 17.0\se{1.8} & 8.1\se{6.8} & 84.7\se{8.6}\\
Self-verification & 71.8 & 72.8 & 52.8 & 63.9 & 5.2 & 70.8 & 56.2\se{9.2} & 14.2\se{2.0} & 29.6\se{11.1} & 76.6\se{12.8}\\
Warning (last condition) & 71.8 & 74.0 & 66.9 & 68.8 & 33.5 & 63.7 & 63.1\se{9.7} & 11.4\se{2.4} & 25.4\se{10.2} & 85.5\se{8.9}\\
Warning + tip & 45.6 & 56.3 & 29.8 & 41.7 & 7.9 & 42.1 & 37.2\se{9.7} & 9.8\se{1.9} & 52.9\se{11.3} & 82.5\se{11.5}\\
Auto-repair & 1.2 & 3.2 & 2.6 & 4.0 & 2.4 & 4.6 & 3.0\se{1.2} & 57.3\se{6.6} & 39.7\se{6.4} & 83.3\se{9.7}\\
Answer gate (last cond.) & 32.3 & 27.4 & 32.3 & 56.5 & 25.6 & 23.2 & 32.9\se{10.6} & 5.2\se{1.8} & 61.9\se{10.7} & 84.9\se{8.6}\\
Enforce & 7.1 & 15.5 & 4.6 & 7.1 & 9.5 & 8.9 & 8.8\se{3.1} & 3.0\se{0.4} & 88.2\se{3.2} & 86.3\se{8.4}\\
Judge + blocking & 15.7 & 5.6 & 4.2 & 3.0 & 2.8 & 5.2 & 6.1\se{2.2} & 3.5\se{0.4} & 90.4\se{2.1} & 81.7\se{9.3}\\
OM + blocking & 87.3 & 86.1 & 89.7 & 73.2 & 3.6 & 85.5 & 70.9\se{8.5} & 16.6\se{1.8} & 12.5\se{10.2} & 86.5\se{9.4}\\
Judge, no evidence, + blocking & 87.9 & 86.7 & 87.5 & 73.4 & 3.0 & 86.3 & 70.8\se{8.2} & 16.7\se{1.8} & 12.5\se{10.0} & 83.7\se{10.6}\\
Judge, no evidence & 86.5 & 86.9 & 87.5 & 73.0 & 4.4 & 92.1 & 71.7\se{8.7} & 16.3\se{2.1} & 11.9\se{10.7} & 80.7\se{12.3}\\
Warning (last cond.) + may abstain & 6.2 & 4.2 & 8.3 & 44.8 & 4.0 & 4.8 & 12.1\se{2.1} & 3.8\se{0.6} & 84.2\se{2.2} & 83.9\se{12.2}\\
Warning & 82.7 & 83.9 & 80.8 & 73.2 & 19.2 & 80.6 & 70.1\se{8.7} & 14.1\se{2.2} & 15.9\se{10.3} & 86.3\se{8.5}\\
Warning + may abstain & 23.8 & 23.6 & 16.3 & 17.7 & 6.5 & 21.2 & 18.2\se{9.3} & 5.0\se{2.0} & 76.8\se{11.1} & 83.5\se{11.2}\\
\rowcolor{ours}\textbf{SilentGuard} & 1.6 & 2.0 & 1.4 & 2.6 & 1.4 & 1.4 & 1.7\se{0.5} & 2.4\se{0.3} & 95.8\se{0.7} & 84.9\se{11.3}\\
\bottomrule
\end{tabular}
\caption{AgentOrchBench, all methods (mean over 7 open models, \%, $\pm$ standard error). Wrong, correct and unknown are shares of the runs with a silent fault. Clean is \% correct with no fault. Auto-repair has the most correct answers but its wrong answers are released as verified (Table~\ref{tab:repair}). Last condition: the earlier check of the last condition only. All other harness rows use the full check. No evidence: the judge does not see the consistency queries. OM: Outcome Monitors.}
\label{tab:app-aob-full}
\end{table*}

\begin{table*}[t]
\centering
\footnotesize
\setlength{\tabcolsep}{8pt}
\begin{tabular}{@{}lcccc@{}}
\toprule
\rowcolor{head}
\textbf{AgentOrchBench} & Wrong$\downarrow$ & Correct$\uparrow$ & Unknown & Clean$\uparrow$\\
\midrule
No guard & [56.8, 91.3] & [11.9, 20.9] & [0.0, 30.8] & [59.1, 100.0]\\
Caution prompt & [55.5, 91.2] & [11.1, 21.6] & [0.0, 33.1] & [63.9, 100.0]\\
Self-verification & [33.6, 78.8] & [9.2, 19.2] & [2.4, 56.7] & [45.4, 100.0]\\
Re-query & [62.6, 87.1] & [12.6, 21.4] & [0.0, 24.7] & [63.6, 100.0]\\
Blind retry & [62.0, 86.9] & [13.9, 20.7] & [0.0, 24.1] & [70.1, 100.0]\\
LLM judge & [52.1, 90.3] & [12.2, 22.5] & [0.0, 35.2] & [65.8, 100.0]\\
Checklist & [52.3, 92.2] & [10.9, 21.6] & [0.0, 36.6] & [58.2, 100.0]\\
Reviewer & [60.0, 87.6] & [11.5, 20.2] & [0.0, 28.0] & [56.0, 100.0]\\
Outcome Monitors & [52.3, 92.1] & [12.6, 20.7] & [0.0, 34.5] & [64.5, 100.0]\\
Judge, no evidence & [50.4, 93.0] & [11.2, 21.4] & [0.0, 38.1] & [50.8, 100.0]\\
Judge, no evidence, + blocking & [50.6, 90.9] & [12.2, 21.2] & [0.0, 37.0] & [57.7, 100.0]\\
Judge + blocking & [0.6, 11.5] & [2.6, 4.4] & [85.2, 95.7] & [59.0, 100.0]\\
OM + blocking & [50.2, 91.6] & [12.3, 20.9] & [0.0, 37.3] & [63.6, 100.0]\\
Warning + may abstain & [0.0, 41.0] & [0.2, 9.8] & [49.6, 100.0] & [56.2, 100.0]\\
\rowcolor{ours}\textbf{SilentGuard (ours)} & [0.4, 3.0] & [1.6, 3.3] & [94.2, 97.5] & [57.2, 100.0]\\
\midrule
\rowcolor{head}\textbf{Retail domain} & Wrong$\downarrow$ & Correct$\uparrow$ & Clean$\uparrow$ & \\
\midrule
No guard & [54.3, 100.0] & [1.3, 13.1] & [8.1, 64.3] & \\
Auto-repair & [5.2, 24.8] & [5.9, 55.0] & [6.9, 64.6] & \\
Enforce & [2.9, 21.6] & [1.0, 12.1] & [5.8, 61.9] & \\
\rowcolor{ours}\textbf{SilentGuard (ours)} & [4.1, 17.8] & [2.3, 11.9] & [10.3, 66.6] & \\
\midrule
\rowcolor{head}\textbf{Real APIs (9)} & Wrong$\downarrow$ & Correct$\uparrow$ & Right docs$\uparrow$ & \\
\midrule
No guard & [76.9, 100.0] & [0.0, 10.4] & [76.2, 100.0] & \\
Blind retry & [75.3, 100.0] & [0.0, 9.3] & [79.5, 100.0] & \\
Spec check alone & [9.8, 32.1] & [54.0, 87.6] & [76.2, 100.0] & \\
Typed calls & [17.3, 39.9] & [52.5, 80.2] & [77.7, 100.0] & \\
Typed calls + ours & [15.9, 40.6] & [51.9, 78.9] & [80.4, 100.0] & \\
\rowcolor{ours}\textbf{SilentGuard (ours)} & [12.1, 30.5] & [51.2, 87.9] & [74.6, 100.0] & \\
\midrule
\rowcolor{head}\textbf{ToolMaze} & Silent$\uparrow$ & Loud$\uparrow$ & Clean$\uparrow$ & \\
\midrule
Checklist & [15.1, 31.3] & [28.0, 48.6] & [45.5, 65.6] & \\
Outcome Monitors & [27.3, 33.7] & [35.2, 46.2] & [48.3, 66.0] & \\
Checklist + OM & [27.4, 38.1] & [33.3, 46.2] & [44.7, 64.9] & \\
Blind retry & [39.4, 48.0] & -- & [47.8, 60.6] & \\
\rowcolor{ours}\textbf{SilentGuard (ours)} & [29.9, 42.1] & [32.7, 46.9] & [42.1, 62.9] & \\
\bottomrule
\end{tabular}
\caption{95\% confidence intervals for Tables~\ref{tab:main} and~\ref{tab:other} ($t$ interval over models, clipped to [0, 100]).}
\label{tab:app-ci}
\end{table*}

\begin{table*}[t]
\centering
\small
\setlength{\tabcolsep}{5pt}
\begin{tabular}{@{}lccccccccc@{}}
\toprule
\rowcolor{head}
\textbf{Model} & None & Check. & Rev. & OM & Judge & Retry & Self-v. & Enf. & \textbf{Ours}\\
\midrule
Llama-4-Scout & 86.6 & 79.2 & 79.6 & 81.2 & 83.6 & 78.0 & 79.6 & \underline{7.2} & \textbf{2.1}\\
Llama-4-Maverick & 80.8 & 80.3 & 79.9 & 79.6 & 80.1 & 80.1 & \underline{36.3} & \textbf{0.5} & \textbf{0.5}\\
Llama-3.3-70B & 80.6 & 80.8 & 80.6 & 80.6 & 78.7 & 80.3 & 69.2 & \underline{1.2} & \textbf{0.0}\\
gpt-oss-120b & 80.8 & 80.6 & 79.8 & 80.8 & 74.1 & 80.8 & 53.5 & \underline{24.8} & \textbf{1.2}\\
Nemotron-3-Ultra & 80.6 & 80.6 & 80.6 & 80.6 & 79.9 & 80.6 & 78.5 & \underline{12.0} & \textbf{3.7}\\
Gemma-3-27B & 76.9 & 81.0 & 76.2 & 78.9 & 77.3 & 77.3 & 63.9 & \underline{6.5} & \textbf{3.5}\\
Phi-4 & 32.2 & 23.4 & 40.0 & 23.4 & 24.8 & 44.0 & 12.5 & \underline{9.5} & \textbf{1.2}\\
\midrule
GPT-4.1 & 100.0\,{\scriptsize(14\%)} & -- & -- & -- & 66.7\,{\scriptsize(57\%)} & -- & 47.2\,{\scriptsize(29\%)} & -- & --\\
GPT-4o & 83.3\,{\scriptsize(14\%)} & -- & -- & -- & -- & -- & 2.8\,{\scriptsize(14\%)} & -- & --\\
GPT-4o-mini & 98.6\,{\scriptsize(14\%)} & -- & -- & -- & 98.6\,{\scriptsize(43\%)} & -- & -- & 5.6\,{\scriptsize(14\%)} & --\\
\bottomrule
\end{tabular}
\caption{AgentOrchBench, wrong answers per model and method (\%). None: no guard. Check.: checklist. Rev.: reviewer. OM: Outcome Monitors. Judge: LLM judge. Retry: blind retry. Self-v.: self-verification. Enf.: Enforce. Ours: SilentGuard. Bold: best in the row. Underlined: second. The three older GPT models below the line finished only part of their runs (finished share in brackets) and are not ranked. The GPT-5.6 models are in Table~\ref{tab:frontier}.}
\label{tab:app-permodel}
\end{table*}

\begin{table}[t]
\centering
\small
\setlength{\tabcolsep}{3pt}
\resizebox{\columnwidth}{!}{%
\begin{tabular}{@{}lcccccc@{}}
\toprule
\rowcolor{head}
 & \multicolumn{3}{c}{\textbf{AgentOrchBench} (wrong $\downarrow$)} & \multicolumn{3}{c}{\textbf{ToolMaze} (success $\uparrow$)}\\
\rowcolor{head}\cmidrule(lr){2-4}\cmidrule(l){5-7}
\rowcolor{head}\textbf{Model} & None & Ours & $\Delta$ & C+OM & Ours & $\Delta$\\
\midrule
Llama-4-Scout & 86.6 & \textbf{2.1} & \textcolor{gain}{$-84.5$} & 31.7 & \textbf{38.6} & \textcolor{gain}{$+6.9$}\\
Llama-4-Maverick & 80.8 & \textbf{0.5} & \textcolor{gain}{$-80.3$} & 38.9 & \textbf{45.6} & \textcolor{gain}{$+6.7$}\\
Llama-3.3-70B & 80.6 & \textbf{0.0} & \textcolor{gain}{$-80.6$} & 36.9 & \textbf{38.6} & \textcolor{gain}{$+1.7$}\\
gpt-oss-120b & 80.8 & \textbf{1.2} & \textcolor{gain}{$-79.6$} & \textbf{38.6} & 35.6 & \textcolor{loss}{$-3.1$}\\
Nemotron-3-Ultra & 80.6 & \textbf{3.7} & \textcolor{gain}{$-76.9$} & 31.9 & \textbf{37.5} & \textcolor{gain}{$+5.6$}\\
Gemma-3-27B & 76.9 & \textbf{3.5} & \textcolor{gain}{$-73.4$} & 23.6 & \textbf{32.2} & \textcolor{gain}{$+8.6$}\\
Phi-4 & 32.2 & \textbf{1.2} & \textcolor{gain}{$-31.0$} & \textbf{27.5} & 24.2 & \textcolor{loss}{$-3.3$}\\
\midrule
\textit{Ours better} & \multicolumn{3}{c}{7/7, $p=0.016$} & \multicolumn{3}{c}{5/7, $p=0.125$}\\
\bottomrule
\end{tabular}}
\caption{\textbf{The gain holds for every model on AgentOrchBench and for 5 of 7 on ToolMaze.} AgentOrchBench: \% wrong answers under silent faults, no guard against SilentGuard. ToolMaze: \% task success under silent faults, the strongest published mix (Checklist + Outcome Monitors, C+OM) against SilentGuard. $\Delta$: ours minus the other, computed before rounding (green: ours better, red: ours worse, bold: the better of the pair). $p$: exact sign-flip test over the 7 models.}
\label{tab:permodel}
\end{table}

\begin{table}[h]
\centering
\small
\setlength{\tabcolsep}{3pt}
\resizebox{\columnwidth}{!}{%
\begin{tabular}{@{}lccccc@{}}
\toprule
\rowcolor{head}
\textbf{Method} & P3$\uparrow$ & P4$\uparrow$ & P3+P4$\uparrow$ & P0$\uparrow$ & P1+P2$\uparrow$\\
\midrule
Checklist & 26.8\se{4.3} & 19.5\se{2.6} & 23.2\se{3.3} & 55.5\se{4.1} & 38.3\se{4.2}\\
Outcome Monitors & 36.3\se{2.2} & 24.8\se{1.6} & 30.5\se{1.3} & 57.1\se{3.6} & 40.7\se{2.2}\\
Checklist + OM & 39.9\se{3.1} & 25.6\se{2.0} & 32.7\se{2.2} & 54.8\se{4.1} & 39.8\se{2.6}\\
Blind retry & 60.4\se{3.1} & 27.0\se{1.4} & 43.7\se{1.7} & 54.2\se{2.6} & --\\
\rowcolor{ours}\textbf{SilentGuard} & 47.6\se{3.7} & 24.4\se{1.5} & 36.0\se{2.5} & 52.5\se{4.2} & 39.8\se{2.9}\\
\bottomrule
\end{tabular}}
\caption{ToolMaze, all modes (\% task success by its own judge, mean $\pm$ SE over 7 models, 180 fresh workflows). P3: silent fault on the first call. P4: silent fault on every call. P0: no fault. P1, P2: explicit errors. Blind retry was not run on P1/P2.}
\label{tab:app-toolmaze}
\end{table}

\begin{table}[t]
\centering
\small
\setlength{\tabcolsep}{2.6pt}
\resizebox{\columnwidth}{!}{%
\begin{tabular}{@{}lccccc@{}}
\toprule
\rowcolor{head}
 & \multicolumn{3}{c}{\textbf{ToolMaze} success$\uparrow$} & \textbf{Real APIs} & \textbf{Repeated}\\
\rowcolor{head}\cmidrule(lr){2-4}
\rowcolor{head}\textbf{Harness} & 1-call fault & 2-call fault & Lasting & wrong$\downarrow$ & actions$\downarrow$\\
\midrule
No re-check & 39.9 & 32.5 & \underline{25.6} & -- & --\\
Blind retry & \textbf{60.4} & \textbf{56.0} & \textbf{27.0} & 89.5 & 2.36\\
\rowcolor{ours}\textbf{SilentGuard} & \underline{47.6} & \underline{46.4} & 24.4 & \textbf{21.3} & \textbf{0.24}\\
\bottomrule
\end{tabular}}
\caption{\textbf{Blind retry helps only when the fault clears.} Blind retry (the no re-check harness plus every call sent twice, second result kept) clears a short fault that lasts one or two calls. On a fault that lasts (ToolMaze P4) it adds little over the same harness without retry, and on real APIs, which ignore a misnamed filter every time, it does not reduce wrong answers. It also repeats bookings, payments and messages. No re-check: the checklist with Outcome Monitors, without retry. Mean over 7 open models. Values are \% except repeated actions (repeated action calls per ToolMaze run). Bold: best. Underlined: second.}
\label{tab:retry}
\end{table}

\begin{table}[h]
\centering
\small
\setlength{\tabcolsep}{4pt}
\resizebox{\columnwidth}{!}{%
\begin{tabular}{@{}lccc@{}}
\toprule
\rowcolor{head}
\textbf{Model} & \textbf{No re-check} & \textbf{Blind retry} & \textbf{SilentGuard}\\
\midrule
Llama-4-Scout & 32.8 & 60.0 & 50.6\\
Llama-4-Maverick & 31.7 & 56.1 & 47.2\\
Llama-3.3-70B & 32.2 & 66.1 & 57.2\\
gpt-oss-120b & 36.4 & 51.2 & 43.4\\
Nemotron-3-Ultra & 38.9 & 65.6 & 53.9\\
Gemma-3-27B & 12.2 & 34.4 & 22.8\\
Phi-4 & 43.3 & 58.9 & 50.0\\
\bottomrule
\end{tabular}}
\caption{ToolMaze short fault lasting two calls, per model (\% task success, 180 workflows, runs whose model calls failed at the provider are left out). No re-check: the checklist with Outcome Monitors.}
\label{tab:app-twocall}
\end{table}

\begin{table}[h]
\centering
\small
\setlength{\tabcolsep}{3pt}
\resizebox{\columnwidth}{!}{%
\begin{tabular}{@{}lcccc@{}}
\toprule
\rowcolor{head}
\textbf{Method} & Wrong$\downarrow$ & Correct$\uparrow$ & Unknown & Right docs$\uparrow$\\
\midrule
No guard & 89.8\se{5.3} & 3.5\se{2.8} & 6.0\se{4.6} & 92.1\se{6.5}\\
Blind retry & 89.5\se{5.8} & 3.2\se{2.5} & 4.8\se{3.7} & 93.0\se{5.5}\\
Value-swap check alone & 17.8\se{2.5} & 25.7\se{11.4} & 55.2\se{12.8} & 91.1\se{7.8}\\
Spec check alone & 21.0\se{4.6} & 70.8\se{6.8} & 7.0\se{4.6} & 92.1\se{6.5}\\
\rowcolor{ours}\textbf{SilentGuard} & 21.3\se{3.8} & 69.5\se{7.5} & 7.9\se{4.9} & 92.4\se{7.3}\\
\bottomrule
\end{tabular}}
\caption{Agents on nine real public APIs, mean $\pm$ standard error over 7 open models (45 runs per model, method and documentation). Wrong, correct, unknown: \% of runs with slightly wrong documentation (the rest ran out of turns with no answer). Right docs: \% correct with the right documentation. Value-swap check alone: the harness withholds a count whose filter had no effect, with no spec check. It leaves the fewest wrong answers but few correct ones. The spec check does most of the work, because it rejects the misnamed filter and the agent rewrites the query.}
\label{tab:app-realapi}
\end{table}

\begin{table*}[t]
\centering
\small
\setlength{\tabcolsep}{7pt}
\begin{tabular}{@{}lcccccc@{}}
\toprule
\rowcolor{head}
 & \multicolumn{4}{c}{\textbf{AgentOrchBench}} & \multicolumn{2}{c}{\textbf{Retail}}\\
\rowcolor{head}\cmidrule(lr){2-5}\cmidrule(l){6-7}
\rowcolor{head}\textbf{Harness action} & Wrong$\downarrow$ & Correct$\uparrow$ & Unknown & Clean$\uparrow$ & Wrong$\downarrow$ & Clean$\uparrow$\\
\midrule
Warning (last condition) & 63.1 & 11.4 & 25.4 & \underline{85.5} & 71.5 & 35.7\\
Warning & 70.1 & \underline{14.1} & 15.9 & \textbf{86.3} & -- & --\\
Warning + tip & 37.2 & 9.8 & 52.9 & 82.5 & 59.6 & \underline{36.9}\\
Auto-repair & \underline{3.0} & \textbf{57.3} & 39.7 & 83.3 & 15.0 & 35.7\\
Answer gate (last condition) & 32.9 & 5.2 & 61.9 & 84.9 & -- & --\\
Enforce & 8.8 & 3.0 & 88.2 & \textbf{86.3} & \underline{12.3} & 33.8\\
\rowcolor{ours}\textbf{SilentGuard} & \textbf{1.7} & 2.4 & 95.8 & 84.9 & \textbf{11.0} & \textbf{38.5}\\
\bottomrule
\end{tabular}
\caption{\textbf{Ablation of the harness action after a failed check} (mean over 7 open models, \%). Warnings keep wrong answers high. Auto-repair gives the most correct answers but releases wrong ones marked as verified (Table~\ref{tab:repair}). SilentGuard withholds only what an exact check proves wrong. Last condition: these two earlier arms check the last condition only. All other rows use the full check. Warning + tip: the warning also names the recovery tools. Enforce: only checked counts are accepted. Unknown: the answer says the value cannot be determined. --: not run on retail. Bold: best. Underlined: second.}
\label{tab:ablation}
\end{table*}

\section{Predictions and Method History}
\label{app:prereg}
\label{app:history}

Every planned experiment was written down with its predictions before it ran, and the code was
frozen by file hashes, so runs refuse to start if the code changed. The prediction log and the hash files
are in the code repository. We kept every result, including failed predictions. Of the \num{prereg-n} predictions of the planned experiments, \num{prereg-held} held
(Table~\ref{tab:app-prereg}). The count leaves out the earlier ToolMaze design studies, which shaped the
method, and $\tau$-bench (App.~\ref{app:tau}). Four failed predictions were about blind retry on ToolMaze
(\S\ref{sec:res-retry}). Two failed because \sgp{} did better than predicted (C1 on
counts, and L1 on lists for the 6 models it was written for, while on all 7 models its gaps of
\num{l1-gap7-seen} and \num{l1-gap7-unseen} points are within the predicted 3), one because the LLM judge with blocking raised fewer false alarms than we expected
(K3), and two because typed tool calls did worse than the spec check and adding \sgp{} to them did not
help (T1, T3). Three were about permission to abstain. With the
last-condition check it worked better than we predicted (X3), and with the full check worse (Y1, Y2).
One was about recovery in the value-format study (V4).

The method had three versions. The first withheld a flagged result at once. On ToolMaze it beat
Outcome Monitors on average but not on every model. The second added the harness re-check, which
raised success on one-time faults but lost when withholding blocked good data after a false alarm of
a learned check. The final \sgp{} sets the action from the check's estimated false-alarm rate. Each
version was tested only on ToolMaze workflows it had never seen, and \sgp{} was frozen before its
planned runs on AgentOrchBench, the retail domain, real APIs and the controls. The 1\% cutoff
$\tau$ was chosen after the second version, which we list as a limitation. An earlier pilot sweep,
with a stricter grader and other fault settings, led to the answer rule. Its numbers are not
comparable with Table~\ref{tab:main} and are kept in our run logs. The frontier-model runs were
added last and had no predictions.

\section{List Answers}
\label{app:lists}
\label{sec:res-records}

To test a second kind of answer, the 30 list tasks ask for the IDs of all matching records instead of
a count (Table~\ref{tab:records}), using the record checks of Table~\ref{tab:checks}. With
\sgp{}, wrong
lists fall from \pct{rec-none-seen-wrong} to \pct{rec-full-seen-wrong} on fault types seen in design
and from \pct{rec-none-unseen-wrong} to \pct{rec-full-unseen-wrong} on five fault types written
later, the fewest of any harness action, and clean-task accuracy is \pct{rec-full-clean} (\pct{rec-none-clean} with no
guard). Correct lists fall too, from \pct{rec-none-unseen-correct} to \pct{rec-full-unseen-correct} on
unseen faults. Auto-repair, which keeps only the
returned records that meet the filter, has more wrong lists (\pct{rec-rep-unseen-wrong} on unseen
faults) but more correct ones. Record checks can remove a record that should not be there, but they
cannot reveal one the API never returned.

\begin{table}[t]
\centering
\small
\setlength{\tabcolsep}{2.3pt}
\begin{tabular}{@{}lccccc@{}}
\toprule
\rowcolor{head}
 & \multicolumn{2}{c}{\textbf{Seen faults}} & \multicolumn{2}{c}{\textbf{Unseen faults}} & \textbf{Clean}\\
\rowcolor{head}\cmidrule(lr){2-3}\cmidrule(lr){4-5}\cmidrule(l){6-6}
\rowcolor{head}\textbf{Method} & Wrong$\downarrow$ & Corr.$\uparrow$ & Wrong$\downarrow$ & Corr.$\uparrow$ & Corr.$\uparrow$\\
\midrule
No guard & 49.3 & \underline{43.7} & 37.8 & \textbf{56.7} & 80.5\\
Auto-repair & 14.6 & \textbf{47.9} & 16.9 & \underline{37.5} & \underline{81.0}\\
Checked lists only & \underline{11.2} & 22.1 & \underline{11.7} & 7.5 & \underline{81.0}\\
\rowcolor{ours}\textbf{SilentGuard} & \textbf{9.1} & 21.1 & \textbf{11.5} & 7.1 & \textbf{84.8}\\
\bottomrule
\end{tabular}
\caption{\textbf{The same rule gives the fewest wrong lists.} The 30 tasks ask for the IDs of all matching records. Six fault types were seen in design and five were written later (unseen). The record checks test that every returned record meets the filter, that no record repeats, and that the list matches the reported total and the count identity. Checked lists only: the answer rule alone. SilentGuard adds the re-check, withholding and the result checklist. Corr.: correct. Bold: best. Underlined: second. Mean over 7 open models, \% of runs.}
\label{tab:records}
\end{table}

\section{Cost and Fine-Tuning}
\label{app:cost}

This section gives the cost of each method and the fine-tuning study.

\paragraph{Cost.}
\sgp{} takes \num{cost-sgp-secs}\,s and \num{cost-sgp-calls} tool calls per faulted run, against
\num{cost-none-secs}\,s for no guard (Table~\ref{tab:app-cost}). The extra time is the harness's own
check queries, and no extra model call is made. \sgp{} takes less time than the reviewer (\num{cost-reviewer-secs}\,s) and the LLM judge
(\num{cost-judge-secs}\,s), which add model calls, but uses more tokens than the judge
(\num{cost-sgp-tokens} against \num{cost-judge-tokens}), because the agent takes more turns.
The check can be made cheaper. Checking conditions in random order and stopping at the first
failure catches as many faults (\pct{bud-seq-caught}) with \num{bud-seq-calls} extra queries on
faulted results (Table~\ref{tab:app-budget}).

\paragraph{Fine-tuning.}
We fine-tuned two small models with the harness check shown to the model as a warning
(Table~\ref{tab:app-training}). These are Phi-4-mini \citep{abouelenin2025phi4mini} with LoRA
\citep{hu2021lora} supervised fine-tuning (SFT) and with reinforcement learning (RL) using DAPO
\citep{yu2025dapo}, and Mistral-7B \citep{jiang2023mistral} with QLoRA \citep{dettmers2023qlora} SFT. Training to
refuse transfers to unseen fault types, but the trained model still acts on the harness check it was
trained with. Training makes the model follow the check and does not replace it. Training to repair
does not transfer, because on unseen faults it turns refusals into confident wrong answers.

\begin{table}[h]
\centering
\footnotesize
\setlength{\tabcolsep}{2.5pt}
\begin{tabular}{@{}lrrrr@{}}
\toprule
\rowcolor{head}
 & \textbf{Tool} & & \textbf{Time} & \textbf{Extra LLM}\\
\rowcolor{head}\textbf{Method} & \textbf{calls} & \textbf{Tokens} & \textbf{(s)} & \textbf{calls}\\
\midrule
No guard & 2.0 & 3{,}183 & 9.2 & 0.0\\
Caution prompt & 2.0 & 3{,}734 & 11.3 & 0.0\\
LLM judge & 3.2 & 6{,}287 & 47.2 & 2.0\\
Checklist & 1.9 & 3{,}429 & 8.7 & 0.0\\
Reviewer & 1.9 & 9{,}276 & 25.0 & 2.6\\
Outcome Monitors & 1.9 & 3{,}058 & 8.8 & 0.0\\
Blind retry & 1.9 & 3{,}052 & 8.0 & 0.0\\
Re-query & 1.8 & 3{,}041 & 7.9 & 0.0\\
Self-verification & 5.0 & 12{,}422 & 33.3 & 0.0\\
Warning (last condition) & 4.7 & 9{,}620 & 23.5 & 0.0\\
Warning + tip & 5.6 & 13{,}831 & 30.6 & 0.0\\
Auto-repair & 2.9 & 5{,}782 & 16.6 & 0.0\\
Answer gate (last cond.) & 5.0 & 10{,}963 & 27.4 & 0.0\\
Enforce & 5.0 & 11{,}745 & 29.1 & 0.0\\
Warning & 4.4 & 8{,}299 & 26.9 & 0.0\\
\rowcolor{ours}\textbf{SilentGuard} & 3.8 & 9{,}055 & 18.9 & 0.0\\
\bottomrule
\end{tabular}
\caption{\textbf{Cost per faulted run} on AgentOrchBench (mean over 7 open models). Columns give agent tool calls, LLM tokens, wall time and extra LLM calls (judge or reviewer) on top of the agent's own. SilentGuard's checks run in the harness and need no extra LLM call.}
\label{tab:app-cost}
\end{table}

\begin{table*}[h]
\centering
\small
\setlength{\tabcolsep}{4pt}
\resizebox{\textwidth}{!}{%
\begin{tabular}{@{}llccccc|ccc@{}}
\toprule
\rowcolor{head}
 & & \multicolumn{5}{c|}{Greedy decoding} & \multicolumn{3}{c}{Sampling, $T=0.7$, 3 seeds}\\
\rowcolor{head}\textbf{Model} & \textbf{Training} & Seen$\downarrow$ & Unseen$\downarrow$ & Consistent lie$\downarrow$ & Reworded$\downarrow$ & Clean$\uparrow$ & Seen$\downarrow$ & Unseen$\downarrow$ & Clean$\uparrow$\\
\midrule
Phi-4-mini & Base model & 100.0 & 73.3 & 100.0 & 100.0 & 86.7 & 57.8 & 72.8 & 25.6\\
Phi-4-mini & SFT to refuse (trained on drop-last) & 0.0 & 6.7 & 100.0 & 0.0 & 100.0 & 0.0 & 6.1 & 100.0\\
Phi-4-mini & SFT to repair & 0.0 & 45.3 & 100.0 & 0.0 & 100.0 & 1.1 & 90.0 & 100.0\\
Phi-4-mini & RL (DAPO, false-equation penalty) & 0.0 & 28.7 & 100.0 & 0.0 & 100.0 & 0.0 & 59.4 & 100.0\\
Mistral-7B & Base model & 0.0 & 0.0 & 0.0 & 0.0 & 0.0 & 2.2 & 1.1 & 0.0\\
Mistral-7B & SFT to refuse (trained on drop-last) & 0.0 & 8.7 & 100.0 & 0.0 & 100.0 & 3.3 & 0.6 & 100.0\\
Mistral-7B & SFT to repair & 0.0 & 42.0 & 100.0 & 0.0 & 100.0 & 0.0 & 81.7 & 100.0\\
\bottomrule
\end{tabular}}
\caption{Training the model itself, full results (\% wrong, clean: \% correct). Seen: drop-last, the fault used in training. Unseen: cap-total, AND$\to$OR, ignore-NOT, off-by-one and prefix. Consistent lie: a wrong answer the API backs up in every query. Reworded: the seen fault with reworded tasks. With sampling, unseen covers AND$\to$OR and off-by-one only. SFT: supervised fine-tuning. RL: reinforcement learning with DAPO. Mistral-7B base cannot use the tools, so its 0\% wrong is not a success.}
\label{tab:app-training}
\end{table*}

\section{Check Budget, Cutoff and a Check-Aware API}
\label{app:budget}

This section tests cheaper versions of the count check, other cutoffs and an API built to pass the check.

\begin{table}[t]
\centering
\small
\setlength{\tabcolsep}{4pt}
\begin{tabular}{@{}cccc@{}}
\toprule
\rowcolor{head}
\textbf{Cutoff} $\tau$ & \textbf{Min.\ support} & \textbf{Withheld (\%)} & \textbf{Changed}\\
\midrule
0.1\% & $\geq$998 & 0.0 & 0\\
0.5\% & $\geq$198 & 0.0 & 0\\
1\% & $\geq$98 & 0.0 & 0\\
2\% & $\geq$48 & 4.4 & 108\\
5\% & $\geq$18 & 34.6 & 847\\
10\% & $\geq$8 & 73.0 & 1{,}788\\
20\% & $\geq$3 & 87.7 & 2{,}148\\
\bottomrule
\end{tabular}
\caption{Sensitivity to the cutoff on ToolMaze. A learned check is treated as exact when $1/(n+2)\le\tau$, that is, when it held on at least the listed number of clean examples. Over all 2{,}448 withhold-or-warn decisions logged in our ToolMaze runs, the largest support is 52, so every cutoff below 1.85\% (1/54) gives the same decisions as 1\% (Changed: decisions that differ from 1\%). On AgentOrchBench and real APIs every check is exact and the cutoff has no effect.}
\label{tab:app-gate}
\end{table}

\paragraph{Check budget.} One run of the full check gives the identity result at every position, so
the detection rate of checking only $b$ randomly chosen conditions can be computed exactly over all
$\binom{k}{b}$ choices, with no sampling. We ran the harness alone (no model) on the 36 count tasks
under each fault type and under no fault (Table~\ref{tab:app-budget}). No budget raised a false alarm
on a clean result. The original single check (last condition only) misses an ignored field at any
other position. One random condition catches it only \pct{bud-dropfield-r1} of the time, and the
full check \pct{bud-dropfield-all} (the rest are tasks where the ignored field does not change the
count).

\begin{table*}[t]
\centering
\small
\setlength{\tabcolsep}{7pt}
\begin{tabular}{@{}lccccc@{}}
\toprule
\rowcolor{head}
 & \multicolumn{2}{c}{\textbf{Extra queries}} & \multicolumn{2}{c}{\textbf{Seen faults}} & \textbf{Check-aware}\\
\rowcolor{head}\cmidrule(lr){2-3}\cmidrule(lr){4-5}
\rowcolor{head}\textbf{Positions checked} & clean & faulted & caught$\uparrow$ & wrong$\downarrow$ & caught$\uparrow$\\
\midrule
No check & 0.0 & 0.0 & 0.0 & 80.6 & 0.0\\
Last condition only & 2.0 & 2.0 & 88.0 & 8.8 & 0.0\\
1 random condition & 2.0 & 2.0 & 88.4 & 8.6 & 0.9\\
2 random conditions & 4.0 & 4.0 & 94.4 & 2.8 & 1.9\\
All, stop at first failure & 5.2 & 2.3 & 97.2 & 0.0 & 2.8\\
\rowcolor{ours}\textbf{All conditions (SilentGuard)} & 5.2 & 5.2 & 97.2 & 0.0 & 2.8\\
\bottomrule
\end{tabular}
\caption{How many conditions to check, measured on the harness alone (no model, 36 count tasks per fault type, detection averaged exactly over every choice of positions). Extra queries per filter call on clean and faulted runs. Caught: share of faulted results the check flags. Wrong: share released with a wrong count under the answer rule. No budget raised a false alarm on clean runs. Check-aware: the consistent lie, an API built to keep the identity true. No budget catches more than 2.8\% of it.}
\label{tab:app-budget}
\end{table*}

\paragraph{Cutoff.} Table~\ref{tab:app-gate} counts, over all \num{gate-decisions} withhold-or-warn decisions in
our ToolMaze runs, how many would change at other cutoffs.

\paragraph{Check-aware API.} The consistent-lie fault inflates every filtered count
and answers the checker's complement queries so that the identity still holds. The last-condition
check never catches it, and checking every position, or random positions, catches at most
\pct{bud-all-lie}. This limits what any check built only from the API's own answers can do.

\section{$\tau$-bench}
\label{app:tau}
We also ran $\tau$-bench retail \citep{yao2024tau} with two silent read faults on \texttt{get\_order\_details}. There,
\sgp{} tied the published methods, because most injected faults do not affect what the task needs. The
predictions for this run were written before the runs, are not counted in Table~\ref{tab:app-prereg}, and are kept in
our run logs.

\section{Further Discussion}
\label{app:faq}

This section answers points a reader may raise and that the main text covers only briefly.

\paragraph{Each fault type is modeled on a kind of API bug.} Table~\ref{tab:faults} lists them, and
three kinds occur in the real APIs we probed. These are an ignored parameter in \num{e3n-silent} of
\num{e3n-apis}, a silently capped page and an ignored filter value (Table~\ref{tab:app-faultkinds}).
The other fault types are not measured in real APIs. \sgp{} also has the fewest wrong answers on the
retail schema, cuts wrong answers on those real APIs, and is slightly ahead of the published methods
on ToolMaze silent faults, a benchmark built by others with different tasks and faults.

\paragraph{The method is not limited to counts.} It uses whatever checks a tool allows. These are an exact
identity for count queries, record checks for lists (App.~\ref{app:lists}), the API's own parameter
list for documented APIs, and learned contracts for everything else. Table~\ref{tab:checks} lists
exact checks for other common operations. On ToolMaze, whose workflows book, pay, message and look up
records, only learned contracts are available, and \sgp{} is slightly ahead of the published methods
on silent faults.

\paragraph{Answering unknown is not refusing.} Under a fault that repeats on every call, the API
cannot give the correct count, so the real choice is between a wrong number and saying so. On the
same tasks with no fault, \sgp{} is as accurate as no guard (\pct{aob-sgp-clean} against
\pct{aob-none-clean}). On real APIs, where the agent can rewrite the query, \sgp{} raises correct
answers from \pct{api-none-correct} to \pct{api-sgp-correct}.

\paragraph{Repair releases wrong answers as verified.} Auto-repair gives more correct answers on the
fault types it was built for, but on unseen fault types it released \num{ob-vbw} wrong answers marked
as verified (Table~\ref{tab:repair}). A wrong answer that says it is verified is worse than no
answer.

\paragraph{Blind retry wins on ToolMaze because half its faults clear.} Half of ToolMaze's silent
faults hit only the first call, so any retry clears them. On faults that last, and on real APIs,
retrying gives no gain, and it repeats bookings, payments and messages (Table~\ref{tab:retry}). We
predicted otherwise for the two-call fault and report the failure (Table~\ref{tab:app-prereg}).

\paragraph{A better prompt is not enough.} A warning from the count check plus explicit permission to
abstain cuts wrong answers to \pct{x6-abst}, against \pct{x6-sgp} for \sgp{} on the same models
(Table~\ref{tab:twobytwo}). But the prompt still needs the check's evidence, wrong answers still range
from \num{x6-abst-min}\% to \num{x6-abst-max}\% across models, and nothing stops a model from ignoring
the prompt. With the harness the result depends far less on the model.

\paragraph{Stronger models do not notice the fault.} Without a guard, GPT-5.6 Sol, Terra and Luna
report a wrong count in \num{fr-none-wrong-min}--\num{fr-none-wrong-max}\% of faulted runs
(Table~\ref{tab:frontier}). Told to run the check themselves they give few wrong answers
(\pct{fr-selfverify-wrong}) but refuse most clean tasks as well (\pct{fr-selfverify-clean} correct).
A silently wrong count looks like a right one. The evidence has to be produced by extra queries, and
the harness is the place that can do so without changing how the model answers clean tasks.

\paragraph{The comparison with published methods is fair.} All methods share one harness, the same models
and the same tasks, and their prompts follow each paper's design (App.~\ref{app:baselines}). We gave
the LLM judge the harness's own consistency queries and a strong judge model, and the reviewer up to
five loops. The published methods either leave the decision to the model or use checks learned from
value patterns, and none of them issues extra queries to test a result. On ToolMaze, a benchmark built by
others for tool failures, \sgp{} is still slightly ahead of them.

\paragraph{No method was tuned on the test data.} Method design used separate ToolMaze workflows,
\sgp{} was frozen by file hashes before the planned runs, and every prediction was written down
before its run (App.~\ref{app:prereg}).

\paragraph{Cost.} On AgentOrchBench the cost is the harness's own check queries, with no extra model call
(Table~\ref{tab:app-cost}). On ToolMaze, false alarms of learned checks cost
\num{gain-toolmaze-clean} points of clean success against Outcome Monitors (Table~\ref{tab:other}).

\paragraph{A consistently lying API defeats every check.} We tested this directly. The consistent-lie
fault is an API built to keep the count identity true on the queries a checker sends. The full check
catches only \pct{bud-all-lie} of its answers, and choosing the checked conditions at random does not
help (Table~\ref{tab:app-budget}). When an API answers every query in one consistent false world, no
check built from its own answers can tell. Only an independent source of truth can, and every method
we tested reports the wrong count. We list this as a limitation.

\paragraph{The cutoff is not tuned to the results.} No decision depends on it in the range that
matters. Every cutoff below \pct{gate-safe-tau} gives the same decisions on ToolMaze, and on
AgentOrchBench and real APIs every check is exact (Table~\ref{tab:app-gate}).

\begin{table*}[t]
\centering
\small
\setlength{\tabcolsep}{4pt}
\begin{tabular}{@{}lp{9.2cm}lc@{}}
\toprule
\rowcolor{head}
\textbf{ID} & \textbf{Prediction, written before the run} & \textbf{Observed} & \textbf{Outcome}\\
\midrule
A1 & SilentGuard $>$ blind retry on lasting faults (ToolMaze P4) & 24.4 vs 27.0 & \textbf{\textcolor{loss}{failed}}\\
A2 & SilentGuard $\geq$ blind retry $-2$ on one-time faults (P3) & 47.6 vs 60.4 & \textbf{\textcolor{loss}{failed}}\\
A3 & Blind retry repeats action calls & 2.36 vs 0.24 per run & held\\
B1 & Loud faults: SilentGuard within 3 points of Checklist + OM & 39.8 vs 39.8 & held\\
C1 & AgentOrchBench wrong within 5 points of Enforce & 1.7 vs 8.8 & \textbf{\textcolor{loss}{failed}}\\
C2 & AgentOrchBench correct $\geq$ answer gate $-5$ & 2.4 vs 5.2 & held\\
D1 & Checklist, Reviewer, Outcome Monitors all above 40\% wrong & 72.3, 73.8, 72.2 & held\\
D2 & SilentGuard fewer wrong than each published method on 7/7 models & 7/7, 7/7, 7/7 & held\\
D3 & Outcome Monitors within 10 points of no guard & 72.2 vs 74.0 & held\\
E1 & Retail: wrong and clean within 5 points of Enforce & 11.0/38.5 vs 12.3/33.8 & held\\
F1 & Real APIs (first 4): wrong $\leq$ best single check $+3$, and correct $\geq$ spec check $-5$ & 38.6 vs 35.7; 50.0 vs 51.4 & held\\
R1 & AgentOrchBench: blind retry within 5 points of no guard & 74.4 vs 74.0 & held\\
R2 & AgentOrchBench: re-query within 10 points of no guard & 74.9 vs 74.0 & held\\
R3 & SilentGuard fewer wrong than blind retry on 7/7 models & 7/7 & held\\
BR1 & Two-call fault: blind retry within 5 points of no re-check & 56.0 vs 32.5 & \textbf{\textcolor{loss}{failed}}\\
BR2 & Two-call fault: SilentGuard $\geq$ blind retry & 46.4 vs 56.0 & \textbf{\textcolor{loss}{failed}}\\
BR3 & Real APIs (first 4): blind retry within 5 points of no guard & 90.7 vs 87.9 & held\\
L1 & Lists: SilentGuard wrong within 3 points of checked lists only (seen, unseen) & 9.5/11.4 vs 13.1/13.7 & \textbf{\textcolor{loss}{failed}}\\
L2 & Lists: SilentGuard fewer wrong than no guard and auto-repair (seen, unseen) & 9.5/11.4 vs 51.0/40.0, 17.0/19.7 & held\\
L3 & Lists: clean correct within 5 points of no guard & 82.2 vs 77.8 & held\\
K1 & Judge + blocking has more wrong answers than SilentGuard on 6/6 models & 6/6 & held\\
K2 & Outcome Monitors + blocking within 10 points of Outcome Monitors (wrong) & 69.4 vs 70.8 & held\\
K3 & Judge + blocking clean correct at least 10 points below no guard & 78.7 vs 82.6 & \textbf{\textcolor{loss}{failed}}\\
P1 & Judge + blocking without check results: at least 20 points more wrong than with them & 69.4 vs 6.1 & held\\
P2 & Judge + blocking without check results: clean correct within 10 points of with them & 81.2 vs 78.7 & held\\
X1 & Judge without check results, advising: wrong within 10 points of no guard & 70.4 vs 73.0 & held\\
X2 & Judge in four settings: evidence and blocking at least 40 points below each other cell & 6.1 vs 70.4, 69.4, 69.8 & held\\
X3 & Warning (last-condition check) + permission: fewer wrong than the warning alone, but more than 20\% & 11.9 vs 61.3 & \textbf{\textcolor{loss}{failed}}\\
Y1 & Full-check warning + permission: wrong on ignore field below 15\% & 19.2 & \textbf{\textcolor{loss}{failed}}\\
Y2 & Full-check warning + permission: overall wrong between 3 and 8\% & 19.6 & \textbf{\textcolor{loss}{failed}}\\
Y3 & Full-check warning + permission: more wrong than SilentGuard on at least 4 of 6 models & 5/6 & held\\
T1 & Typed tool calls: wrong within 5 points of the spec check (real APIs) & 28.6 vs 21.0 & \textbf{\textcolor{loss}{failed}}\\
T2 & Typed tool calls: more than 30\% wrong on NCBI, UniProt and Europe PMC & 86, 37, 57 & held\\
T3 & Typed calls + SilentGuard: no more wrong than typed calls on at least 6 of 7 models & 4/7 & \textbf{\textcolor{loss}{failed}}\\
V1 & Wrong value format (iNaturalist): no guard wrong at least 80\% & 97.1 & held\\
V2 & Wrong value format: spec check alone within 10 points of no guard (wrong) & 100.0 vs 97.1 & held\\
V3 & Wrong value format: SilentGuard wrong at most 20\% and at least 50 points below the spec check & 20.0 vs 100.0 & held\\
V4 & Wrong value format: SilentGuard correct at least 50\% & 2.9 & \textbf{\textcolor{loss}{failed}}\\
V5 & Wrong value format: no guard with the right docs correct at least 85\% & 88.6 & held\\
\bottomrule
\end{tabular}
\caption{Predictions of the confirmatory experiments, written down before each run, and their outcomes (26 of 39 held). All were written, with the method frozen by file hashes, before the runs. Failed predictions (bold red) are reported and discussed in the text (blind retry, \S\ref{sec:res-retry}). IDs are the labels used in the prediction log. Observed values are means over the 7 open models (\%), or the number of models with fewer wrong answers. Rows L, K, P, X and Y use the 6 models the predictions were written for, so their no-guard values differ slightly from the 7-model tables. Within $x$ points: the absolute difference is at most $x$. The rows from L down are control runs added after the first results. Each was written before its own run.}
\label{tab:app-prereg}
\end{table*}

\end{document}